\documentclass[preprint,12pt]{elsarticle}

\usepackage{graphicx}
\usepackage{amssymb}

\usepackage{hyperref}

\usepackage{lineno}

\usepackage{tabularx}
\usepackage{makecell}

\usepackage{tikz}
\usetikzlibrary{tikzmark,calc,matrix,positioning}

\tikzstyle{inputNode}=[draw,circle,minimum size=10pt,inner sep=0pt]
\tikzstyle{stateTransition}=[-stealth, thick]

\journal{Experimental Neurology}

\begin{document}

\begin{frontmatter}


\title{Promises and pitfalls of deep neural networks in neuroimaging-based psychiatric research}



\author[label1]{Fabian Eitel\corref{cor1}} 
\author[label1]{Marc-André Schulz\corref{cor1}} 
\author[label1]{Moritz Seiler\corref{cor1}}
\author[label1]{Henrik Walter} 
\author[label1]{Kerstin Ritter\fnref{fn1}} 

\address[label1]{Charité -- Universitätsmedizin Berlin, corporate member of Freie
Universität Berlin, Humboldt-Universität zu Berlin, and Berlin Institute of
Health; Department of Psychiatry and Psychotherapy, Bernstein Center for Computational Neuroscience; 10117 Berlin, Germany}

\cortext[cor1]{Equal contribution}
\fntext[fn1]{Corresponding author: kerstin.ritter@charite.de}

\begin{abstract}
By promising more accurate diagnostics and individual treatment recommendations, deep neural networks and in particular convolutional neural networks have advanced to a powerful tool in medical imaging. Here, we first give an introduction into methodological key concepts and resulting methodological promises including representation and transfer learning, as well as modelling domain-specific priors. After reviewing recent applications within neuroimaging-based psychiatric research, such as the diagnosis of psychiatric diseases, delineation of disease subtypes, normative modeling, and the development of neuroimaging biomarkers, we discuss current challenges. 
This includes for example the difficulty of training models on small, heterogeneous and biased data sets, the lack of validity of clinical labels, algorithmic bias, and the influence of confounding variables. 

\end{abstract}


\end{frontmatter}

%

\section{Introduction}
\label{S:1}

By setting new standards in image and speech recognition tasks, deep neural networks advanced to a key technology in research \citep{Lecun2015, schmidhuber2015deep}. In medical imaging, a number of applications have reached or even exceeded human level performance, especially in cases where the sample size was rather large ($N > 15000$) and the prediction task was well defined (i.e., the pathology is clearly identifiable in the imaging data). This includes, for example, the detection of diabetes from fundus images, skin cancer from photographs, and pneumonia from chest X-rays \citep{esteva2017dermatologist, gulshan2016development}.
Given these success stories, it has been asked to what extent deep neural networks are also capable of identifying brain diseases based on neuroimaging data, e.g., data obtained from magnetic resonance imaging (MRI; \citep{litjens2017}). While most neurological diseases are associated with measurable brain damage, such as atrophy or lesions, that is visible in structural MRI, brain alterations in psychiatric disorders are considered to be more subtle, mostly functional and still under debate \citep{goodkind2015identification, lui2016psychoradiology}. 
Nevertheless, in the last two decades, neuroimaging has become one of the cornerstones in the search for biomarkers that explain neurobiological variance associated with psychiatric disease \citep{lui2016psychoradiology,fu2013neuroimaging}. 
Promising biomarkers include regional atrophy and reduced cortical thickness \citep{zhao2014brain, schmaal2016subcortical}, brain age \citep{cole2017predicting} as well as alterations in task-induced functional MRI activity or resting-state connectivity \citep{ patel2012neurocircuitry}; for a review, see \citet{lui2016psychoradiology}.  
However, such biomarkers are not necessarily disease-specific and have a high overlap across psychiatric (and neurological) diseases \citep{abi2016search}.

In addition to traditional statistical analyses, where mean differences between groups (e.g., patients and controls) have been investigated, neuro\-imaging-based biomarkers have also been employed in machine learning analyses with the goal to draw conclusions about individual subjects \citep{Kloeppel2011, Orru2012, Wolfers2015, bzdok2018machine, walter2019translational}.
By being settled in the framework of precision medicine, machine learning is considered to provide a huge promise for transforming healthcare in general \citep{Obermeyer2016} and psychiatry in particular \citep{durstewitz2019deep}. While machine learning approaches can been applied to different kinds of data including deep phenotyping, genetics, and metabolomics, they are in particular a suitable candidate for analyzing neuroimaging data due to their ability to flexibly handle high-dimensional data where the number of variables (e.g., voxels or regional volumes) commonly exceeds the number of samples (see Table \ref{ml_terms} for a brief description of machine learning related terms).
The learning settings range from automatic disease diagnosis and prognosis to subtype discovery and prediction of treatment outcome \citep{Wolfers2015,Marquand2016,lueken2016neurobiological, Woo2017BuildingNeuroimaging}.

\begin{table}[h]
    \centering
    {\scriptsize
    \begin{tabularx}{\linewidth}{lX}
        \hline
        \textbf{Term} & \textbf{Description}\\
        \hline
        & \\
        \textbf{Machine learning} &  
\makecell[Xt]{The field of \emph{machine learning} comprises algorithms that learn to perform a task solely based on data, without being explicitly programmed to do so.}\\
        \tikzmark{s1} & \\
         \hspace{0.3cm} 
          \textbf{Deep learning} & \makecell[Xt]{\emph{Deep learning} is a particular subtype of machine learning which is based on DNNs.}\\
         \hspace{0.3cm} 
         \makecell[Xt]{Artificial neural network (ANN)/\\Deep neural network (DNN)}& \makecell[Xt]{\emph{Artificial neural networks} describe  a  class  of  models that consists of a  collection of  connected  units  or  artificial  neurons within layers.\\ If the network architecture  consists of multiple layers between the input and output layer, an ANN is also called a \emph{deep neural network}.}\\
        \hspace{0.3cm} \tikzmark{s2} & \\
         \hspace{0.6cm} 
          \textit{Fully-connected neural network} & \makecell[Xt]{A \emph{fully-connected neural network} is the most basic type of ANN. It consists of layers that connected every artificial neuron with every artificial neuron in the subsequent layer.}\\
         \hspace{0.6cm} 
          \textit{Convolutional neural network (CNN)} & \makecell[Xt]{A \emph{convolutional neural network} is a particular type of ANN that uses the mathematical convolution operation and is predominantly used in image analysis.}\\
         \hspace{0.6cm} 
          \textit{Recurrent neural network (RNN)} & \makecell[Xt]{A \emph{recurrent neural network} is another type of ANN which introduces recurrent connections between the artificial neurons; and is, thus, primarily used on sequential data.}\\
        \tikzmark{s3} \hspace{0.2cm} \tikzmark{s4} & \\
        \hline
    \end{tabularx}}
    \caption{Description and interdependencies of central terms in machine learning and deep learning.}
    \label{ml_terms}
\begin{tikzpicture}[overlay,remember picture,xshift=-1.6em,line width=.7pt,draw=gray!60]
\path (pic cs:s1)|-coordinate(E)(pic cs:s3);
\draw ($(pic cs:s1)+(0,\ht\strutbox)$)--(E) 
($(pic cs:s2)+(0,\ht\strutbox)$)--(pic cs:s4); 
\end{tikzpicture}
\end{table}

Most classical machine learning algorithms learn comparably simple input-output functions. They therefore have difficulties in processing raw MRI data and rely on hand-crafted features such as cortical thickness or connectivity matrices \citep{Lecun2015, Vieira2017UsingApplications, 2017inference, bzdok2018points}. In psychiatry, the design of those features is build on neurobiological disease models and usually reflects only one part of disease pathology, but does not meet the multifactorial nature of many psychiatric diseases \citep{abi2016search}.
Deep learning approaches, on the other hand, can learn hierarchical representations directly from the raw data and thus are capable of solving more complex problems \citep{Lecun2015, schmidhuber2015deep}. 
Major breakthroughs have been achieved in natural image recognition by using a specialized deep learning architecture called convolutional neural networks (CNNs); which are artificial neural networks (ANNs) that exploit the structural dependencies of data coming from arrays such as image, audio, and video signals \citep{Lecun2015}.
The key idea behind CNNs is inspired by the mechanism of receptive fields in the primate’s visual cortex and relates to the application of local convolutional filters in combination with downsampling \citep{Hubel1968, schmidhuber2015deep}. In addition to their utilization in industry for diverse image and speech recognition tasks, they have advanced to a strong instrument for analyzing medical imaging data, including some aspects of neuroimaging data \citep{litjens2017,Vieira2017UsingApplications, LUNDERVOLD2019102}. Besides promising preliminary results in the field of neurology and psychiatry \citep{Vieira2017UsingApplications, valliani2019deep}, a number of open challenges still exist. 
On the one hand, these challenges can be technology-related, such as the difficulty of deep neural networks to learn robust models on small, heterogeneous data sets or to give meaningful explanations for complex model decisions \citep{Boehle2019, Eitel2019MS, Schulz2019dlbrains, he2020deep}. On the other hand, these challenges are driven by factors that are controversial within psychiatry itself, e.g., the heterogeneity of psychiatric diseases in their clinical presentation and the reliance on clinical symptoms rather than neurobiological substrates for establishing disease categories \citep{rokham2020addressing, Cuthbert2013, Insel2015}.

In this review, we will first give a short introduction into MRI and its role in psychiatry (section \ref{sec:MRI}) and then introduce the basic concepts of machine and deep learning with a focus on CNNs as the most popular type of deep neural network currently applied to neuroimaging data (section \ref{sec:methods}; please see \citet{durstewitz2019deep} for an introduction into recurrent neural networks 
for analyzing temporal data in psychiatry).  Both concepts are needed in order to understand the following sections; readers interested in more general introductions of machine learning in medicine and psychiatry are referred to \citep{bzdok2018machine,rajkomar2019machine,  rowe2019introduction}. 
In section \ref{sec:meth_prom}, we will outline three methodological promises for the use of deep learning in psychiatry, namely representation learning, transfer learning, and the use of model architectures incorporating neuroimaging-specific priors (a.k.a. inductive bias).
In sections \ref{S:3} and \ref{sec:applications}, we will first present promising application scenarios within psychiatry, namely automated diagnostics, subtype identification, and development of new biomarkers; and then review existing applications for major psychiatric disorders including Alzheimer's disease, schizophrenia, substance abuse, neurodevelopmental disorders, and internalizing disorders such as depression, anxiety, and obsessive-compulsive disorder.
Finally, current challenges and implications in the use of
deep neural networks will be discussed in section \ref{sec:challenges}. 


\section{Magnetic Resonance Imaging (MRI) and its role in psychiatry}\label{sec:MRI}
 After a short introduction into the role of MRI in psychiatric research, we will present the most common MRI modalities (and preprocessing steps) that have been used at the intersection of psychiatry and machine learning. 
 MRI is a non-invasive medical imaging technique which has become an important tool in the diagnosis and monitoring of neurological diseases including stroke, multiple sclerosis and tumor detection \citep{schellinger2010evidence, geraldes2018current}. In psychiatry, MRI data is mainly used to rule out other causes (e.g., brain tumors) that might explain changes in behavior and feelings but also became a cornerstone for investigating neurobiological correlates in psychiatric diseases \citep{lui2016psychoradiology}. The focus in psychiatric research, using structural MRI (sMRI), is on high-resolution T1-weighted MRI data (e.g., Magnetization Prepared RApid Gradient Echo (MPRAGE) \citep{mugler1990three}) providing the best contrast between grey and white matter which is therefore useful for measuring cortical thickness and regional atrophy in the brain.
White matter abnormalities can be identified using T2-weighted imaging (e.g., Fluid-Attenuated Inversion Recovery (FLAIR)), magnetization transfer imaging, and diffusion tensor imaging but thus far play only a minor role in machine learning analyses. 
Although structural changes in patients with psychiatric diseases are considered to be small and difficult to detect by visual inspection, some relative consistent findings have been reported, e.g., smaller hippocampi in patients with depression or larger ventricles in patients with schizophrenia \citep{moncrieff2010systematic, campbell2004role, lui2016psychoradiology}. 

Functional MRI (fMRI), in contrast to sMRI, measures brain activity by changes in blood oxygenation and blood flow as a result of neural activity (the so-called blood oxygen level dependent (BOLD) response \citep{glover2011overview}). Whereas in task-based fMRI, the activation during specific tasks (e.g., processing of emotional faces) is examined, in resting-state fMRI it is the intrinsic network activity during rest (i.e., no task) that is investigated.
Specifically, resting-state functional connectivity matrices capturing spatial correlations in BOLD fluctuations are
thought to give important insights about the functional organization of the healthy and diseased brains and have been employed in machine and deep learning \citep{khosla2019machine, pervaiz2020optimising}. However, the within- and between-subject variability is considerable and makes finding robust functional biomarkers challenging \citep{specht2019current}. 

Most studies applying CNNs on MRI so far have focused on sMRI (in particular T1-weighted MPRAGE) rather than fMRI. There are two main reasons for this focus. First, sMRI shares more properties with the modalities where CNNs have been successful (e.g., natural images); fMRI, in contrast, does not contain clearly visible hierarchical objects such as brain regions with high-frequency edges. Second, given that fMRI is 4-dimensional, it makes computation very expensive on the raw data (see section \ref{sec:compu}), or leads to the aggregation across time or space, potentially losing relevant information. In the future, advanced MRI techniques such as quantitative MRI may play an important role \citep{LUNDERVOLD2019102}. 

To study the same brain regions across subjects, MRI data are usually spatially normalized to a common space (e.g., the Montreal Neurological Institute [MNI] template) \citep{Ashburner2007}. 
Deep learning, on the other hand, does not necessarily require spatial normalization but it may help in reducing variance in the data especially in light of small sample sizes. Therefore, most deep learning studies at least linearly register the data to MNI space \citep{Eitel2019MS}. Additionally, deep learning algorithms have also been used to perform nonlinear registration by themselves (for an overview of deep learning applications in MRI aside from clinical applications, see \citep{LUNDERVOLD2019102}). 



To analyze diseases trans-diagnostically and on a larger scale, a number of large multi-site imaging cohorts have been elicited
(e.g., HCP\footnote{\url{http://www.humanconnectomeproject.org/}}, 
UK biobank\footnote{\url{https://www.ukbiobank.ac.uk/}}, ENIGMA\footnote{\url{http://enigma.ini.usc.edu/}}, and IMAGEN\footnote{\url{https://imagen-europe.com/resources/imagen-dataset/}}). The existence of large data bases is essential for the success of machine and deep learning techniques in psychiatric research (see section \ref{sec:challenges}).

\section{Key ideas of machine and deep learning}\label{sec:methods}
To make the methodological promises and challenges of deep learning technology within psychiatric research more accessible, we first briefly introduce the reader to some fundamental concepts of the more general field of machine learning, including different learning settings and validation schemes. We then concentrate on the sub-field of deep learning, which is the focus of the later explanations. Therefore, we briefly describe the underlying principles of deep learning, namely artificial neural networks (ANNs), and convolutional neural networks (CNNs) in particular since this type of ANN is predominantly used to analyze image data, as well as methods to explain individual predictions of these types of models.
 While experienced readers might skip this section, novice readers are encouraged to consider \citep{Bishop2007, Goodfellow2016} for a comprehensive coverage of the fundamentals of machine learning and deep learning or \citep{Wolfers2015, bzdok2018machine, rajkomar2019machine, rowe2019introduction} for its applications to medicine.

\subsection{Machine learning}

\subsubsection{Basic concepts}

\begin{figure}[t]
\def\svgwidth{\linewidth}
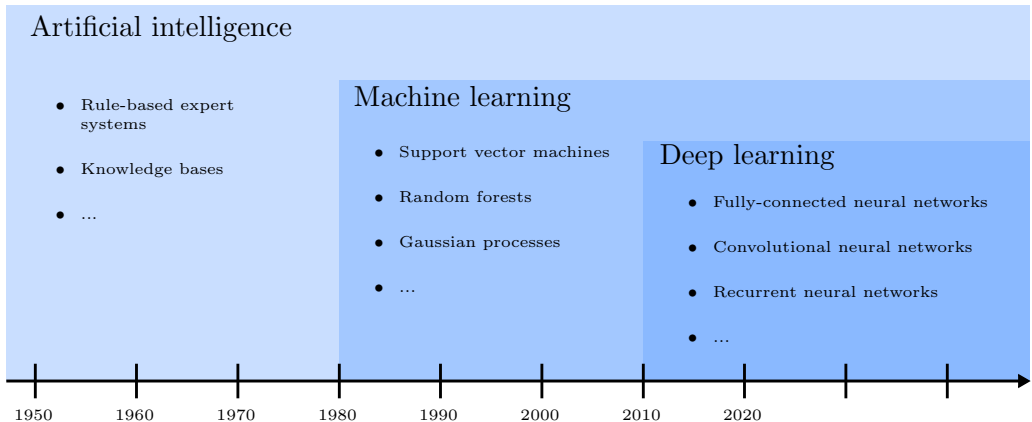
\caption{Artificial intelligence became popular in the mid' 1900s using large tables of hand designed decision strategies, which for example beat human players in chess. Starting in the 1980s the field of machine learning gained traction and showed that the previously hand designed strategies could be learned from data. The so far unpopular methods around ANNs had a breakthrough in the early 2010s when large data sets and computing power allowed them to become deeper and more complex, and thus outperformed other methods in many disciplines including image and speech recognition.}
\label{fig:ml_dl}
\end{figure}

Machine learning is an interdisciplinary field at the intersection of computer science, statistics, mathematics, and others, which in general defines algorithms that learn from data. In contrast to traditional rule-based systems in artificial intelligence, where human knowledge is encoded in rigid rules, machine learning algorithms can learn to perform a specific task (e.g., discriminating between images of dogs and cats) without being explicitly programmed \citep{Bishop2007, Buchanan1983} (see Figure \ref{fig:ml_dl}).
This flexibility allows learning algorithms to be applied to complex problems which are not feasible for hand-tuned approaches \citep{Juang1991, Rosten2006, Libbrecht2015}.

\begin{figure}
    \begin{center}
        \input{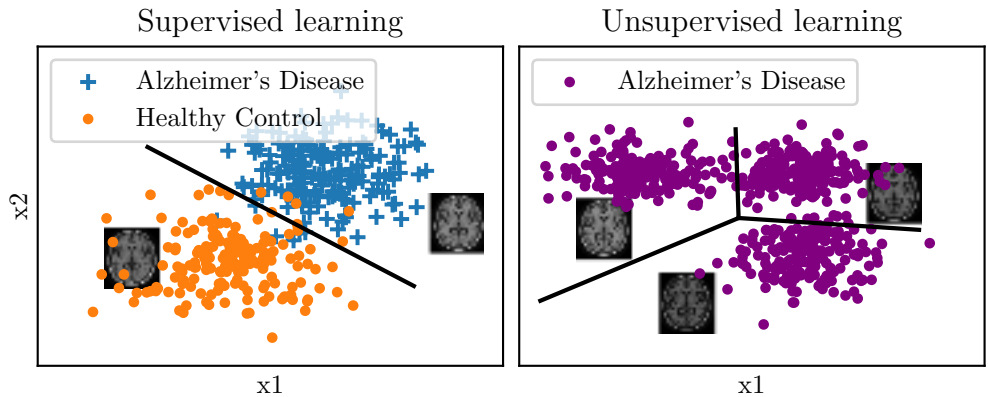}
    \end{center}
    \caption{Examples of supervised and unsupervised machine learning. In supervised machine learning such as classification (left) each brain MRI is a data point (represented as circles and crosses) and has a pre-determined class such as patient or control. The task is to find a model (black line) that discriminates between those classes. In unsupervised machine learning such as clustering (right) each brain MRI is again a data point but conversely does not have a class label. Here, the task is to find clusters in the data which can be well separated. These clusters could for example represent subgroups of a disease.}\label{fig:un_sup}
\end{figure}

Most machine learning algorithms can be categorized into two main para\-digms: supervised learning and unsupervised learning (see Figure \ref{fig:un_sup}). 
In supervised learning, a data set contains examples of input-output pairs, and a function that maps the input features to the output labels is learned. 
In psychiatry, the input features could be a set of biomarkers (e.g., volumes of particular brain structures), while the label is given by 
group membership (e.g., patients or healthy controls), disease severity (e.g., the extent of positive symptoms in schizophrenia), prognosis, or treatment outcome (e.g., response to cognitive-behavioral psychotherapy). After learning this mapping, the algorithm can then be used to make statements about new, unseen subjects. 
If the output labels are discrete classes, the prediction task is called classification, while for continuous labels, it is known as regression.
In contrast, in unsupervised learning, the data set only contains input features without any information about the labels. 
Often, the goal of this paradigm is to discover a compact and more informative representation of the data. 
An example here is the identification of disease clusters or subtypes solely based on neuroimaging data.

To enable a machine learning algorithm to learn performing such tasks from data, four components are usually used to define the algorithm: (1) a data set (which is split into training and test set), (2) a statistical model to learn an approximate representation of the data, (3) a loss function to measure the goodness of the model, and (4) an optimization procedure to alter the model parameters in a way that the loss is minimized. 
Concerning the statistical models, machine learning comprises a large number of model classes, of which in particular, support vector machines (SVMs), Gaussian processes, random forests, logistic regression, and ANNs are common choices in the neuroimaging domain \citep{Wolfers2015,Bishop2007, mateoperez2018survey}.

\subsubsection{Validation and performance metrics}\label{sec:val_metrics} 
The central challenge in machine learning is the ability of the learning algorithm to perform well on new, unobserved data; this is also called generalization \citep{hastie2009elements}. Therefore, a quantitative measure of predictive performance, such as the model error, is required to test this ability. Additionally, according to the `No Free Lunch Theorem' \cite{Wolpert1997} in machine learning, there is no single model which performs universally best on all possible data sets. Therefore, two different procedures are performed to identify the best performing model on a data set which furthermore generalizes well to unseen data: model selection and model assessment \cite{hastie2009elements}. 

Model selection is the process of selecting the best of many competing models. Since most machine learning algorithms have hyperparameters which must be set manually, finding the optimal hyperparameter configuration is performed as part of this model selection step. This step is crucial to adjust the expressive capacity (often called expressivity) of the model to match the complexity of the task. A mismatch between model capacity and task complexity is a common problem in machine learning.
In particular, overfitting, in which the model learns noisy and too complex patterns in the data set without improving the generalization, results from too much model capacity for the task, and therefore, requires regularization to penalize this capacity. 
In the model assessment, the generalization ability of a model is evaluated based on the generalization error of the model. 

This generalization error of the learning algorithm must be approximated, which is usually done by the holdout validation method. The data set is split by randomly partitioning it into three disjoint subsets: a training set, a validation set, and a test set. These subsets are assumed to be independent and identically distributed (i.i.d.) samples from an unknown data generating distribution. The i.i.d. assumption is crucial to receive an unbiased estimate of the generalization error, although violations are common in neuroimaging-based machine learning \cite{wen2019serious}, for example, resulting from data leakage by splitting clinical data containing repeated measurements on the observation-level rather than the subject-level \cite{wen2020convolutional}.
Based on the training set, different models are fitted and evaluated on the validation set by using the model errors as an estimate of the generalization error. The best model is then selected and tested on the holdout test set using the test error as an estimate of the generalization error of the model.\footnote{Please note that the model selection error on the validation set will underestimate the true error of the models and should therefore not be used as an estimate of the models' generalization ability.}


In practice, especially in clinical applications, the data set sizes are usually small such that the single train-validation-test split might result in an inaccurate estimate of the generalization error \cite{hastie2009elements}. 
The most common technique to address this issue is cross-validation (CV). In CV, the data set is partitioned into several different subsets, so-called folds. By choosing one of the folds as the test set, while the others are used as a training set, it is possible to repeat this training and testing computation until each fold has been used as a test set. The estimate of the generalization error is then an average over the errors on the test set of each repetition. A special case of CV is leave-one-out CV, where the number of folds is equivalent to the number of samples in the data set. This form of CV has been used in clinical research (e.g., \citep{Kloppel2008, Weygandt2011}) due to particularly small data set sizes. However, \citet{Kohavi1995} has empirically shown that although leave-one-out estimates are almost unbiased, the variance of the estimates can be large. Besides, it has been argued that repeated random splits lead to more stable results \citep{varoquaux2017assessing}. More recently, in multi-site studies, so-called leave-one-site-out CV is used (e.g, \citep{jollans2019quantifying, koutsouleris2018prediction}). Here, the data from different sites are pooled so that the data of each site being a separate fold in the CV framework.
If both model selection and model evaluation are performed, CV needs to be extended to receive an unbiased estimate of the generalization error \cite{varma2006bias}. This extension is called nested CV and is merely a nesting of two CV loops, performing model selection in the inner loop while evaluating the final model in the outer loop. 

The predictive performance of the final model is then reported using a performance metric. Common metrics include (balanced) accuracy, sensitivity and specificity, area under the receiver operating characteristic curve (ROC AUC), F-1 score for classification, and mean-squared-error (MSE) for regression. While some metrics such as accuracy might be strongly influenced by the class distributions, others try to correct for that, such as balanced accuracy, ROC AUC and F-1 score.




\subsection{Deep learning}

\subsubsection{Artificial neural networks}
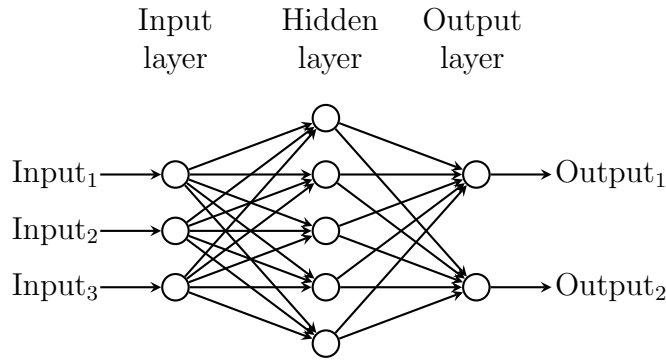
\begin{figure}[ht!]
    \centering
    \begin{tikzpicture}
    	\node[inputNode, thick] (i1) at (6, 0.75) {};
    	\node[inputNode, thick] (i2) at (6, 0) {};
    	\node[inputNode, thick] (i3) at (6, -0.75) {};
    	
    	\node[inputNode, thick] (h1) at (8, 1.5) {};
    	\node[inputNode, thick] (h2) at (8, 0.75) {};
    	\node[inputNode, thick] (h3) at (8, 0) {};
    	\node[inputNode, thick] (h4) at (8, -0.75) {};
    	\node[inputNode, thick] (h5) at (8, -1.5) {};
    	
    	\node[inputNode, thick] (o1) at (10, 0.75) {};
    	\node[inputNode, thick] (o2) at (10, -0.75) {};
    	
    	\draw[stateTransition] (5, 0.75) -- node[xshift=-1cm] {Input$_1$} (i1);
    	\draw[stateTransition] (5, 0) -- node[xshift=-1cm] {Input$_2$} (i2);
    	\draw[stateTransition] (5, -0.75) -- node[xshift=-1cm] {Input$_3$} (i3);
    	
    	\draw[stateTransition] (i1) -- (h1);
    	\draw[stateTransition] (i1) -- (h2);
    	\draw[stateTransition] (i1) -- (h3);
    	\draw[stateTransition] (i1) -- (h4);
    	\draw[stateTransition] (i1) -- (h5);
    	\draw[stateTransition] (i2) -- (h1);
    	\draw[stateTransition] (i2) -- (h2);
    	\draw[stateTransition] (i2) -- (h3);
    	\draw[stateTransition] (i2) -- (h4);
    	\draw[stateTransition] (i2) -- (h5);
    	\draw[stateTransition] (i3) -- (h1);
    	\draw[stateTransition] (i3) -- (h2);
    	\draw[stateTransition] (i3) -- (h3);
    	\draw[stateTransition] (i3) -- (h4);
    	\draw[stateTransition] (i3) -- (h5);
    	
    	\draw[stateTransition] (h1) -- (o1);
    	\draw[stateTransition] (h1) -- (o2);
    	\draw[stateTransition] (h2) -- (o1);
    	\draw[stateTransition] (h2) -- (o2);
    	\draw[stateTransition] (h3) -- (o1);
    	\draw[stateTransition] (h3) -- (o2);
    	\draw[stateTransition] (h4) -- (o1);
    	\draw[stateTransition] (h4) -- (o2);
    	\draw[stateTransition] (h5) -- (o1);
    	\draw[stateTransition] (h5) -- (o2);
    	
    	\node[above=of i1, align=center] (l1) {Input \\ layer};
    	\node[right=1.5em of l1, align=center] (l2) {Hidden \\ layer};
    	\node[right=0.7em of l2, align=center] (l3) {Output \\ layer};
    	
    	\draw[stateTransition] (o1) -- node[xshift=1.2cm] {Output$_1$} (11, 0.75);
    	\draw[stateTransition] (o2) -- node[xshift=1.2cm] {Output$_2$} (11, -0.75);

    \end{tikzpicture}
    \caption{Architecture of a fully-connected network with interconnected groups of neurons in the input layer, a single hidden layer and the output layer. }
    \label{mlp_architecture}
\end{figure}

Deep learning is a particular subtype of machine learning which is based on ANNs. These ANNs describe a large class of models that consists of a collection of connected units or artificial neurons. Although these models have recently gained attraction after achieving state-of-the-art results in fields such as computer vision \cite{Szegedy2014} or natural language processing \cite{Sutskever2014}, the origins of these models date back to the 1940s and were used to study biologically inspired representations of information processing \cite{Mcculloch1943}.
To describe the working mechanisms of ANNs, we first introduce the most basic type of these networks, the multilayer perceptron (MLP) or fully-connected neural network. This type is a feed-forward neural network which is constructed by connecting groups of artificial neurons or units organized within layers (see Figure \ref{mlp_architecture}). Unlike recurrent neural networks \citep{hopfield1982neural, hochreiter1997long}, feed-forward neural networks only allow the input information to flow in one direction, from input to output, without loops between the neuron connections. 
For this, an MLP consists of an input layer, one or more hidden layers and an output layer. While the input layer passes the input features to the network using one unit for each input feature (e.g., one unit for every voxel in an MR image), the output layer completes the task 
by outputting a prediction, with the number of units depending on the task (e.g., one unit for patients and one unit for healthy controls in a binary classification task). The hidden layers define the capacity of ANNs, with the number of units in a hidden layer defining the width and the number of hidden layers defining the depth of the network \cite{Goodfellow2016}. An ANN architecture using more than one hidden layer is called a deep neural network; the name deep learning is derived from this terminology and therefore serves as a collective term for ANNs with multiple hidden layers.

In an MLP, a fully-connected layer connects each artificial neuron in a layer with every artificial neuron in the previous layer, enabling the flow of information between sets of artificial neurons. Each unit thus receives an input from every artificial neuron in the previous layer connected to it to compute a weighted sum of these connections using the associated weight parameters. Therefore, the weights can be seen as the relative strength or importance of the connections between consecutive units. Furthermore, inside a unit, a nonlinear activation function (e.g., rectifier function \cite{Glorot2011} or logistic sigmoid function) is applied to the previously computed weighted sum to nonlinearly transform the input to an output of the unit. These outputs then serve as inputs to the units in the subsequent layer, creating a nested chain structure of connected units between layers. Therefore, an MLP can be understood as a complex mathematical function composed of many simpler functions. 

Regarding the expressive power, the Universal Approximation Theorem (e.g., \cite{Cybenko1989}) states that a fully-connected neural network with one hidden layer and a sufficient number of hidden units can theoretically approximate any continuous function.\footnote{Several extensions to this theorem exist, which also include, for example, further neural network architectures (e.g., \citep{zhou2020universality}).} 
However, the required width of the hidden layer may be infeasibly large such that the fully-connected neural network may fail to learn and generalize correctly. Therefore, the capacity of an ANN model is typically increased by adding depth to the network \cite{Bengio2009}. This leads to a series of nonlinear transformations which enable an ANN, or rather a deep neural network, to hierarchically learn multiple levels of representations from raw data to abstract representations. 
Unlike traditional machine learning methods, which often rely on the discriminatory power of handcrafted features and previous feature engineering, ANNs can learn feature representations from raw data by using a general learning procedure \cite{bengio2013representation}. 
Examples of feature engineering include the delineation of lesions to calculate the total lesion load for use in a classifier for multiple sclerosis or computing a gray matter segmentation, dividing it into separate brain regions and using the mean gray matter density per region in a classifier for schizophrenia. Artificial neural networks, however, are able to process the raw images and learn similar or potentially more powerful features themselves.

To learn the optimal model parameters (i.e., weights), ANNs rely on gradient-based optimization using the backpropagation algorithm \citep{rumelhart1986}. 
During the training phase, the backpropagation algorithm uses the chain-rule to compute the partial derivatives of the loss function with respect to the weight parameters to further optimize the weight parameters using a gradient-based optimization procedure.
This whole procedure is then repeated many times to find the optimal parameter configuration of the artificial neural network model. 
An essential aspect of the tremendous success of ANNs is the efficient implementation of this learning procedure using highly optimized programming libraries such as PyTorch \cite{Paszke2019}, Tensorflow \cite{tensorflow2015-whitepaper} or Theano \cite{Bergstra2010} and the support of graphical processing units (GPUs), which enable an efficient parallelized computation of this procedure \cite{Lecun2015}.

Despite the success and the representational power of ANNs, the learning process introduces certain challenges to this class of machine learning algorithms. First, although attempts at automated approaches \cite{Elsken2018} exist, the specification of an ANN model requires manually determined design choices such as the number of hidden layers or hidden units in a layer, regularization techniques, as well as further hyperparameters. These choices are mostly guided by background knowledge and experimentation, making training an ANN quite an art.
Due to this issue, artificial neural network models are usually overparametrized using thousands or even millions of parameters and are prone to overfit the training data. To address the risk of overfitting, regularization techniques such as dropout \cite{Srivastava2014} or weight decay \cite{Krogh1992} are used in the training process to reduce the network's capacity. 
Second, the loss surface, especially of deep neural network models, is highly non-convex and possesses many local minima \cite{Li2018}. Thus, the gradient-based optimization algorithms do not guarantee to converge to the global minimum of the loss surface such that different starting configurations of the network are required. However, \citet{Choromanska2015} have shown that although deep neural networks mostly converge to local minima, the resulting models often generalize well to new data, so the problem of local minima is negligible.

\subsubsection{Convolutional neural networks}
Although a fully-connected neural network can theoretically approximate any continuous function, the wiring of the neurons can lead to certain drawbacks, especially on grid-like topologies such as MR images \cite{LeCun1995}. 
First, the number of parameters to be learned for this often high-dimensional input data can be very large due to the many required connections between the neurons in the input and the subsequent hidden layer. Second, for grid-like data such as images, the pixels or voxels of the input images are treated independently, ignoring the spatial information in form of correlations between pixels and translation invariance of objects in the image. Convolutional neural networks (CNNs) \cite{LeCun1989} use the mathematical convolution operation within a convolutional layer to address these drawbacks by introducing a biologically inspired local receptive field \cite{Hubel1968}, weight sharing, and downsampling.

\begin{figure}[ht!]
    \centering
    \begin{tikzpicture}

	\matrix (mtr) [matrix of nodes,row sep=-\pgflinewidth, nodes={draw}]
	{
		0 & 1 & 1 & |[fill=green!30]| 1 & |[fill=green!30]| 0 & |[fill=green!30]| 0 & 0\\
		0 & 0 & 1 & |[fill=green!30]| 1 & |[fill=green!30]| 1 & |[fill=green!30]| 0 & 0\\
		0 & 0 & 0 & |[fill=green!30]| 1 & |[fill=green!30]| 1 & |[fill=green!30]| 1 & 0\\
		0 & 0 & 0 & 1 & 1 & 0 & 0\\
		0 & 0 & 1 & 1 & 0 & 0 & 0\\
		0 & 1 & 1 & 0 & 0 & 0 & 0\\
		1 & 1 & 0 & 0 & 0 & 0 & 0\\
	};

	\draw[very thick, green] (mtr-1-4.north west) rectangle (mtr-3-6.south east);

	\node [below= of mtr-5-4.south, yshift=-0.2cm] (lm) {Input};

	\node[right = 0.2em of mtr] (str) {$*$};

	\matrix (K) [right=0.2em of str,matrix of nodes,row sep=-\pgflinewidth, nodes={draw, fill=blue!30}]
	{
		1 & 0 & 1 \\
		0 & 1 & 0 \\
		1 & 0 & 1 \\
	};
	\node [below = of K-3-2.south, yshift=-0.2cm] (lk) {Kernel};

	\node [right = 0.2em of K] (eq) {$=$};

	\matrix (ret) [right=0.2em of eq,matrix of nodes,row sep=-\pgflinewidth,nodes in empty cells, nodes={draw,minimum size=5.5mm, anchor=center}]
	{
		1 & 4 & 3 & |[fill=orange!30]| 4 & \\
		 &  &  &  & \\
		 &  &  &  & \\
		 & & & & \\
		 & & &  & \\
	};
	\node [below = of ret-4-3.south, yshift=-0.2cm] (lim) {Feature map};

	\draw[very thick, orange] (ret-1-4.north west) rectangle (ret-1-4.south east);

	\draw[densely dotted, blue, thick] (mtr-1-4.north west) -- (K-1-1.north west);
	\draw[densely dotted, blue, thick] (mtr-3-4.south west) -- (K-3-1.south west);
	\draw[densely dotted, blue, thick] (mtr-1-6.north east) -- (K-1-3.north east);
	\draw[densely dotted, blue, thick] (mtr-3-6.south east) -- (K-3-3.south east);

	\draw[densely dotted, orange, thick] (ret-1-4.north west) -- (K-1-1.north west);
	\draw[densely dotted, orange, thick] (ret-1-4.south west) -- (K-3-1.south west);
	\draw[densely dotted, orange, thick] (ret-1-4.north east) -- (K-1-3.north east);
	\draw[densely dotted, orange, thick] (ret-1-4.south east) -- (K-3-3.south east);

	\matrix (K) [right=0.2em of str,matrix of nodes,row sep=-\pgflinewidth, nodes={draw, fill=blue!10}]
	{
		1 & 0 & 1 \\
		0 & 1 & 0 \\
		1 & 0 & 1 \\
	};

	\draw[very thick, blue] (K-1-1.north west) rectangle (K-3-3.south east);

\end{tikzpicture}
    \caption{Example of a 2-dimensional convolution operation.}
    \label{convolution}
\end{figure}
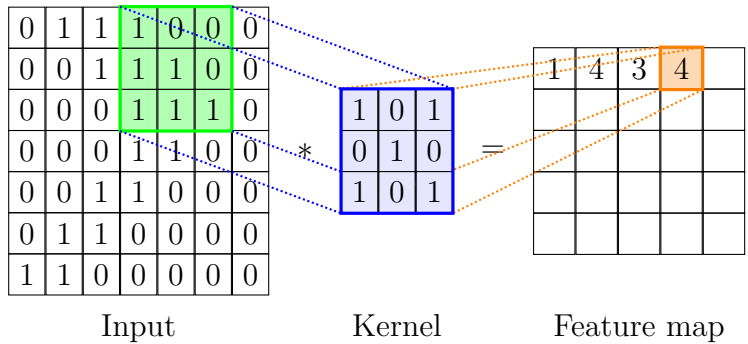

A convolution is a linear mathematical operation (see Figure \ref{convolution}) which computes the weighted sum of an input and the weight parameters of a function, also known as a kernel or filter, at every location in the input space to produce a feature map. Thus, instead of connected groups of neurons, each unit in the feature map is only locally connected to a specific region, also known as a receptive field, in the input space. This enables the extraction of local features, and therefore, preserves the spatial structure of the input space. Moreover, the units in the feature map are restricted to share the same weight parameters for every region in the input space. As a result, the number of parameters to be learnt and thereby the complexity is reduced to a set of parameters equivalent to the size of the kernel. 
Since the convolution is a linear operation, activation functions are again used to nonlinearly transform the resulting feature maps.
Typically, in a CNN, a convolution layer is followed by a pooling layer that replaces the values of a local region in the feature map with summary statistics to reduce the dimensionality of the input. 
This results in translation invariance, which is an important property when the presence of a particular feature is of importance but not its location. 
In a typical CNN architecture, successive pairs of convolutional layers and pooling layers are used to learn different levels of representations of the data, e.g. from an MR image (from edges and blobs to more abstract concepts such as lesions or atrophy). 
The final layers of a CNN architecture are typically fully-connected layers to compute the output predictions for a specific task. Most studies applying neural networks for classification in MRI follow the architectures of successful computer vision models such as DenseNets \citep{Huang2017DenselyNetworks}, ResNets \citep{He2015DeepRecognition} or VGGNet \citep{Simonyan2014a}. Yet, recently researchers have started designing architectures specialized for the properties of raw data or functional connectivity matrices introducing inductive biases (see section \ref{sec:inducbias}). 
For applications of CNNs to neuroimaging data, a couple of specialized software libraries exist (e.g., PHOTONAI \citep{leenings2020photon}, DeepNeuro \citep{beers2020deepneuro}).

\subsubsection{Explaining model predictions}\label{sec:attribution}
Despite their recent success, ANNs, and in particular, deep neural networks, are often criticized for being black-boxes \citep{Castelvecchi2016, Lapuschkin2019}. Although deep neural networks are mostly deterministic models, the large parameter space of these models, often comprising thousands or millions of parameters, and highly nonlinear interactions make it difficult to understand the relationship between the inputs and the outputs from a human perspective. This is problematic particularly in risk-sensitive disciplines such as medicine, where a transparent and verifiable decision-making process is crucial \cite{Xiao2018}. To address this lack of transparency, different methodological approaches have been proposed to understand the behaviour of machine learning models in general (e.g., LIME \citep{ribeiro2016should}, SHAP \citep{lundberg2017shap}) and neural networks in particular \citep{Koh2017, dosovitskiy2016generating, hinton2015distilling, chen2018learning}.

Various attribution methods have been introduced to explain neural network predictions in image classification \citep{ancona2018unifiedattribution}, which attempt to determine the attribution or relevance of each input feature to the predicted output of a neural network. The resulting attributions are then displayed in a heatmap to visualize which input features have positively or negatively influenced the prediction.
The proposed attribution methods can be categorized into two different methodological approaches: perturbation-based methods and backpropagation-based methods.
Perturbation-based methods assign attribution values to input features directly by modifying the input space through removing, masking, or altering features and measuring the difference between the predictions based on the modified and the original inputs. An example is the occlusion method proposed by \citet{Zeiler2014}, which systematically masks a region in the input image to observe potential changes in the target class probability. \citet{Rieke2018} extended this method to an atlas-based occlusion for MR images. Backpropagation-based methods, on the other hand, assign the attribution for all input features using the backpropagation algorithm to either compute partial derivatives of the outputs with respect to the input features \citep{Simonyan2013,sundararajan2017integratedgradients} or to backpropagate a relevance score through the network \citep{shrikumar2017deeplift}. Although the visualizations do not directly relate to output variations, these approaches are computational less expensive compared to perturbation-based methods and thus make them more suitable for high-dimensional MR images \cite{Boehle2019}. A popular method of this kind is layer-wise relevance propagation (LRP; \citep{Bach2015}), in which a relevance score is introduced to the output layer, which is then then propagated back to the input layer using a modified backpropagation algorithm.

\section{Methodological promises}\label{sec:meth_prom}

Deep neural networks present advantages over classical statistics and machine learning methods that are particularly promising for neuroimaging data in psychiatry. In this section, we discuss how deep neural networks can provide meaningful intermediate representations, which facilitate discovery of disease subtypes. Furthermore, we show how these intermediate representations enable transfer learning to deal with comparably small sample sizes in neuroimaging. Lastly, we  discuss how specialized architectures (i.e., inductive biases) can more efficiently exploit the structure in neuroimaging data. These three advantages lay the methodological foundation for the application scenarios in section \ref{S:3}. 

\subsection{Representation learning}\label{sec:repr} 
\begin{figure}[ht!]
\def\svgwidth{\linewidth}
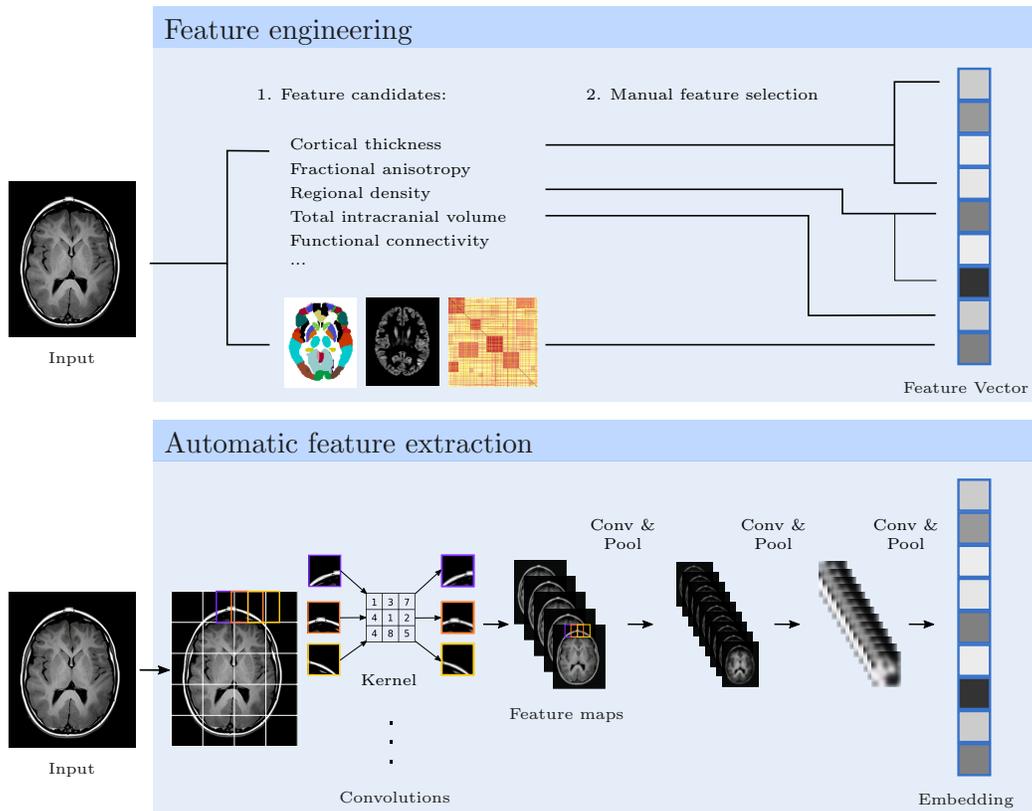
\caption{Deep neural networks have the ability to automatically perform feature extraction (bottom). Automatic feature extraction allows the automatic design and selection (``learning") of representative features. In contrast (top), traditional machine learning models mostly require manual extraction and selection of features. These features are typically designed (``engineered") by human experts and therefore introduce a bias about the experts' knowledge into the features. Furthermore, these methods require additional tools (e.g., Statistical Parametric Mapping (SPM) or Freesurfer) 
 or manual effort (e.g., manual segmentation).} 
\label{fig:feature_engineering}%
\end{figure}

A central advantage of deep neural networks is their ability to perform automatic feature engineering on raw or minimally processed data (e.g., spatially normalized MRI data) as shown in Figure \ref{fig:feature_engineering}. These automatically learned features can be interpreted as a new view or representation of the input data and can be extracted by reading out the activations of neurons in a given layer \citep{bengio2013representation}. Thus, any given data point (e.g., an MR volume of a particular subject) can be described by the activation profile that it generates in a particular layer of a deep neural network. 

In analogy to how a CNN trained to classify natural images will yield successive layers of intermediate representations detecting edges, textures, and, finally, whole objects \citep{lee2009convolutional}, one expects a deep neural network trained on neuroimaging data to provide a  hierarchy of intermediate representations reflecting brain structure or function (e.g., low-level neural structures such as boundaries between grey and white matter, focal abnormalities such as lesions or atrophy, and disease profiles \citep{plis2014deep}).

To create representations that are useful for a specific research question, there exist two general approaches: supervised and unsupervised representation learning. In supervised representation learning, models are trained to predict a target variable, and successive layers of the model discard more and more prediction-irrelevant information  while retaining and recombining prediction-relevant information into increasingly abstract representations of the input data (for a formal discussion, see e.g., \citet{tishby2015deep}). 
In contrast, unsupervised representation learning operates exclusively on the input data. One example are autoencoders that transform the data to pass through a low-dimensional bottleneck and then try to reconstruct the original data from the compressed intermediate representation \citep{kramer1991nonlinear, Goodfellow2016}. This approach implicitly assumes that a compressed representation of the data is likely to mirror abstract higher-order concepts or latent variables that generated the data. Recently popularized ``self-supervised" learning extends this approach to other auxiliary tasks for learning meaningful representations, such as predicting the next word in a sentence  \citep{mikolov2013efficient}, color channels from a grayscale image \citep{zhang2017split}, or  anatomical segmentations from an MR image \citep{balakrishnan2019voxelmorph}.
Intermediate representations in deep neural networks can be constrained to have potentially useful mathematical properties such as normality and independence \citep{kingma2013auto}, or to selectively disentangle factors of variation  \citep{cheung2014discovering}. They are often designed to reduce the dimensionality of the input data (e.g., high-dimensional neuroimaging data) for further analyses, such as clustering (section \ref{sec:subtype}) or to test scientific hypotheses (section \ref{sec:hypothesis}).

\subsection{Transfer learning} 

In many domains, for instance in computer vision, there exist large general purpose data sets (e.g., ImageNet) while data sets for specific applications are often prohibitively small. Recently, a similar situation has arisen in neuroimaging, where small clinical studies are now augmented by large-scale data collection initiatives such as the UK Biobank or ENIGMA. The existence of intermediate representations in deep neural networks can allow for carrying insights gained in large general purpose data sets over to small sample clinical settings \citep{he2020meta, semi2017}. Models can be split at the level of an intermediate representation, allowing to \emph{pre-train} the lower layers of the networks on a large data set, and then \emph{fine-tune} the higher layers on the target data set. This approach is called transfer learning \citep{tan2018survey}.

The motivation for transfer learning in CNNs is simple. Instead of initializing weights randomly, they are initialized based on a related data set. Ideally, the data used for pre-training the weights should have common properties with the target data, such as objects with high frequency edges. If the low-level representations of a CNN are responsible for detecting those edges, then the learned properties of edge detection can be transferred to the new task. The transferred weights can then either be frozen i.e. used as a fixed feature extractor in combination with a classifier (e.g. fully-connected layers), or can be fine-tuned to adapt to the target domain. This setup can permit the application of expressive deep learning models even in situations where the sample size of the target data set would be insufficient for training such a model from scratch.

For analyzing MRI data, transfer learning from natural images \citep{maqsood2019transfer, hon2017transfer} and from related MRI data (e.g., other MRI sequences; \citep{Ghafoorian2017TransferLearning, Eitel2019MS}) has been suggested.
Even though, the number of studies on transfer learning in neuroimaging is growing, studying, for instance, brain lesion segmentation \citep{kamnitsas2017unsupervised, akkus2017deep,Ghafoorian2017TransferLearning,valindria2017reverse} or Alzheimer’s disease classification \citep{Gupta2013, hosseini2018alzheimer, Boehle2019}, the most effective approach for transfer learning in clinical neuroimaging has not yet been established. 

\subsection{Inductive bias}\label{sec:inducbias} 

Every machine learning algorithm must make certain assumptions in how it processes the training data. For instance, ($l_1$-regularized) linear regression assumes a (sparse) linear relationship between input and target variable. The sum of assumptions of a given learning algorithm is called its inductive bias. Inductive biases that reflect the data-generating process or restrict the space of potential solutions to those that are a priori meaningful, can substantially reduce the number of training samples needed to achieve the prediction goal \citep{geman1992neural}. 
The flexible architecture of deep neural networks allows for the creation of complex inductive biases \citep{battaglia2018relational}. The success and subsequent (re-)popularisation of deep learning in the last two decades has been attributed mainly to advances in CNNs, i.e. the development of an inductive bias that allows efficient exploitation of translation invariance in natural images. Other examples include recurrent neural networks which assume time invariance \citep{yu2019review}, or graph neural networks which assume invariance to ordering (except for pair relations) \citep{wu2020comprehensive}.   

In contrast to natural images, MRI brain images do not possess strong translation invariance properties. Due to a standard brain structure (e.g., hippocampus is at the same location in all humans) and further registration to a common template (e.g., MNI), the usefulness of translation invariance as an inductive bias for neuroimaging data has been questioned \citep{Schulz2019dlbrains}.
Recently, specialized architectures of deep neural networks have been proposed that are specifically adapted to neuroimaging data. \citet{kawahara2017brainnetcnn} proposed an architecture designed for connectivity matrices. \citet{eitel2020harnessing} restrict the translation invariance of CNNs to local patches, accounting for spatially normalized MRI data. Graph CNNs \citep{defferrard2016convolutional,bronstein2017geometric, kipf2016semi} have been adapted to neuroimaging data by e.g., Parisot et al. \citep{parisot2017spectral, parisot2018disease}. Further advances in inductive biases for neuroimaging data may be an important prerequisite for fully utilizing deep neural networks in psychiatric research. 


\section{Application scenarios}\label{S:3}
%

\begin{figure}[ht!]
\def\svgwidth{\linewidth}
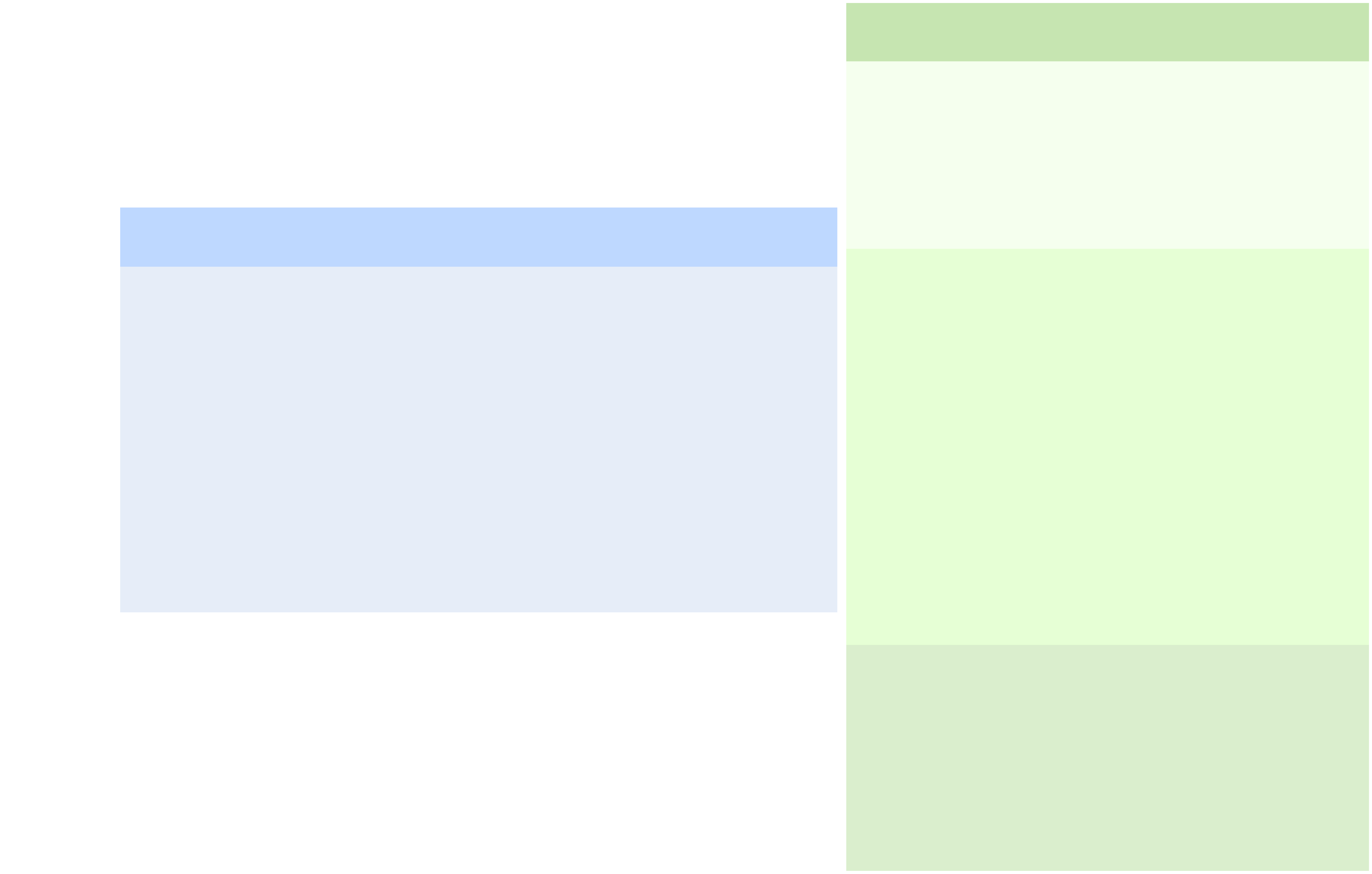
\caption{Three application scenarios for deep learning in neuroimaging-based psychiatric research. The input is processed by a convolutional feature extractor (1.) and then further analyzed (2.) by one of the three described scenarios: (a) automatic disease diagnosis and prognosis, (b) subtype discovery and normative modeling, and (c) hypothesis generation and biomarker identification.}\label{fig:overview}
\end{figure}

Motivated by the ability of CNNs to extract hierarchical representations of complex image data, and several groundbreaking results for medical imaging applications where human-level performance has been surpassed \citep{esteva2017dermatologist, gulshan2016development}, we have selected three broad use cases which we determined as specifically promising for neuroimaging-based psychiatric research. The following section outlines these use cases (see also Figure \ref{fig:overview}), namely 
the automatic diagnosis of diseases and prognosis, the identification of disease subtypes and modeling normative distributions for detecting aberrations, and how deep learning can be used to develop new neuroscientific hypotheses.

\subsection{Automatic disease diagnosis and prognosis} 
The most apparent tasks for deep learning in neuroimaging-based psychiatric research are automatic disease diagnosis, prognosis, and prediction of treatment response. While signs of psychiatric disorders in neuroimaging data are very difficult to detect for human experts, several studies have suggested that there exist subtle, but measurable alterations in brain MRI data \citep{lui2016psychoradiology}. Hence, computer-assisted diagnosis using machine learning models that can pick up even very small and diffuse signals become highly promising.
Most applications so far focused on binary classification tasks, such as patient vs. control \citep{Vieira2017UsingApplications, wen2020convolutional}, converter vs. non-converter \citep{jo2019deep}, and response vs. non-response to treatment \citep{patel2015responsepred}, but the extension to multi-class classification tasks, such as differential diagnosis \citep{schnack2014can, talpalaru2019identifying, liu2015multilabelmulticlass}, is straightforward.


Since deep neural networks specifically profit from data which contains objects that are composed of a hierarchy of abstractions, deep neural networks are a particularly suitable method for sMRI data; the brain is composed of four main lobes, which are each composed of several brain regions and each regions is composed of edges and structures. The complexity of the neurobiology of psychiatric disorders makes nonlinear methods such as deep learning more likely to successfully detect subtle disease-related alterations. Since the aim is to predict an outcome for each subject individually, rather than group level associations, machine learning in general may facilitate personalized medicine \citep{bzdok2020prediction, walter2019translational}. Crucially, transfer learning of deep neural networks may help circumvent the problem of small sample sizes in psychiatric studies.  Furthermore, automatic disease diagnosis is less prone to human bias \citep{saposnik2016cognitive, hamberg2008genderbias}. This is especially true for deep learning as it does not require manual feature extraction which can introduce specific biases such as the importance of some regions over others. 

Beyond end-to-end learning, in which the model is fed with imaging data and outputs a class label, more indirect approaches exist. Here, the model output can include segmentation of white matter hyperintensities \citep{habes2016white} or grey matter segmentation, which can form the basis for atrophy measurements \citep{PAGNOZZI2019116018}. These outputs can either be used to support a human expert's decision or to extract latent features for training another classifier \citep{Suk2014}. 

When using longitudinal data, one can build deep learning models for prognosis, which have shown to outperform other methods in modeling disease progression \citep{alaa2019attentive}. From a clinical perspective, it is highly relevant to be able to assess time-to-event questions such as treatment outcome, survival prediction, relapse probability or potential disease onset. Current data sets are slowly approaching sample sizes that may allow for answering those questions. With future large-scale data sets, new milestones for clinical neuroimaging might be reached \citep{thompson2020enigma}.

\subsection{Subtype discovery and normative modeling}\label{sec:subtype}

Traditional diagnostic categories rely on identifying groups of frequently co-occurring symptoms, which are assumed to represent coherent disease entities. 
While successful in many areas of medicine, this approach tends to fail in psychiatry, where symptoms are far removed from the underlying pathomechanisms.
Mounting evidence suggests that established disease categories (such as ICD-10 or DSM-V \citep{lehman2000diagnostic}) insufficiently match actual biological dysfunction \citep{Karalunas2014,tamminga2014bipolar,lowe2008depression, Cuthbert2013, Insel2010, Insel2015}. 
What appears as a coherent disease entity often turns out to be a mixture of distinct subtypes with heterogeneous biomarkers, treatment responses, and disease progressions. Consequently, robust biomarkers that allow for unambiguous diagnosis and predicting treatment response, common in other areas of medicine, remain elusive \citep{Insel2015}.

Increasingly available large-scale neuroimaging data sets allow for a new approach to disease categorization. Instead of using clusters of often qualitative symptoms as reference points in the search for the biological causes of psychiatric disease, it becomes possible to identify clusters directly from the quantitative neuroimaging data. This data-driven approach to disease categorization has multiple benefits. First, it prioritizes biomarkers. The whole process centers on quantitative neuroimaging data for identifying clusters of dysfunction and thus directly yields historically elusive neuroimaging biomarkers for psychiatric diseases. Second, by considering data that is closer to the level of biological dysfunction than self-reported or clinician-observed symptoms, disease categories may better match the underlying pathomechanism. Third, and consequently, such biomarkers and disease categories should better predict disease progression and treatment response \citep{Insel2015}.

The data-driven search for biologically meaningful disease categories, or disease subtypes (biotypes), relies on cluster analysis \citep{hastie2009elements, karim2020deep, mirzaei2018segmentation, schulz2020inferring} -- a class of machine learning algorithms that automatically subdivide data samples into distinct groups, called clusters, such that data points in the same group are maximally similar while data points assigned to different groups are maximally dissimilar.
The question now becomes how to define similarity. Similarity in the space of voxels of a brain MRI is unlikely to yield clusters that are relevant for psychiatry for two reasons. First, a brain MRI is extremely high-dimensional, so that the variables, i.e. voxels, vastly outnumber the available data samples. This leads to the so-called curse of dimensionality; the higher the dimensionality, the more ways there are to be dissimilar and distance (or similarity) functions become increasingly meaningless.
Second, similarity in the voxel-space of a brain MRI will be dominated by strong dimensions of variation such as sex and age, which tend to drown out the comparatively small dimensions of clinical relevance \citep{panta2016tool}. 
For neuroimaging-based clustering to be useful for psychiatry, the raw voxels of the brain MRI need to be transformed into a more meaningful representation which alleviates the curse of dimensionality and emphasizes clinically relevant variation over irrelevant confounders.
Deep neural networks are particularly well suited to provide these highly abstract representations of neuroimaging data which are crucial for data-driven disease subtyping (see section \ref{sec:repr}). 

Both promises and challenges of neuroimaging-driven subtyping can be illustrated by a study on depression subtypes by \citet{Drysdale2017Resting-stateDepression}. 
These authors used canonical correlation analysis (CCA) \citep{hardoon2004canonical} to generate an intermediate representation, linking resting-state fMRI functional connectivity to the Hamilton Depression Rating. The low-dimensional intermediate representation emphasized specifically those aspects of the fMRI functional connectivity that pertain to the patients' self-reported depression symptoms. Cluster analysis on the intermediate representation identified four subtypes of depression, each with distinct connectivity and clinical symptom profiles, and which were predictive of patients' responses to transcranial magnetic stimulation.
In a replication study, however, \citet{dinga2019evaluating} were unable to reproduce these results on independent data and emphasize that ``[clustering algorithms] always yield clusters, regardless of the structure of the data", showing that comprehensive statistical analysis is necessary to ensure robust findings. 
The CCA that was employed by \citet{Drysdale2017Resting-stateDepression} can be seen as a linear precursor to more expressive models based on deep neural networks. Deep neural network extensions of CCA \citep{andrew2013deep} and deep neural network architectures that are more finely tuned to the structure of neuroimaging data \citep{kawahara2017brainnetcnn, eitel2020harnessing, parisot2018disease} may in the future unlock more intricate nonlinear interactions in the data, which are inaccessible to classical models like CCA.


 A related approach to utilizing intermediate (latent) representations of neuroimaging data for psychiatry is normative modeling \citep{marquand2019conceptualizing, Marquand2016}. Instead of training models on data from diseased subjects to delineate disease profiles, normative modeling relies on data from healthy subjects to model the underlying dimensions of normal variation in neuroimaging data. The high-dimensional neuroimaging data is reduced to a low-dimensional representation, either via traditional machine learning models (see \cite{marquand2019conceptualizing} for an overview), or more recently, by extracting the intermediate representations of deep neural networks (see section \ref{sec:repr};  \citep{pinaya2019using,kia2019neural, pinaya2020normative}).  In the space spanned by the new low-dimensional representation,  the distribution of healthy subjects can be easily characterized, thus defining a normal range of variation. Subjects that fall outside of this normal range can be considered to have abnormal brain structure or function. There are different ways in which subjects can deviate from the normal range. Subjects may be normal in most dimensions of the model's intermediate representation and deviate in one or two specific dimensions, others may be outliers in a whole range of dimensions. Investigating these anomaly profiles could serve as an approach to disease classification in its own right \citep{wolfers2018mapping}.


\subsection{Hypothesis generation and biomarker identification using deep learning}\label{sec:hypothesis}

In contrast to training disease models on a priori defined features (e.g., hippocampal volume in Alzheimer's disease \citep{ACHTERBERG2019hippocampal}), deep learning allows for extracting its own representations (see section \ref{sec:repr}) which can be used to generate new scientific hypotheses and indicate directions for further study.
One approach is to extract the learned representation of a CNN into a comprehensible format, such as a voxel-wise relevance heatmap (see section \ref{sec:attribution}), in order to detect patterns across subjects. For instance, multiple studies proposed different ways to compute relevance heatmaps in Alzheimer's disease for explaining individual network decisions \citep{Rieke2018, Boehle2019, Dyrba2020vis, Jo2020taubetvis}. In accordance with neurobiological evidence, all studies found the hippocampus and other temporal regions as primarily important. On the one hand, relevance scores can be used for model validation and the detection of data set bias when comparing regions with neurobiological evidence. On the other hand, new disease discriminating associations between certain brain regions and psychiatric disorders could be identified. These associations could also include common patterns of relationships between different brain regions, which only in combination elucidate predictiveness. This similarly applies to multi-modal studies which integrate multiple imaging modalities or combine them with clinical and socio-demographic features \citep{whelan2014neuropsychosocial, pettersson2013multimodal, nie2016multimodal, patel2015responsepred, maglanoc2020multimodal}.  Here, more complex relationships between various variables might be detected, such as the covariance of a brain region with the clinical history of a patient. In addition, biomarkers which most human experts would only identify in a specialized neuroimaging modality (e.g., diffusion tensor imaging) might have traces in a more common modality (e.g., T1-weighted MPRAGE). Those traces could be learned in a multi-modal fashion and then be used in cases where only the latter modality is available. It is essential that new hypothesis and biomarkers are tested on independent data, in order to avoid the so called ``double dipping" in which one data set is used in circular analysis \citep{Kriegeskorte2009}.


\section{Applications in psychiatric research}\label{sec:applications} 
In the last 1-2 decades, a number of neuroimaging-based machine learning studies in psychiatric research have been performed (for reviews, see \citep{Wolfers2015, Woo2017BuildingNeuroimaging,walter2019translational, dwyer2018machine, kloppel2012diagnostic}). Regardless of the difficulty of the application scenario, most of the initial studies were quite promising with accuracies above chance level for applications ranging from diagnosing patients with mental illness to identifying patients at risk and clinical subgroups to predicting treatment outcome \citep{koutsouleris2018prediction,koutsouleris2009use, Kloppel2008a, Weygandt2012}. Although the data sets and methods used were quite diverse and difficult to benchmark, usually classical machine learning analyses consisting of some kind of feature extraction (e.g., grey matter density, volumes of brain structures, cortical thickness or functional connectivity) in combination with a linear or nonlinear machine learning algorithm (e.g., SVMs, logistic regression or random forest) have been employed \citep{Woo2017BuildingNeuroimaging}. However, most of the studies relied on relatively small sample sizes ($N < 100$) and internal testing which often lead to overly optimistic performance metrics \citep{VAROQUAUX2018samplesizes, arbabshirani2017single, Woo2017BuildingNeuroimaging, flint2019systematic} in addition to publication bias towards positive findings \citep{koppe2020deep}. Even though deep learning studies within neuroimaging-based psychiatric research are still rare (with the exception of Alzheimer's disease), we review its application on several mental disorders, which have been employed in at least one of the applied machine learning settings described in section \ref{S:3}. 


\subsection{Alzheimer's disease}

Most MRI-based machine and deep learning studies so far have been conducted on Alzheimer's disease (AD), a neuropsychiatric disorder which is the main cause of dementia in the elderly. AD is symptomatically characterized by loss of memory and other cognitive abilities to such an extent that it affects daily life \citep{Bhogal2013}. In contrast to other psychiatric disorders, the neurobiological correlates are rather clear (i.e., neurodegeneration and resulting atrophy starting in the hippocampus). In addition, the existence of the Alzheimer's Disease Neuroimaging Initiative (ADNI) database, a large open data base with easy access for (non-clinical) researchers, considerably boosted the number of machine learning studies \citep{Weiner2013, Woo2017BuildingNeuroimaging}. 
In a recent review of deep learning studies in AD, accuracies of up to 98\% have been reported for the discrimination between patients with AD and healthy controls, and 83\% for the conversion from mild-cognitive impairment (MCI) to AD \citep{jo2019deep}. 
However, \citet{wen2020convolutional} pointed out that a considerable number of those studies suffer from data leakage (see section \ref{sec:val_metrics}) and therefore overestimate the prediction performance. 
From the 16 studies where data leakage could be excluded, accuracies between 76\% and 91\% for the discrimination of patients with AD and healthy controls and 65\% to 83\% for the discrimination between MCI and healthy controls have been reported. Whereas early studies mostly analyzed 2-dimensional slices of MRI data and have drawn on established architectures and pre-training techniques in computer vision (e.g., \citep{Gupta2013, Payan2015}), most studies now analyze 3-dimensional data in either a ROI-based (e.g., hippocampus) or a whole-brain setting \citep{Boehle2019, korolev2017residual, wen2020convolutional}. In \citet{wen2020convolutional}, those different approaches and additional settings regarding pre-processing and pre-training have been benchmarked with accuracies of up to 88\%. Deep learning approaches based on 3-dimensional data were superior to 2-dimensional approaches but not better than a standard machine learning analysis relying on SVMs. The models also generalized well between different data sets when the inclusion criteria and the underlying distributions of demographics was not too different. Only few studies so far have tried to understand and explain what the deep neural networks (i.e., CNNs) have learned during training (see section \ref{sec:attribution}). Different attribution methods, indicating the importance of each voxel for the final classification decision, have been compared \citep{Rieke2018} and especially layer-wise relevance propagation (LRP) has been shown to explain CNN decisions well in AD classification \citep{Bach2015, Boehle2019}. In \citet{wood2019neuro}, a recurrent visual attention model is suggested that processes the data iteratively and thus is easier to interpret than CNN-based approaches.

\subsection{Schizophrenia}
Schizophrenia is a mental disorder characterized by episodes of psychosis for which 
neurobiological abnormalities in sMRI and fMRI data with a focus on enlarged ventricles and an involvement of structures within temporal and frontal lobe have been reported \citep{shenton2001review,Narr2004, van2010self, kircher2019neurobiology}. Based on diverse features ranging from grey matter density to volumes to functional correlates, a number of classical machine learning studies have been performed to either diagnose schizophrenia, predict first-episode psychosis, or identify subtypes \citep{koutsouleris2009use, Davatzikos2009,greenstein2012ROIschizophrenia, gould2014multivariate, du2018classification, talpalaru2019identifying, vieira2020using} (for a recent review, see \citep{de2019machine}). The classification accuracies varied here considerably between 60\% and 95\%; a tendency of lower accuracies for larger data sets. Independent testing suggested that reported accuracies should be taken with caution \citep{Woo2017BuildingNeuroimaging, arbabshirani2017single}. 
Regarding deep learning, a first promising study using a deep belief network has shown that more depth resulted in a strong increase of accuracy (from around 60\% to 90\%) for separating patients with schizophrenia and healthy controls \citep{plis2014deep}. 
Another study using a deep belief network has shown a slightly higher accuracy than for a SVM (74\% to 68\%) \citep{pinaya2016using}. 
Using CNNs and MRI data from patients with schizophrenia and healthy controls from 5 public data sets ($N=873$), \citet{oh2020identifying} report AUCs of up to 0.90 on the left-out data set with a high relevance of the right temporal region. 
However, the importance of a similar training and test distribution was shown by a rather low AUC of 0.71 for an external data set with younger patients and a shorter illness duration. 
Notably, for clinicians (5 psychiatrists, 2 radiologists) the average accuracy was 62\% on 100 randomly selected MRI volumes. 

For resting-state and task-based fMRI data, diverse techniques have been employed to discriminate between patients with schizophrenia and healthy controls including different kinds of
autoencoders \citep{kim2016deep, zeng2018multi, li2020application}, deep generative models \citep{matsubara2019deep}, and CNNs \citep{qureshi20193d, oh2019classification}. 
%
For functional connectivity matrices from a large multi-site sample ($N=734$), accuracies between 81\% and 85\% for leave-one-site-out classification have been reported  \citep{zeng2018multi}. 
In accordance with neurobiological findings, most important for the classification were connectivity values within the cortical-striatal-cerebellar circuit. Compared to whole brain images and graph-based metrics obtained from resting-state fMRI, functional connectivity measures have generally been shown to be more informative in a machine and deep learning framework \citep{lei2019detecting}. In \citet{plis2018reading}, a deep learning-based translation approach is suggested in order to investigate the linkage between MRI-based structure (grey matter) and function (dynamic connectivity). 
For task-based fMRI data, similar accuracies (around 80\% to 84\%) have been reported \citep{li2020application, oh2019classification}, relying on, for example, a sensorimotor task \citep{li2020application}. 
To exploit the temporal structure of fMRI data, \citet{yan2019discriminating} developed a multi-scale recurrent neural network and report accuracies between 80\% and 83\% in a multi-site setting ($N=1100$).
In \citet{pinaya2019using}, a normative model based on a deep autoencoder was trained in a large cohort of healthy controls ($N=1113$) and then the reconstruction error as a measure of deviation was assessed in
patients with schizophrenia.


\subsection{Internalizing disorders}
Internalizing disorders describe a group of psychiatric disorders which are characterized by negative affectivity and include, for instance, depression, anxiety disorders, and post-traumatic stress disorder (PTSD). In contrast to AD and schizophrenia, internalizing disorders have been comparably less studied using classical machine learning and deep learning.
On the one hand, as for most psychiatric disorders, the labels per se have been criticized for being mainly based on symptoms rather than neurobiological correlates and thus being too noisy and unspecific for the use in supervised machine learning studies \citep{rokham2020addressing}. On the other hand, neurobiological correlates obtained from neuroimaging data are less clear for internalizing disorders and might be mediated by underlying subtypes \citep{ionescu2013neurobiology, Drysdale2017Resting-stateDepression}.

For major depressive disorder (MDD), a recent review \citep{gao2018machine} summarized the results of 66 classical machine learning studies equally distributed on structural MRI, resting-state and task-based fMRI. In comparison of modalities, models using resting-state fMRI data tended to yield higher accuracies. Some studies also investigated machine learning algorithms in the context of predicting treatment outcome with a focus on fMRI connectivy as a biomarker. The sample size of studies were mostly below 200 and accuracies for discriminating MDD patients and healthy controls ranged from chance (approx. 50\%) to unrealistic 100\%, again showing the need for caution when interpreting results. Problems such as data-leakage or insufficiently sized test sets (see section \ref{sec:label}) have most likely afflicted at least some of these studies \citep{Drysdale2017Resting-stateDepression, jiang2018smri, gao2018machine}. 
For bipolar disorder, a large recent multi-site study \citep{nunes2018using} based on 13 cohorts from ENIGMA exists ($N = 3020$) that report aggregated subject-level accuracies of about 65\% using a combination of SVMs and extracted MRI features (regional cortical thickness, surface area, and subcortical volumes). 

Using CNNs and recurrent neural networks, \citet{pominova2018voxelwise} report  AUCs of up to 0.66 for sMRI and 0.73 for fMRI for separating patients with MDD and healthy controls. In a large-scale machine learning challenge (PAC 2018\footnote{\url{https://www.photon-ai.com/pac}}
, $N=2240$) a variety of simple and high-complex classifiers (CNNs etc.) were benchmarked by 49 teams for the classification between patients with MDD and healthy controls. However, none of the teams achieved an accuracy higher than 65\% on the holdout set. Given that a number of studies have reported classification accuracies of over 90\% for the classification of depression based on sMRI data (for an overview, see \citet{Wolfers2015,gao2018machine}), those results seem rather surprising on a large well-controlled data set. A possible conclusion from these results is that PAC showed more realistic accuracies than published studies that suffer from overconfidence as previously explained. Based on the same data set, it has been shown that accuracies are systematically overestimated in random subsets \citep{flint2019systematic} replicating small sample size effects, e.g., reported in \citep{Woo2017BuildingNeuroimaging,VAROQUAUX2018samplesizes,  arbabshirani2017single}. In contrast to decoding clinical categories, \citet{pervaiz2020optimising} suggests to predict neurotisicsm as an underlying construct for the potential incidence of mood disorders. However, based on functional connectivity data from the UK biobank ($N\approx 14000$) and diverse optimised pipelines, the average correlation between predicted and true neuroticsm score was below 0.2.

Regarding anxiety disorders including obsessive-compulsive disorder (OCD) and PTSD, a review from 2015 included 8 studies which all had a rather low sample size and report comparable high accuracies \citep{Wolfers2015}. Additional studies investigated the prediction of treatment outcome in anxiety disorders (for a review, see \citet{lueken2016neurobiological}). For OCD, a systematic review found 12 studies with a wide range of accuracies 
based on different modalities but sample sizes were rather small \citep{bruin2019diagnostic}. None of those studies used deep learning.

 
\subsection{Neurodevelopmental disorders} 
Neurodevelopmental disorders describe a group of neurological and psychiatric conditions that originate in childhood. We will focus here on two diseases, namely attention-deficit/hyperactivity disorder (ADHD) and autism spectrum disorder (ASD), which both have been investigated in deep learning frameworks \citep{Vieira2017UsingApplications, khodatars2020deep}. 

ADHD is characterized by an ongoing pattern of inattention, hyperactivity and/or impulsivity, which has been related to smaller brain volumes (e.g., in the dorsolateral prefrontal cortex) and altered functional connectivity \citep{seidman2005structural, yu2007altered}. In a competition from 2012 based on the ADHD-200 data set\footnote{\url{http://fcon_1000.projects.nitrc.org/indi/adhd200/}}, the best predictive model relied only on personal characteristic data including age, gender, and several IQ scales, and led to an accuracy of 62.5\%.\footnote{\url{http://fcon_1000.projects.nitrc.org/indi/adhd200/results.html}} The best image-based model resulted in an accuracy of 60.5\%. In an extended ADHD-200 data set, it has been shown that models based on only personal characteristic data outperform models based on resting-state fMRI data ($75.0$
to $70.7$\%) \citep{brown2012adhd}. However, since then a number of deep learning studies based on resting-state fMRI data have been performed and accuracies of up to 90\% have been reported \citep{kuang2014discrimination, deshpande2015fully, riaz2020deepfmri} (for an overview, see \citep{Vieira2017UsingApplications}). Recently, also based on the ADHD-200 data set, a spatio-temporal model and a multi-channel model for combining resting-state fMRI data as well as demographics have been shown to result in AUCs between 0.74 and 0.8 \citep{mao2019spatio, chen2019multichannel}. 

Subjects with ASD are mainly characterized by repetitive behavior and difficulties in social interactions. A number of structural and functional abnormalities have been described including a slightly thinner temporal cortex, a thicker frontal cortex, reduced volumes of amygdala and nucleus accumbens, as well as a reduced functional connectivity \citep{van2018cortical, kennedy2008intrinsic}. An overview over deep learning studies in ASD can be found in \citet{khodatars2020deep}. Based on sMRI and resting-state fMRI from the Autism Brain Imaging Data Exchange (ABIDE) initiative, accuracies of about 70\% have been reported \citep{heinsfeld2018identification, khodatars2020deep}.



\subsection{Substance abuse}

Substance use disorders (SUDs) describe a class of mental disorders related to problematic consumption of alcohol, tobacco and illicit drug use  
affecting daily and working life. 
For alcohol use disorder (AUD), the most prominent SUD, sMRI and fMRI studies have revealed a number of neurobiological correlates including enlarged ventricles, grey and white matter loss in frontal and reward-related brain areas, as well as altered functional connectivity in the amygdala and nucleus accumbens 
\citep{buhler2011alcohol, zahr2017alcohol, chanraud2007brain, hu2018resting, wang2018disrupted}.
Classical machine learning models have been employed to identify AUD or predict alcohol consumption / binge drinking on different kinds of data including demographics, history of life events, personality traits, cognition, candidate genes, as well as brain structure and function \citep{guggenmos2018decoding, guggenmos2020multimodal, seo2015predicting, zhu2018random, whelan2014neuropsychosocial}. For sMRI data, it has been shown that a computer-based classification approach performed better than a blinded radiologist in diagnosing alcohol dependence based on regional grey matter (74\% to 66\%) and predicting future alcohol consumption \citep{guggenmos2018decoding}.
Compared to demographics, sMRI and task-based fMRI, \citet{fede2019resting} showed that resting-state connectivity resulted in the lowest root-mean-squared error (RMSE) for predicting alcohol severity measured by the Alcohol Use Disorders Identification Test (AUDIT; resting-state fMRI: 8.04, demographics: 9.76, sMRI: 8.11, task-based fMRI: 8.63). 
Alterations in the reward network and executive control network have been reported as informative for diagnosing AUD \citep{zhu2018random}. For predicting alcohol use during adolescence, thinner cortices and less brain activation in frontal and temporal areas have been found \citep{squeglia2017neural}.
The largest classical machine learning study so far has been performed on the IMAGEN data set, as part of which \citet{whelan2014neuropsychosocial} report an AUC of 0.90-0.96 for the separation of 14-year old binge drinkers and 14-year old controls and an AUC of 0.75 for predicting binge drinking at 16 years. 
  Here, a combination of history, personality traits, and brain features were most predictive, supporting the hypothesis that multiple causal factors shape later alcohol use. In another study, sex-specific psychosocial and neurobiological features have been identified for the initiation of cannabis use \citep{spechler2019initiation}.

To the best of our knowledge, only three recent studies (from a single group) exist that have used CNN models to identify AUD \citep{wang2018alcoholism,  wang2019alcoholism, wang2020alcoholism}. They applied different CNN models 
to 2-dimensional slices and reported accuracies around 97\% in discriminating patients with abstinent long-term chronic AUD ($N = 188$) and healthy controls ($N = 191$). Due to the excessive amounts of alcohol consumed, the CNN models presumably captured neurotoxic effects rather than neurobiological underpinnings explaining AUD.



\subsection{Brain age}
As an overarching biomarker relevant in many diseases including AD, schizophrenia, and substance abuse, brain age as opposed to chronological age has been suggested \citep{guggenmos2018decoding, elliott2019brain, bashyam2020mri, cole2018brain}. Using CNNs, highly successful and robust models for brain age have been developed on raw MRI data \citep{cole2017predicting} and parcellations \citep{jiang2019predicting} with average deviations (mean absolute error) of 4-6 years. However, similar deviations were achieved using Gaussian processes applied to grey matter volumetric maps \citep{cole2017predicting}. Based on a large and heterogeneous multi-site cohort ($N=11729$), \citet{bashyam2020mri} have shown that moderately fitting brain aging models (based on 2-dimensional CNNs) might lead to a better separation between patients (with AD, MCI, schizophrenia or depression) and healthy controls than tightly-fitting models. 

\section{Challenges}\label{sec:challenges} 

Besides the potential of deep learning in neuroimaging-based psychiatric research outlined in sections \ref{sec:meth_prom} and \ref{S:3} and early promising studies presented in section \ref{sec:applications}, the field faces some substantial challenges hindering immediate application in clinical routine. 

Noise in MRI data and disease labels affects prediction performance and increases the number of data samples required for training deep neural networks  (section \ref{sec:noise_both}). In section \ref{sec:models}, we discuss difficulties of model choice and model comparisons within the neuroimaging domain. For instance, it is still contentious, under which circumstances deep neural networks are useful for neuroimaging data and whether neuroimaging data require nonlinear models (section \ref{sec:nonlin}). And even within different neural network architectures, comparing model performances is impeded by a lack of neuroimaging benchmark data sets (section \ref{sec:bench}). Furthermore, the computational expense of 3-dimensional CNNs hinders widespread application (section \ref{sec:compu}).

Even when a deep neural network achieves high prediction performance on an independent test set, pitfalls with respect to validity and explainability remain (section \ref{sec:valid}). Imbalances in the training data can lead to systematically misdiagnosing minorities (\ref{sec:ab}). Correction for confounding variables becomes substantially more challenging than in classical statistics (\ref{sec:cm}). In addition, complex machine learning models are hard to interpret and require sophisticated engineering to extract explanations for a models' predictions (\ref{sec:blackbox}).


\subsection{Noise in MRI data}\label{sec:noise_both}

\subsubsection{Low signal-to-noise ratio}\label{sec:noise}

The most basic precondition for machine learning to succeed is the existence of exploitable mutual information between brain images and target variables. While this precondition is surely fulfilled for sex or age, brain lesions or atrophy, the situation is less clear for more intricate variables representing aspects of human cognition and affectivity which are relevant in psychiatry \citep{uttal2011mind, Woo2017BuildingNeuroimaging}. It is, for instance, controversial to what extent subtle traits such as intelligence are reflected in brain morphometry measured via currently available sMRI technology \citep{mihalik2019abcd, hilger2020predicting}. Similarly, fMRI does not directly reflect neuronal activity and instead relies on the haemodynamic response as a proxy, giving it a temporal and spatial resolution that is far removed from the actual underlying phenomenon \citep{Aine1995, Woo2017BuildingNeuroimaging}. The actual signal, for example the cognitive process in question, will realistically only occupy a small fraction of a diverse conglomerate of different signals \citep{bellon1986mr, liu2016noise}. Unrelated anatomical differences, differences in physiology, unrelated physiological processes such as breathing or head movements, thermal and system noise from the scanner, even effects from unrelated cognitive processes may obscure a particular trait or cognitive process to the point of invisibility.
In comparison to areas where deep learning is highly effective, such as natural language, natural images, and even some areas of biomedicine (e.g., histology \citep{kather2019deep}, or dermatology \citep{esteva2017dermatologist}), neuroimaging data in the context of psychiatry contains high levels of noise and often little trace of the phenomenon under investigation.  Low signal-to-noise poses challenges to the application of deep learning, requiring a higher amount of training data (\ref{sec:label}) and may impede the extraction of nonlinear structure (\ref{sec:nonlin}).

\subsubsection{Lack of ground truth labels}\label{sec:label}

The success of supervised machine learning approaches in neuroimaging-based psychiatric research may be limited by insufficiently valid and reliable labels. Although psychiatric research is not imaginable without clinical labels such as depression or schizophrenia, those diagnostic categories have been severely criticized for not incorporating underlying neurobiological correlates and for their limited ability to account for heterogeneity as well as comorbidities within and across clinical categories \citep{Insel2010,Insel2015,Cuthbert2013,Karalunas2014}. Moreover, both low inter-rater and low test-retest reliability have been repeatedly documented \citep{specht2019current}. To address this issue, the National Institute of Mental Health (NIMH) suggested to study mental health along the so-called Research Domain Criteria (RDoC), and by this to shift focus away from classical clinical labels towards carefully designed domains of neurocognitive functioning across different levels of analysis assumed to be relevant to mental health \citep{Insel2010,Cuthbert2013}.

In machine learning terminology, the field of psychiatry would be described as suffering from high ``label noise". The higher the label noise, the more data samples are necessary to reliably characterize the statistical relationships between brain image and clinical label. This reduced sample efficiency, i.e. the reduced amount of information gained per sample, further exacerbates the problem of insufficient sample sizes in neuroimaging psychiatry, where generating additional training data is often prohibitively expensive.  Especially for small sample sizes, high label noise increases the likelihood of unrepresentative, spurious high-accuracy results \citep{VAROQUAUX2018samplesizes, arbabshirani2017single}. These can arise not only by overfitting or data leakage, but also out of sheer coincidence; the particular subsample of data used for model evaluation may randomly contain more correct or more easy-to-classify labels than what would be representative for the data set \citep{flint2019systematic}.   

\subsection{Model choice and model comparison}\label{sec:models}

\subsubsection{Linear vs. nonlinear models} \label{sec:nonlin} 

Even though deep learning has revolutionized the fields of computer vision and natural language processing, it is yet unclear if these successes will fully translate to neuroimaging. Results on deep learning based prediction of demographic or behavioral phenotypes have been mixed. While various early successes have been reported \citep{plis2014deep,Vieira2017UsingApplications}, deep learning models have often failed to outperform linear baselines \citep{Schulz2019dlbrains, he2020deep}. The mixed literature results are mirrored in recent challenges and competitions for machine learning in neuroimaging.
In a number of competitions, for instance in recent ABCD \citep{mihalik2019abcd}, PAC \citep{flint2019systematic}, and TADPOLE \citep{marinescu2018tadpole} challenges, classical machine learning models outperformed more complex deep learning approaches. In their review of deep learning in neuroimaging, \citet{Vieira2017UsingApplications} conclude that “despite the success of [deep learning] in several scientific areas, the superiority of this analytical approach in neuroimaging is yet to be demonstrated".

One possible explanation is the comparatively low sample size of neuroimaging studies. While computer vision and natural language processing (NLP) models are regularly trained on millions of data samples, a typical neuroimaging study will have only hundreds or at most thousands of participants. Given the arguably lower signal-to-noise ratio of neuroimaging data, one may expect a need for larger rather then smaller sample sizes to extract complex nonlinear interactions from neuroimaging data. 
This view is supported by research indicating that the performance of linear models on neuroimaging data is not yet saturated at $N\approx10000$ \citep{Schulz2019dlbrains}. If present sample sizes are insufficient to even fully characterize linear effects, then it seems unlikely that more complex nonlinear interactions can be extracted without further efforts in large scale data collection.

An alternative explanation for the difficulty of applying deep learning to psychiatry pertains to the high levels of noise in neuroimaging data (see section \ref{sec:noise}). \citet{Schulz2019dlbrains} argue that high levels of additive noise can in certain instances linearize the true decision boundary. In this case, no amount of additional training samples would allow complex nonlinear models such as deep neural networks to outperform linear baselines. Some nonlinear relationships in neuroimaging data could be inaccessible in principle.  

Whether inability to exploit nonlinear structure in some neuroimaging data is due to insufficient sample size, or inaccessible due to noise remains an open question. Nevertheless, recent research suggests that machine learning for psychiatry will not necessarily be solved by more and more complex and expressive models alone.
Researchers may instead need to focus on stronger inductive biases, which impose structural priors on the data to more efficiently learn statistical relationships from limited data \citep{battaglia2018relational} . Inductive biases that are specifically adapted to neuroimaging data (see section \ref{sec:inducbias}) may unlock prediction successes in the future.

\subsubsection{Lack of benchmark data sets} \label{sec:bench}
The literature contains a plethora of different deep neural network architectures, and new designs are continually being invented. Thus, reliable model comparisons are necessary to make an informed choice about which particular deep neural network architecture (or even which linear model) to use on a given brain imaging data set. Standardized benchmark data sets are crucial for a reliable and fair comparison of models and methods, and are an integral part of machine learning research (see, e.g., Imagenet  \citep{russakovsky2015imagenet} or  Cityscapes \citep{cordts2016cityscapes} for computer vision).
However, to the best of our knowledge, no large-scale neuroimaging benchmark data sets for the detection of psychiatric disorders are openly available and 
\citet{arbabshirani2017single} even state that comparison of accuracies between studies is ``essentially meaningless”. 
The necessity of benchmark data sets for model comparisons, particularly in neuroimaging-based psychiatric research, is due to broadly four reasons.

First, benchmark data sets can control for heterogeneity of the data sources. Cityscapes, for example, contains street scenery from 50 different cities, procured in different seasons and weather conditions. In multi-site neuroimaging data sets, heterogeneity is both a promise and a pitfall. Since the imaging site has a significant effect on prediction \citep{Glocker2019MachineLW}, it is beneficial for generalization to include several imaging sites, but it is also required to have similar label distributions across sites and to have high counts of samples per site.

Second, no further pre-processing of the data is required or code is available for additional steps such as cropping or intensity harmonization. In neuroimaging, MRI pre-processing is considerably more complex than on natural images. Consequently, the selection of pre-processing steps and tools have a significant impact on task-driven neurobiological inference and patient-wise prediction \citep{Bhagwat2020preprocessing}. Neuroimaging benchmark data sets therefore need to either come fully pre-processed or include code to repeat the pre-processing. The aim is not to find an optimal pre-processing strategy but rather to ensure replicability and comparability between studies. 

Third, the same motivation holds for a unique split of the data into training and testing before publication. As the heterogeneity of a data set is proportional to its size $N$ \citep{Schnack2016samplesize, VAROQUAUX2018samplesizes}, the prediction result will change drastically with different train and test splits. Therefore, in order to compare model performances between studies it is required that all studies use the same train and test split. Cityscapes has carefully designed a 3-way split of the data (train/validation/test) which attributes equal amounts of data based on geographical location, city size and season. Similar efforts would be suggested for neuroimaging data. Another downside of not sharing unique data splits is the risk of data leakage as described in section \ref{sec:val_metrics} which has been shown for AD classification in \citep{wen2020convolutional}.

And fourth, a benchmark data set for neuroimaging would need to provide an adequate level of task difficulty. 
On the one hand, the prediction task needs to be difficult enough for a superior model to clearly and reliably outperform simpler baselines. If a task is too simple and models are approaching the tasks' irreducible error (which might be the case for sex or AD classification), then it is unlikely that differentiation of model performance can be reliably observed. 
On the other hand, the task cannot be too complex, so that models can be successfully trained with available sample sizes. 

Creating large benchmark data sets typically requires the collaboration of several clinics to form a multi-site study. Even though, conducting multi-site machine learning studies seems highly promising, and is most likely a requirement for building generalizable models, properly designing multi-site studies remains a challenge. The main issue with multi-site data sets is that the data collected will have measurable differences between sites and perfectly matching participants between sites is not feasible. Most often, different sites have a different ratio of cases and controls. When aggregating those sites, covariates such as age, sex, total intracranial volume etc. might correlate strongly with the case-control ratio. High capacity machine learning models will easily learn to pickup those correlations to differentiate the MR images, rather than using disease specifying properties (see section \ref{sec:ab}). Hence, benchmarks on multi-site data sets might be overly optimistic in their results. Investigating the issue could be done by training a machine learning model solely on those covariates. If the covariates lead to a result which is not significantly different than the neuroimaging based model, then the results of the latter might have no meaning beyond portraying covariate differences. 

Even if not matching the criteria of benchmark data sets, the publication of large and open data sets is almost always valuable for machine learning research. However, for creating better benchmarks, study designers should ensure that the aforementioned points are met. This includes reporting any data source heterogeneity, the inclusion of a standardized processing pipeline, a pre-determined data set split and determination of clinically relevant questions to address, which have an appropriate difficulty. In addition, it is necessary to control for the correlation of covariates within and between sites. 

\subsubsection{Computational challenges}\label{sec:compu} %
Other impediments to the application of deep neural networks for neuroimaging pertain to computational cost. Although general processing GPUs (gpGPUs) have enabled deep learning to take off in many fields, computational resources remain expensive and especially neuroimaging-based deep learning is still limited by technical capacities. In contrast to other applications where cloud computing can be used, medical data typically needs to stay within the boundaries of the hospital and cannot be transferred to the cloud provider's network. 
Furthermore, as sMRI data is 3-dimensional it requires cubic operations leading to extensive GPU memory utilization. With images that often have more than a million voxels, the memory of a typical GPU is quickly exceeded when using very deep neural networks, large batch sizes or wide fully-connected layers. In fMRI, which has an additional time dimension, the issue becomes even more severe. Therefore, most researchers are highly limited in the amount of architecture search they can do. Many deep learning applications are extremely sensitive to hyperparameters. However, the optimization problem of finding hyperparameters is ill defined and often results in random searches. This means that usually the search space of hyperparameters is underexplored and one cannot prove that no better settings exist. This issue is similarly pressing when training simpler baseline models and comparing the suggested model to those baselines. Here, benchmark data sets as described in section \ref{sec:bench} could help by giving many researchers the opportunity to search for hyperparameters, allowing them to reference the current state-of-the-art results as baselines.

\subsection{Validity and explainability}\label{sec:valid}

\subsubsection{Algorithmic bias}\label{sec:ab} 

Challenges arise not only in training deep neural networks for psychiatry, but also in generalisation. How do trained models deal with new data samples that were generated after training and validation was already completed? In day-to-day clinical use, new data samples may differ from the training data in subtle but critical ways. The MRI scanner may be of a different model, scanning protocols or pre-processing pipelines may differ, and certain ethnicities or comorbidities may not have been included in the training data sample.
This can lead to critical errors. The predictions of a deep neural network are meaningful only for new data samples from the same distribution that generated the original training data \citep{szegedy2013intriguing}. \citet{goodfellow2014explaining} argue that ``classifiers based on modern machine learning techniques, even those that obtain excellent performance on the test set, are not learning the true underlying concepts that determine the correct output label. Instead, these algorithms have built a Potemkin village that works well on naturally occurring data, but is exposed as a fake when one visits points in space that do not have high probability in the data distribution".
A deep neural network will still make predictions for out-of-distribution samples, but these prediction will often be meaningless or misleading.


Furthermore, the expressivity of deep neural networks exacerbates distortions arising from sampling biases and confounding variables in the data \citep{wachinger2019quantifying, tommasi2017deeper}. A deep neural network has neither the goal nor the ability to prefer causal relationships over a spurious associations and will exploit whatever serves to predict the target variable \citep{Lapuschkin2019}. Hence, biases that exist in the training data will be reproduced in the learned model \citep{lewinn2017sample}. This phenomenon is well studied and to some extent under control for simple linear models, where (known) confounders can be anticipated and corrected for \citep{snoek2019control}. 
However, the highly expressive deep neural networks may pick up highly complex and unexpected biases in the training data. For instance, a deep neural network trained to classify skin lesions learned that dermatologists tend to include a ruler in the photos of skin lesions they where particularly concerned about it, and used the ruler as a proxy for malignancy \citep{narla2018automated}. Thus, one flip-side of the high expressive capacity of deep neural networks is the resulting difficulty to control for potential biases in the predictions.  
Such ``algorithmic biases" have diverse practical impacts. Minorities, who are often underrepresented in training data sets, have a high chance of being systematically misclassified, with potentially severe clinical consequences \citep{genomics2019, sirugo2019missing}.  Deep neural network's ability to exploit the most subtle biases in the training data represents a significant problem for multi-site studies where deep learning may pick up on even the slightest sampling differences between sites and thus use measurement site as a proxy for the actually studied phenomenon \citep{wachinger2019quantifying}, further corrupting predictions on new data.


\subsubsection{Confound modeling}\label{sec:cm}
While the preceding section was most concerned with prediction errors, similar complications arise in the interpretation of results. Frequently, researchers will not only attempt to predict a target variable but will investigate the relative contributions of different inputs on a model's accuracy, for instance to delineate neural correlates of a psychopathology from confounds like gender, age, or education which may influence the risk of disease.

The high expressivity of deep neural networks and the necessity of comparatively large sample sizes for training them 
may require novel approaches to disentangle the impact of confounds on prediction performance, as traditional approaches become increasingly impractical.
For instance, a priori balancing of the training data (i.e., matched, case controlled study design) is costly, limiting the achievable sample size and thus the prediction performance. Post hoc balancing by subsampling \citep{Rao2017-jj, Chyzhyk2018-fi} similarly reduces the effective sample size available for training the neural network.
Regressing out the influence of confounds directly from the neuroimaging data \citep{Dukart2011-ra,Todd2013-yh, Kostro2014-gs, Rao2017-jj} works for simple linear models, but will not completely eradicate complex nonlinear traces of the confounding variables from the input. Machine learning models, particularly the highly expressive deep neural networks, may still exploit residual information that could not be regressed out. In a model agnostic approach, \citet{Dinga2020-pa} engage this problem by controlling for confounds post hoc on the level of model predictions.

Deep learning does not only complicate confound modelling, but also allows for novel solutions to this problem. Some researchers propose using confounding variables as additional targets in a multi-task setting and to subsequently zero the model weights which are important for the confounding variable but not the target variable \citep{Ma2018-lt}. Others advocate incorporating confounds directly into the model (e.g., by disentangling them from the intermediate representations \citep{cheung2014discovering, pinaya2019using}), or even using adversarial optimisation objectives \citep{Adeli2019-qp} to train deep neural networks that correct against the influence of confounding variables.

Although properties  of deep neural networks may exacerbate problems of algorithmic bias and confound modelling, the flexibility of deep neural networks to incorporate and isolate confounding variables within the model may in the future offer uniquely powerful solutions to deal with heterogeneous biomedical imaging data.



\subsubsection{Explainability} 
\label{sec:blackbox}
Another way to check for biased or implausible predictions is to interrogate the models decision making process (see section \ref{sec:attribution}). However, even the features of linear models are not unambiguous in neuroimaging \citep{HAUFE201496,kindermans2017learning}. Since deep neural networks have orders of magnitudes more parameters than most other machine learning models, they are extremely hard to comprehend from a human perspective. While in computer vision we are able to visualize the filters of convolutional layers and to assign meaning to them (i.e. edge detectors or more abstract concepts such as eyes for face recognition \citep{Zeiler2014}), this becomes more challenging in neuroimaging as sMRI is 3-dimensional and abstract concepts are less immediate (e.g., how would a filter for atrophy look like?). 

Attribution methods provide a popular approach for reasoning about deep learning models. In high-risk settings such as medicine, the performance of an algorithm might have a vital impact on an individual. Hence, it has been argued that explainability of medical algorithms might become a prerequisite for clinical adoption. On the other side, it has been discussed (in, e.g., \citet{london2019accvsexpl}) whether explainability should be a requirement for machine learning systems that have a high accuracy, considering that the comprehension of many other causal mechanisms in medicine, such as the mechanisms of certain drugs or psychotherapy, is incomplete. In any case, \citet{Lapuschkin2019} have shown how post-hoc analyses of models with high accuracy can identify artifacts in the training data which the models misuse for classification. This showed that attribution methods can help to identify bias or subpar learning strategies, which in neuroimaging could for example be a focus on scanner artifacts, human motion, or random correlations between brain features and group membership. As a caveat, it was recently shown that most attribution methods only minimally change their produced heatmaps when several layers of the model are being randomized \citep{adebayo2018sanity, sixt2019explanations}; how to quantitatively assess attribution methods remains a challenge. Here, neuroimaging creates an opportunity, as the quantitative validation of attribution maps using neurobiological knowledge is more objective than the sole visual inspection \citep{Boehle2019, Eitel2019MS,eitel2019testing}.



        
        
        
        

\section{Conclusion}
While machine and in particular deep learning approaches provide a huge potential for transforming neuroimaging-based psychiatric research towards precision psychiatry, the field is still at the very beginning. From a structural perspective, the further collection of large, multi-site, standardized, and open-source neuroimaging data sets with overlapping disease categories and thorough clinical and behavioral characterization is essential. For addressing questions about disease prognosis and treatment outcome, longitudinal data will be increasingly important. In addition, efforts should be made in order to make analyses between different groups comparable, e.g., by creating benchmark data sets. The key to success will then be whether clinically meaningful and transparent representations can be learned from neuroimaging data that additionally account for individual and site-specific covariates.

\section{Acknowledgements}
We acknowledge support from the German Research Foundation (DFG, 389563835; 402170461-TRR 265; 414984028-CRC 1404), the Deutsche Multiple Sklerose Gesellschaft (DMSG), the Manfred and Ursula-Müller Stiftung, and the Brain \& Behavior Research Foundation (NARSAD grant; USA).









\bibliographystyle{elsarticle-num-names}
\bibliography{sample.bib}

\begin{thebibliography}{286}
\expandafter\ifx\csname natexlab\endcsname\relax\def\natexlab#1{#1}\fi
\providecommand{\url}[1]{\texttt{#1}}
\providecommand{\href}[2]{#2}
\providecommand{\path}[1]{#1}
\providecommand{\DOIprefix}{doi:}
\providecommand{\ArXivprefix}{arXiv:}
\providecommand{\URLprefix}{URL: }
\providecommand{\Pubmedprefix}{pmid:}
\providecommand{\doi}[1]{\href{http://dx.doi.org/#1}{\path{#1}}}
\providecommand{\Pubmed}[1]{\href{pmid:#1}{\path{#1}}}
\providecommand{\bibinfo}[2]{#2}
\ifx\xfnm\relax \def\xfnm[#1]{\unskip,\space#1}\fi
\bibitem[{LeCun et~al.(2015)LeCun, Bengio, and Hinton}]{Lecun2015}
\bibinfo{author}{Y.~LeCun}, \bibinfo{author}{Y.~Bengio},
  \bibinfo{author}{G.~Hinton},
\newblock \bibinfo{title}{{Deep learning}},
\newblock \bibinfo{journal}{Nature} \bibinfo{volume}{521}
  (\bibinfo{year}{2015}) \bibinfo{pages}{436--444}.
\bibitem[{Schmidhuber(2015)}]{schmidhuber2015deep}
\bibinfo{author}{J.~Schmidhuber},
\newblock \bibinfo{title}{Deep learning in neural networks: An overview},
\newblock \bibinfo{journal}{Neural Networks} \bibinfo{volume}{61}
  (\bibinfo{year}{2015}) \bibinfo{pages}{85--117}.
\bibitem[{Esteva et~al.(2017)Esteva, Kuprel, Novoa, Ko, Swetter, Blau, and
  Thrun}]{esteva2017dermatologist}
\bibinfo{author}{A.~Esteva}, \bibinfo{author}{B.~Kuprel},
  \bibinfo{author}{R.~A. Novoa}, \bibinfo{author}{J.~Ko},
  \bibinfo{author}{S.~M. Swetter}, \bibinfo{author}{H.~M. Blau},
  \bibinfo{author}{S.~Thrun},
\newblock \bibinfo{title}{Dermatologist-level classification of skin cancer
  with deep neural networks},
\newblock \bibinfo{journal}{Nature} \bibinfo{volume}{542}
  (\bibinfo{year}{2017}) \bibinfo{pages}{115--118}.
\bibitem[{Gulshan et~al.(2016)Gulshan, Peng, Coram, Stumpe, Wu, Narayanaswamy,
  Venugopalan, Widner, Madams, Cuadros et~al.}]{gulshan2016development}
\bibinfo{author}{V.~Gulshan}, \bibinfo{author}{L.~Peng},
  \bibinfo{author}{M.~Coram}, \bibinfo{author}{M.~C. Stumpe},
  \bibinfo{author}{D.~Wu}, \bibinfo{author}{A.~Narayanaswamy},
  \bibinfo{author}{S.~Venugopalan}, \bibinfo{author}{K.~Widner},
  \bibinfo{author}{T.~Madams}, \bibinfo{author}{J.~Cuadros}, et~al.,
\newblock \bibinfo{title}{Development and validation of a deep learning
  algorithm for detection of diabetic retinopathy in retinal fundus
  photographs},
\newblock \bibinfo{journal}{JAMA} \bibinfo{volume}{316} (\bibinfo{year}{2016})
  \bibinfo{pages}{2402--2410}.
\bibitem[{Litjens et~al.(2017)Litjens, Kooi, Bejnordi, Setio, Ciompi,
  Ghafoorian, Van Der~Laak, Van~Ginneken, and S{\'a}nchez}]{litjens2017}
\bibinfo{author}{G.~Litjens}, \bibinfo{author}{T.~Kooi}, \bibinfo{author}{B.~E.
  Bejnordi}, \bibinfo{author}{A.~A.~A. Setio}, \bibinfo{author}{F.~Ciompi},
  \bibinfo{author}{M.~Ghafoorian}, \bibinfo{author}{J.~A. Van Der~Laak},
  \bibinfo{author}{B.~Van~Ginneken}, \bibinfo{author}{C.~I. S{\'a}nchez},
\newblock \bibinfo{title}{A survey on deep learning in medical image analysis},
\newblock \bibinfo{journal}{Medical Image Analysis} \bibinfo{volume}{42}
  (\bibinfo{year}{2017}) \bibinfo{pages}{60--88}.
\bibitem[{Goodkind et~al.(2015)Goodkind, Eickhoff, Oathes, Jiang, Chang,
  Jones-Hagata, Ortega, Zaiko, Roach, Korgaonkar
  et~al.}]{goodkind2015identification}
\bibinfo{author}{M.~Goodkind}, \bibinfo{author}{S.~B. Eickhoff},
  \bibinfo{author}{D.~J. Oathes}, \bibinfo{author}{Y.~Jiang},
  \bibinfo{author}{A.~Chang}, \bibinfo{author}{L.~B. Jones-Hagata},
  \bibinfo{author}{B.~N. Ortega}, \bibinfo{author}{Y.~V. Zaiko},
  \bibinfo{author}{E.~L. Roach}, \bibinfo{author}{M.~S. Korgaonkar}, et~al.,
\newblock \bibinfo{title}{Identification of a common neurobiological substrate
  for mental illness},
\newblock \bibinfo{journal}{JAMA Psychiatry} \bibinfo{volume}{72}
  (\bibinfo{year}{2015}) \bibinfo{pages}{305--315}.
\bibitem[{Lui et~al.(2016)Lui, Zhou, Sweeney, and
  Gong}]{lui2016psychoradiology}
\bibinfo{author}{S.~Lui}, \bibinfo{author}{X.~J. Zhou}, \bibinfo{author}{J.~A.
  Sweeney}, \bibinfo{author}{Q.~Gong},
\newblock \bibinfo{title}{Psychoradiology: the frontier of neuroimaging in
  psychiatry},
\newblock \bibinfo{journal}{Radiology} \bibinfo{volume}{281}
  (\bibinfo{year}{2016}) \bibinfo{pages}{357--372}.
\bibitem[{Fu and Costafreda(2013)}]{fu2013neuroimaging}
\bibinfo{author}{C.~H. Fu}, \bibinfo{author}{S.~G. Costafreda},
\newblock \bibinfo{title}{Neuroimaging-based biomarkers in psychiatry: clinical
  opportunities of a paradigm shift},
\newblock \bibinfo{journal}{The Canadian Journal of Psychiatry}
  \bibinfo{volume}{58} (\bibinfo{year}{2013}) \bibinfo{pages}{499--508}.
\bibitem[{Zhao et~al.(2014)Zhao, Du, Huang, Lui, Chen, Liu, Luo, Wang, Kemp,
  and Gong}]{zhao2014brain}
\bibinfo{author}{Y.-J. Zhao}, \bibinfo{author}{M.-Y. Du},
  \bibinfo{author}{X.-Q. Huang}, \bibinfo{author}{S.~Lui},
  \bibinfo{author}{Z.-Q. Chen}, \bibinfo{author}{J.~Liu},
  \bibinfo{author}{Y.~Luo}, \bibinfo{author}{X.-L. Wang},
  \bibinfo{author}{G.~Kemp}, \bibinfo{author}{Q.-Y. Gong},
\newblock \bibinfo{title}{Brain grey matter abnormalities in medication-free
  patients with major depressive disorder: a meta-analysis},
\newblock \bibinfo{journal}{Psychological Medicine} \bibinfo{volume}{44}
  (\bibinfo{year}{2014}) \bibinfo{pages}{2927--2937}.
\bibitem[{Schmaal et~al.(2016)Schmaal, Veltman, van Erp, S{\"a}mann, Frodl,
  Jahanshad, Loehrer, Tiemeier, Hofman, Niessen
  et~al.}]{schmaal2016subcortical}
\bibinfo{author}{L.~Schmaal}, \bibinfo{author}{D.~J. Veltman},
  \bibinfo{author}{T.~G. van Erp}, \bibinfo{author}{P.~S{\"a}mann},
  \bibinfo{author}{T.~Frodl}, \bibinfo{author}{N.~Jahanshad},
  \bibinfo{author}{E.~Loehrer}, \bibinfo{author}{H.~Tiemeier},
  \bibinfo{author}{A.~Hofman}, \bibinfo{author}{W.~Niessen}, et~al.,
\newblock \bibinfo{title}{Subcortical brain alterations in major depressive
  disorder: findings from the {ENIGMA} {M}ajor {D}epressive {D}isorder working
  group},
\newblock \bibinfo{journal}{Molecular Psychiatry} \bibinfo{volume}{21}
  (\bibinfo{year}{2016}) \bibinfo{pages}{806--812}.
\bibitem[{Cole et~al.(2017)Cole, Poudel, Tsagkrasoulis, Caan, Steves, Spector,
  and Montana}]{cole2017predicting}
\bibinfo{author}{J.~H. Cole}, \bibinfo{author}{R.~P. Poudel},
  \bibinfo{author}{D.~Tsagkrasoulis}, \bibinfo{author}{M.~W. Caan},
  \bibinfo{author}{C.~Steves}, \bibinfo{author}{T.~D. Spector},
  \bibinfo{author}{G.~Montana},
\newblock \bibinfo{title}{Predicting brain age with deep learning from raw
  imaging data results in a reliable and heritable biomarker},
\newblock \bibinfo{journal}{NeuroImage} \bibinfo{volume}{163}
  (\bibinfo{year}{2017}) \bibinfo{pages}{115--124}.
\bibitem[{Patel et~al.(2012)Patel, Spreng, Shin, and
  Girard}]{patel2012neurocircuitry}
\bibinfo{author}{R.~Patel}, \bibinfo{author}{R.~N. Spreng},
  \bibinfo{author}{L.~M. Shin}, \bibinfo{author}{T.~A. Girard},
\newblock \bibinfo{title}{Neurocircuitry models of posttraumatic stress
  disorder and beyond: a meta-analysis of functional neuroimaging studies},
\newblock \bibinfo{journal}{Neuroscience \& Biobehavioral Reviews}
  \bibinfo{volume}{36} (\bibinfo{year}{2012}) \bibinfo{pages}{2130--2142}.
\bibitem[{Abi-Dargham and Horga(2016)}]{abi2016search}
\bibinfo{author}{A.~Abi-Dargham}, \bibinfo{author}{G.~Horga},
\newblock \bibinfo{title}{The search for imaging biomarkers in psychiatric
  disorders},
\newblock \bibinfo{journal}{Nature Medicine} \bibinfo{volume}{22}
  (\bibinfo{year}{2016}) \bibinfo{pages}{1248}.
\bibitem[{Klöppel et~al.(2011)Klöppel, Abdulkadir, Jack, Koutsouleris,
  Mourao-Miranda, and Vemuri}]{Kloeppel2011}
\bibinfo{author}{S.~Klöppel}, \bibinfo{author}{A.~Abdulkadir},
  \bibinfo{author}{C.~R. Jack}, \bibinfo{author}{N.~Koutsouleris},
  \bibinfo{author}{J.~Mourao-Miranda}, \bibinfo{author}{P.~Vemuri},
\newblock \bibinfo{title}{{Diagnostic neuroimaging across diseases}},
\newblock \bibinfo{journal}{NeuroImage} \bibinfo{volume}{61}
  (\bibinfo{year}{2011}) \bibinfo{pages}{457--463}.
\bibitem[{Orr{\`{u}} et~al.(2012)Orr{\`{u}}, Pettersson-Yeo, Marquand, Sartori,
  and Mechelli}]{Orru2012}
\bibinfo{author}{G.~Orr{\`{u}}}, \bibinfo{author}{W.~Pettersson-Yeo},
  \bibinfo{author}{A.~F. Marquand}, \bibinfo{author}{G.~Sartori},
  \bibinfo{author}{A.~Mechelli},
\newblock \bibinfo{title}{{Using support vector machine to identify imaging
  biomarkers of neurological and psychiatric disease: a critical review}},
\newblock \bibinfo{journal}{Neuroscience \& Biobehavioral Reviews}
  \bibinfo{volume}{36} (\bibinfo{year}{2012}) \bibinfo{pages}{1140--1152}.
\bibitem[{Wolfers et~al.(2015)Wolfers, Buitelaar, Beckmann, Franke, and
  Marquand}]{Wolfers2015}
\bibinfo{author}{T.~Wolfers}, \bibinfo{author}{J.~K. Buitelaar},
  \bibinfo{author}{C.~Beckmann}, \bibinfo{author}{B.~Franke},
  \bibinfo{author}{A.~F. Marquand},
\newblock \bibinfo{title}{{From estimating activation locality to predicting
  disorder: a review of pattern recognition for neuroimaging-based psychiatric
  diagnostics}},
\newblock \bibinfo{journal}{Neuroscience {\&} Biobehavioral Reviews}
  \bibinfo{volume}{57} (\bibinfo{year}{2015}) \bibinfo{pages}{328--349}.
\bibitem[{Bzdok and Meyer-Lindenberg(2018)}]{bzdok2018machine}
\bibinfo{author}{D.~Bzdok}, \bibinfo{author}{A.~Meyer-Lindenberg},
\newblock \bibinfo{title}{Machine learning for precision psychiatry:
  opportunities and challenges},
\newblock \bibinfo{journal}{Biological Psychiatry: Cognitive Neuroscience and
  Neuroimaging} \bibinfo{volume}{3} (\bibinfo{year}{2018})
  \bibinfo{pages}{223--230}.
\bibitem[{Walter et~al.(2019)Walter, Alizadeh, Jamalabadi, Lueken, Dannlowski,
  Walter, Olbrich, Colic, Kambeitz, Koutsouleris
  et~al.}]{walter2019translational}
\bibinfo{author}{M.~Walter}, \bibinfo{author}{S.~Alizadeh},
  \bibinfo{author}{H.~Jamalabadi}, \bibinfo{author}{U.~Lueken},
  \bibinfo{author}{U.~Dannlowski}, \bibinfo{author}{H.~Walter},
  \bibinfo{author}{S.~Olbrich}, \bibinfo{author}{L.~Colic},
  \bibinfo{author}{J.~Kambeitz}, \bibinfo{author}{N.~Koutsouleris}, et~al.,
\newblock \bibinfo{title}{Translational machine learning for psychiatric
  neuroimaging},
\newblock \bibinfo{journal}{Progress in Neuro-Psychopharmacology and Biological
  Psychiatry} \bibinfo{volume}{91} (\bibinfo{year}{2019})
  \bibinfo{pages}{113--121}.
\bibitem[{Obermeyer and Emanuel(2016)}]{Obermeyer2016}
\bibinfo{author}{Z.~Obermeyer}, \bibinfo{author}{E.~J. Emanuel},
\newblock \bibinfo{title}{{Predicting the Future — Big Data, Machine
  Learning, and Clinical Medicine}},
\newblock \bibinfo{journal}{New England Journal of Medicine}
  \bibinfo{volume}{375} (\bibinfo{year}{2016}) \bibinfo{pages}{1216--1219}.
\bibitem[{Durstewitz et~al.(2019)Durstewitz, Koppe, and
  Meyer-Lindenberg}]{durstewitz2019deep}
\bibinfo{author}{D.~Durstewitz}, \bibinfo{author}{G.~Koppe},
  \bibinfo{author}{A.~Meyer-Lindenberg},
\newblock \bibinfo{title}{Deep neural networks in psychiatry},
\newblock \bibinfo{journal}{Molecular Psychiatry} \bibinfo{volume}{24}
  (\bibinfo{year}{2019}) \bibinfo{pages}{1583--1598}.
\bibitem[{Marquand et~al.(2016)Marquand, Rezek, Buitelaar, and
  Beckmann}]{Marquand2016}
\bibinfo{author}{A.~F. Marquand}, \bibinfo{author}{I.~Rezek},
  \bibinfo{author}{J.~Buitelaar}, \bibinfo{author}{C.~F. Beckmann},
\newblock \bibinfo{title}{{Understanding Heterogeneity in Clinical Cohorts
  Using Normative Models: Beyond Case-Control Studies}},
\newblock \bibinfo{journal}{Biological Psychiatry} \bibinfo{volume}{80}
  (\bibinfo{year}{2016}) \bibinfo{pages}{552--561}.
\bibitem[{Lueken et~al.(2016)Lueken, Zierhut, Hahn, Straube, Kircher, Reif,
  Richter, Hamm, Wittchen, and Domschke}]{lueken2016neurobiological}
\bibinfo{author}{U.~Lueken}, \bibinfo{author}{K.~C. Zierhut},
  \bibinfo{author}{T.~Hahn}, \bibinfo{author}{B.~Straube},
  \bibinfo{author}{T.~Kircher}, \bibinfo{author}{A.~Reif},
  \bibinfo{author}{J.~Richter}, \bibinfo{author}{A.~Hamm},
  \bibinfo{author}{H.-U. Wittchen}, \bibinfo{author}{K.~Domschke},
\newblock \bibinfo{title}{Neurobiological markers predicting treatment response
  in anxiety disorders: A systematic review and implications for clinical
  application},
\newblock \bibinfo{journal}{Neuroscience \& Biobehavioral Reviews}
  \bibinfo{volume}{66} (\bibinfo{year}{2016}) \bibinfo{pages}{143--162}.
\bibitem[{Woo et~al.(2017)Woo, Chang, Lindquist, and
  Wager}]{Woo2017BuildingNeuroimaging}
\bibinfo{author}{C.-W. Woo}, \bibinfo{author}{L.~J. Chang},
  \bibinfo{author}{M.~A. Lindquist}, \bibinfo{author}{T.~D. Wager},
\newblock \bibinfo{title}{{Building better biomarkers: brain models in
  translational neuroimaging}},
\newblock \bibinfo{journal}{Nature Neuroscience} \bibinfo{volume}{20}
  (\bibinfo{year}{2017}) \bibinfo{pages}{365--377}.
\bibitem[{Vieira et~al.(2017)Vieira, Pinaya, and
  Mechelli}]{Vieira2017UsingApplications}
\bibinfo{author}{S.~Vieira}, \bibinfo{author}{W.~H. Pinaya},
  \bibinfo{author}{A.~Mechelli},
\newblock \bibinfo{title}{{Using deep learning to investigate the neuroimaging
  correlates of psychiatric and neurological disorders: Methods and
  applications}},
\newblock \bibinfo{journal}{Neuroscience {\&} Biobehavioral Reviews}
  \bibinfo{volume}{74} (\bibinfo{year}{2017}) \bibinfo{pages}{58--75}.
\bibitem[{Bzdok and Yeo(2017)}]{2017inference}
\bibinfo{author}{D.~Bzdok}, \bibinfo{author}{B.~T. Yeo},
\newblock \bibinfo{title}{Inference in the age of big data: Future perspectives
  on neuroscience},
\newblock \bibinfo{journal}{NeuroImage} \bibinfo{volume}{155}
  (\bibinfo{year}{2017}) \bibinfo{pages}{549--564}.
\bibitem[{Bzdok et~al.(2018)Bzdok, Altman, and Krzywinski}]{bzdok2018points}
\bibinfo{author}{D.~Bzdok}, \bibinfo{author}{N.~Altman},
  \bibinfo{author}{M.~Krzywinski},
\newblock \bibinfo{title}{Points of significance: Statistics versus machine
  learning},
\newblock \bibinfo{journal}{Nature Methods} \bibinfo{volume}{15}
  (\bibinfo{year}{2018}) \bibinfo{pages}{233--235}.
\bibitem[{Hubel and Wiesel(1968)}]{Hubel1968}
\bibinfo{author}{D.~H. Hubel}, \bibinfo{author}{T.~N. Wiesel},
\newblock \bibinfo{title}{{Receptive fields and functional architecture of
  monkey striate cortex}},
\newblock \bibinfo{journal}{The Journal of Physiology} \bibinfo{volume}{195}
  (\bibinfo{year}{1968}) \bibinfo{pages}{215--243}.
\bibitem[{Lundervold and Lundervold(2019)}]{LUNDERVOLD2019102}
\bibinfo{author}{A.~S. Lundervold}, \bibinfo{author}{A.~Lundervold},
\newblock \bibinfo{title}{{An overview of deep learning in medical imaging
  focusing on MRI}},
\newblock \bibinfo{journal}{Zeitschrift f{\"{u}}r Medizinische Physik}
  \bibinfo{volume}{29} (\bibinfo{year}{2019}) \bibinfo{pages}{102--127}.
\bibitem[{Valliani et~al.(2019)Valliani, Ranti, and Oermann}]{valliani2019deep}
\bibinfo{author}{A.~A.-A. Valliani}, \bibinfo{author}{D.~Ranti},
  \bibinfo{author}{E.~K. Oermann},
\newblock \bibinfo{title}{Deep learning and neurology: A systematic review},
\newblock \bibinfo{journal}{Neurology and Therapy}  (\bibinfo{year}{2019})
  \bibinfo{pages}{1--15}.
\bibitem[{B{\"{o}}hle et~al.(2019)B{\"{o}}hle, Eitel, Weygandt, and
  Ritter}]{Boehle2019}
\bibinfo{author}{M.~B{\"{o}}hle}, \bibinfo{author}{F.~Eitel},
  \bibinfo{author}{M.~Weygandt}, \bibinfo{author}{K.~Ritter},
\newblock \bibinfo{title}{{Layer-Wise Relevance Propagation for Explaining Deep
  Neural Network Decisions in MRI-Based Alzheimer's Disease Classification}},
\newblock \bibinfo{journal}{Frontiers in Aging Neuroscience}
  \bibinfo{volume}{11} (\bibinfo{year}{2019}) \bibinfo{pages}{194}.
\bibitem[{Eitel et~al.(2019)Eitel, Soehler, Bellmann-Strobl, Brandt, Ruprecht,
  Giess, Kuchling, Asseyer, Weygandt, Haynes, Scheel, Paul, and
  Ritter}]{Eitel2019MS}
\bibinfo{author}{F.~Eitel}, \bibinfo{author}{E.~Soehler},
  \bibinfo{author}{J.~Bellmann-Strobl}, \bibinfo{author}{A.~U. Brandt},
  \bibinfo{author}{K.~Ruprecht}, \bibinfo{author}{R.~M. Giess},
  \bibinfo{author}{J.~Kuchling}, \bibinfo{author}{S.~Asseyer},
  \bibinfo{author}{M.~Weygandt}, \bibinfo{author}{J.-D. Haynes},
  \bibinfo{author}{M.~Scheel}, \bibinfo{author}{F.~Paul},
  \bibinfo{author}{K.~Ritter},
\newblock \bibinfo{title}{{Uncovering convolutional neural network decisions
  for diagnosing multiple sclerosis on conventional MRI using layer-wise
  relevance propagation}},
\newblock \bibinfo{journal}{NeuroImage: Clinical} \bibinfo{volume}{24}
  (\bibinfo{year}{2019}) \bibinfo{pages}{102003}.
\bibitem[{Schulz et~al.(2020)Schulz, Yeo, Vogelstein, Mourao-Miranada, Kather,
  Kording, Richards, and Bzdok}]{Schulz2019dlbrains}
\bibinfo{author}{M.-A. Schulz}, \bibinfo{author}{B.~T. Yeo},
  \bibinfo{author}{J.~T. Vogelstein}, \bibinfo{author}{J.~Mourao-Miranada},
  \bibinfo{author}{J.~N. Kather}, \bibinfo{author}{K.~Kording},
  \bibinfo{author}{B.~Richards}, \bibinfo{author}{D.~Bzdok},
\newblock \bibinfo{title}{Different scaling of linear models and deep learning
  in {UKB}iobank brain images versus machine-learning datasets},
\newblock \bibinfo{journal}{Nature Communications} \bibinfo{volume}{11}
  (\bibinfo{year}{2020}) \bibinfo{pages}{1--15}.
\bibitem[{He et~al.(2020)He, Kong, Holmes, Nguyen, Sabuncu, Eickhoff, Bzdok,
  Feng, and Yeo}]{he2020deep}
\bibinfo{author}{T.~He}, \bibinfo{author}{R.~Kong}, \bibinfo{author}{A.~J.
  Holmes}, \bibinfo{author}{M.~Nguyen}, \bibinfo{author}{M.~R. Sabuncu},
  \bibinfo{author}{S.~B. Eickhoff}, \bibinfo{author}{D.~Bzdok},
  \bibinfo{author}{J.~Feng}, \bibinfo{author}{B.~T. Yeo},
\newblock \bibinfo{title}{Deep neural networks and kernel regression achieve
  comparable accuracies for functional connectivity prediction of behavior and
  demographics},
\newblock \bibinfo{journal}{NeuroImage} \bibinfo{volume}{206}
  (\bibinfo{year}{2020}) \bibinfo{pages}{116276}.
\bibitem[{Rokham et~al.(2020)Rokham, Pearlson, Abrol, Falakshahi, Plis, and
  Calhoun}]{rokham2020addressing}
\bibinfo{author}{H.~Rokham}, \bibinfo{author}{G.~Pearlson},
  \bibinfo{author}{A.~Abrol}, \bibinfo{author}{H.~Falakshahi},
  \bibinfo{author}{S.~Plis}, \bibinfo{author}{V.~D. Calhoun},
\newblock \bibinfo{title}{Addressing inaccurate nosology in mental health: A
  multi label data cleansing approach for detecting label noise from structural
  magnetic resonance imaging data in mood and psychosis disorders},
\newblock \bibinfo{journal}{Biological Psychiatry: Cognitive Neuroscience and
  Neuroimaging}  (\bibinfo{year}{2020}).
\bibitem[{Cuthbert and Insel(2013)}]{Cuthbert2013}
\bibinfo{author}{B.~N. Cuthbert}, \bibinfo{author}{T.~R. Insel},
\newblock \bibinfo{title}{Toward the future of psychiatric diagnosis: the seven
  pillars of {RD}o{C}},
\newblock \bibinfo{journal}{BMC Medicine} \bibinfo{volume}{11}
  (\bibinfo{year}{2013}) \bibinfo{pages}{126}.
\bibitem[{Insel and Cuthbert(2015)}]{Insel2015}
\bibinfo{author}{T.~R. Insel}, \bibinfo{author}{B.~N. Cuthbert},
\newblock \bibinfo{title}{{Brain disorders? Precisely}},
\newblock \bibinfo{journal}{Science} \bibinfo{volume}{348}
  (\bibinfo{year}{2015}) \bibinfo{pages}{499--500}.
\bibitem[{Rajkomar et~al.(2019)Rajkomar, Dean, and
  Kohane}]{rajkomar2019machine}
\bibinfo{author}{A.~Rajkomar}, \bibinfo{author}{J.~Dean},
  \bibinfo{author}{I.~Kohane},
\newblock \bibinfo{title}{Machine learning in medicine},
\newblock \bibinfo{journal}{New England Journal of Medicine}
  \bibinfo{volume}{380} (\bibinfo{year}{2019}) \bibinfo{pages}{1347--1358}.
\bibitem[{Rowe(2019)}]{rowe2019introduction}
\bibinfo{author}{M.~Rowe},
\newblock \bibinfo{title}{An introduction to machine learning for clinicians},
\newblock \bibinfo{journal}{Academic Medicine} \bibinfo{volume}{94}
  (\bibinfo{year}{2019}) \bibinfo{pages}{1433--1436}.
\bibitem[{Schellinger et~al.(2010)Schellinger, Bryan, Caplan, Detre, Edelman,
  Jaigobin, Kidwell, Mohr, Sloan, Sorensen et~al.}]{schellinger2010evidence}
\bibinfo{author}{P.~Schellinger}, \bibinfo{author}{R.~Bryan},
  \bibinfo{author}{L.~Caplan}, \bibinfo{author}{J.~Detre},
  \bibinfo{author}{R.~Edelman}, \bibinfo{author}{C.~Jaigobin},
  \bibinfo{author}{C.~Kidwell}, \bibinfo{author}{J.~Mohr},
  \bibinfo{author}{M.~Sloan}, \bibinfo{author}{A.~Sorensen}, et~al.,
\newblock \bibinfo{title}{Evidence-based guideline: the role of diffusion and
  perfusion {MRI} for the diagnosis of acute ischemic stroke: report of the
  {T}herapeutics and {T}echnology {A}ssessment {S}ubcommittee of the {A}merican
  {A}cademy of {N}eurology},
\newblock \bibinfo{journal}{Neurology} \bibinfo{volume}{75}
  (\bibinfo{year}{2010}) \bibinfo{pages}{177--185}.
\bibitem[{Geraldes et~al.(2018)Geraldes, Ciccarelli, Barkhof, De~Stefano,
  Enzinger, Filippi, Hofer, Paul, Preziosa, Rovira
  et~al.}]{geraldes2018current}
\bibinfo{author}{R.~Geraldes}, \bibinfo{author}{O.~Ciccarelli},
  \bibinfo{author}{F.~Barkhof}, \bibinfo{author}{N.~De~Stefano},
  \bibinfo{author}{C.~Enzinger}, \bibinfo{author}{M.~Filippi},
  \bibinfo{author}{M.~Hofer}, \bibinfo{author}{F.~Paul},
  \bibinfo{author}{P.~Preziosa}, \bibinfo{author}{A.~Rovira}, et~al.,
\newblock \bibinfo{title}{The current role of {MRI} in differentiating multiple
  sclerosis from its imaging mimics},
\newblock \bibinfo{journal}{Nature Reviews Neurology} \bibinfo{volume}{14}
  (\bibinfo{year}{2018}) \bibinfo{pages}{199}.
\bibitem[{Mugler~III and Brookeman(1990)}]{mugler1990three}
\bibinfo{author}{J.~P. Mugler~III}, \bibinfo{author}{J.~R. Brookeman},
\newblock \bibinfo{title}{Three-dimensional magnetization-prepared rapid
  gradient-echo imaging (3{D} {MP} {RAGE})},
\newblock \bibinfo{journal}{Magnetic Resonance in Medicine}
  \bibinfo{volume}{15} (\bibinfo{year}{1990}) \bibinfo{pages}{152--157}.
\bibitem[{Moncrieff and Leo(2010)}]{moncrieff2010systematic}
\bibinfo{author}{J.~Moncrieff}, \bibinfo{author}{J.~Leo},
\newblock \bibinfo{title}{A systematic review of the effects of antipsychotic
  drugs on brain volume},
\newblock \bibinfo{journal}{Psychological Medicine} \bibinfo{volume}{40}
  (\bibinfo{year}{2010}) \bibinfo{pages}{1409}.
\bibitem[{Campbell and MacQueen(2004)}]{campbell2004role}
\bibinfo{author}{S.~Campbell}, \bibinfo{author}{G.~MacQueen},
\newblock \bibinfo{title}{The role of the hippocampus in the pathophysiology of
  major depression.},
\newblock \bibinfo{journal}{Journal of Psychiatry \& Neuroscience}
  (\bibinfo{year}{2004}).
\bibitem[{Glover(2011)}]{glover2011overview}
\bibinfo{author}{G.~H. Glover},
\newblock \bibinfo{title}{Overview of functional magnetic resonance imaging},
\newblock \bibinfo{journal}{Neurosurgery Clinics} \bibinfo{volume}{22}
  (\bibinfo{year}{2011}) \bibinfo{pages}{133--139}.
\bibitem[{Khosla et~al.(2019)Khosla, Jamison, Ngo, Kuceyeski, and
  Sabuncu}]{khosla2019machine}
\bibinfo{author}{M.~Khosla}, \bibinfo{author}{K.~Jamison},
  \bibinfo{author}{G.~H. Ngo}, \bibinfo{author}{A.~Kuceyeski},
  \bibinfo{author}{M.~R. Sabuncu},
\newblock \bibinfo{title}{Machine learning in resting-state f{MRI} analysis},
\newblock \bibinfo{journal}{Magnetic Resonance Imaging} \bibinfo{volume}{64}
  (\bibinfo{year}{2019}) \bibinfo{pages}{101--121}.
\bibitem[{Pervaiz et~al.(2020)Pervaiz, Vidaurre, Woolrich, and
  Smith}]{pervaiz2020optimising}
\bibinfo{author}{U.~Pervaiz}, \bibinfo{author}{D.~Vidaurre},
  \bibinfo{author}{M.~W. Woolrich}, \bibinfo{author}{S.~M. Smith},
\newblock \bibinfo{title}{Optimising network modelling methods for f{MRI}},
\newblock \bibinfo{journal}{NeuroImage} \bibinfo{volume}{211}
  (\bibinfo{year}{2020}) \bibinfo{pages}{116604}.
\bibitem[{Specht(2019)}]{specht2019current}
\bibinfo{author}{K.~Specht},
\newblock \bibinfo{title}{Current challenges in translational and clinical
  f{MRI} and future directions},
\newblock \bibinfo{journal}{Frontiers in Psychiatry} \bibinfo{volume}{10}
  (\bibinfo{year}{2019}).
\bibitem[{Ashburner and Friston(2007)}]{Ashburner2007}
\bibinfo{author}{J.~Ashburner}, \bibinfo{author}{K.~J. Friston},
\newblock \bibinfo{title}{Chapter 6 - segmentation},
\newblock in: \bibinfo{editor}{K.~J. Friston}, \bibinfo{editor}{J.~T.
  Ashburner}, \bibinfo{editor}{S.~Kiebel}, \bibinfo{editor}{T.~Nichols},
  \bibinfo{editor}{W.~D. Penny} (Eds.), \bibinfo{booktitle}{{Statistical
  Parametric Mapping: The Analysis of Functional Brain Images}},
  \bibinfo{publisher}{Academic Press}, \bibinfo{year}{2007}, pp.
  \bibinfo{pages}{81--91}.
\bibitem[{Bishop(2007)}]{Bishop2007}
\bibinfo{author}{C.~M. Bishop}, \bibinfo{title}{{Pattern {R}ecognition and
  {M}achine {L}earning}}, \bibinfo{publisher}{Springer}, \bibinfo{year}{2007}.
\bibitem[{Goodfellow et~al.(2016)Goodfellow, Bengio, and
  Courville}]{Goodfellow2016}
\bibinfo{author}{I.~Goodfellow}, \bibinfo{author}{Y.~Bengio},
  \bibinfo{author}{A.~Courville}, \bibinfo{title}{Deep learning},
  volume~\bibinfo{volume}{1}, \bibinfo{publisher}{MIT Press},
  \bibinfo{year}{2016}.
\bibitem[{Buchanan and Duda(1983)}]{Buchanan1983}
\bibinfo{author}{B.~G. Buchanan}, \bibinfo{author}{R.~O. Duda},
\newblock \bibinfo{title}{Principles of rule-based expert systems},
\newblock in: \bibinfo{booktitle}{Advances in Computers},
  volume~\bibinfo{volume}{22}, \bibinfo{publisher}{Elsevier},
  \bibinfo{year}{1983}, pp. \bibinfo{pages}{163--216}.
\bibitem[{Juang and Rabiner(1991)}]{Juang1991}
\bibinfo{author}{B.~H. Juang}, \bibinfo{author}{L.~R. Rabiner},
\newblock \bibinfo{title}{Hidden markov models for speech recognition},
\newblock \bibinfo{journal}{Technometrics} \bibinfo{volume}{33}
  (\bibinfo{year}{1991}) \bibinfo{pages}{251--272}.
\bibitem[{Rosten and Drummond(2006)}]{Rosten2006}
\bibinfo{author}{E.~Rosten}, \bibinfo{author}{T.~Drummond},
\newblock \bibinfo{title}{Machine learning for high-speed corner detection},
\newblock in: \bibinfo{booktitle}{European Conference on Computer Vision},
  \bibinfo{organization}{Springer}, \bibinfo{year}{2006}, pp.
  \bibinfo{pages}{430--443}.
\bibitem[{Libbrecht and Noble(2015)}]{Libbrecht2015}
\bibinfo{author}{M.~W. Libbrecht}, \bibinfo{author}{W.~S. Noble},
\newblock \bibinfo{title}{Machine learning applications in genetics and
  genomics},
\newblock \bibinfo{journal}{Nature Reviews Genetics} \bibinfo{volume}{16}
  (\bibinfo{year}{2015}) \bibinfo{pages}{321--332}.
\bibitem[{Mateos-Pérez et~al.(2018)Mateos-Pérez, Dadar, Lacalle-Aurioles,
  Iturria-Medina, Zeighami, and Evans}]{mateoperez2018survey}
\bibinfo{author}{J.~M. Mateos-Pérez}, \bibinfo{author}{M.~Dadar},
  \bibinfo{author}{M.~Lacalle-Aurioles}, \bibinfo{author}{Y.~Iturria-Medina},
  \bibinfo{author}{Y.~Zeighami}, \bibinfo{author}{A.~C. Evans},
\newblock \bibinfo{title}{Structural neuroimaging as clinical predictor: A
  review of machine learning applications},
\newblock \bibinfo{journal}{NeuroImage: Clinical} \bibinfo{volume}{20}
  (\bibinfo{year}{2018}) \bibinfo{pages}{506 -- 522}.
\bibitem[{Hastie et~al.(2009)Hastie, Tibshirani, and
  Friedman}]{hastie2009elements}
\bibinfo{author}{T.~Hastie}, \bibinfo{author}{R.~Tibshirani},
  \bibinfo{author}{J.~Friedman}, \bibinfo{title}{The elements of statistical
  learning: data mining, inference, and prediction},
  \bibinfo{publisher}{Springer Science \& Business Media},
  \bibinfo{year}{2009}.
\bibitem[{Wolpert and Macready(1997)}]{Wolpert1997}
\bibinfo{author}{D.~H. Wolpert}, \bibinfo{author}{W.~G. Macready},
\newblock \bibinfo{title}{No free lunch theorems for optimization},
\newblock \bibinfo{journal}{IEEE Transactions on Evolutionary Computation}
  \bibinfo{volume}{1} (\bibinfo{year}{1997}) \bibinfo{pages}{67--82}.
\bibitem[{Wen et~al.(2019)Wen, Thibeau, Samper-Gonz{\'a}lez, Routier, Bottani,
  Dormont, Durrleman, Colliot, Burgos et~al.}]{wen2019serious}
\bibinfo{author}{J.~Wen}, \bibinfo{author}{E.~Thibeau},
  \bibinfo{author}{J.~Samper-Gonz{\'a}lez}, \bibinfo{author}{A.~Routier},
  \bibinfo{author}{S.~Bottani}, \bibinfo{author}{D.~Dormont},
  \bibinfo{author}{S.~Durrleman}, \bibinfo{author}{O.~Colliot},
  \bibinfo{author}{N.~Burgos}, et~al.,
\newblock \bibinfo{title}{How serious is data leakage in deep learning studies
  on {A}lzheimer's disease classification?},
\newblock in: \bibinfo{booktitle}{{2019 OHBM Annual meeting - Organization for
  Human Brain Mapping}}, \bibinfo{year}{2019}.
\bibitem[{Wen et~al.(2020)Wen, Thibeau-Sutre, Diaz-Melo, Samper-Gonz{\'a}lez,
  Routier, Bottani, Dormont, Durrleman, Burgos, Colliot
  et~al.}]{wen2020convolutional}
\bibinfo{author}{J.~Wen}, \bibinfo{author}{E.~Thibeau-Sutre},
  \bibinfo{author}{M.~Diaz-Melo}, \bibinfo{author}{J.~Samper-Gonz{\'a}lez},
  \bibinfo{author}{A.~Routier}, \bibinfo{author}{S.~Bottani},
  \bibinfo{author}{D.~Dormont}, \bibinfo{author}{S.~Durrleman},
  \bibinfo{author}{N.~Burgos}, \bibinfo{author}{O.~Colliot}, et~al.,
\newblock \bibinfo{title}{Convolutional neural networks for classification of
  {A}lzheimer's disease: Overview and reproducible evaluation},
\newblock \bibinfo{journal}{Medical Image Analysis}  (\bibinfo{year}{2020})
  \bibinfo{pages}{101694}.
\bibitem[{Klöppel et~al.(2008)Klöppel, Stonnington, Chu, Draganski, Scahill,
  Rohrer, Fox, Jack, Ashburner, and Frackowiak}]{Kloppel2008}
\bibinfo{author}{S.~Klöppel}, \bibinfo{author}{C.~M. Stonnington},
  \bibinfo{author}{C.~Chu}, \bibinfo{author}{B.~Draganski},
  \bibinfo{author}{R.~I. Scahill}, \bibinfo{author}{J.~D. Rohrer},
  \bibinfo{author}{N.~C. Fox}, \bibinfo{author}{C.~R. Jack},
  \bibinfo{author}{J.~Ashburner}, \bibinfo{author}{R.~S.~J. Frackowiak},
\newblock \bibinfo{title}{{Automatic classification of {MR} scans in
  {A}lzheimer's disease}},
\newblock \bibinfo{journal}{Brain} \bibinfo{volume}{131} (\bibinfo{year}{2008})
  \bibinfo{pages}{681--689}.
\bibitem[{Weygandt et~al.(2011)Weygandt, Hackmack, Pfueller, Bellmann-Strobl,
  Paul, Zipp, and Haynes}]{Weygandt2011}
\bibinfo{author}{M.~Weygandt}, \bibinfo{author}{K.~Hackmack},
  \bibinfo{author}{C.~Pfueller}, \bibinfo{author}{J.~Bellmann-Strobl},
  \bibinfo{author}{F.~Paul}, \bibinfo{author}{F.~Zipp}, \bibinfo{author}{J.-D.
  Haynes},
\newblock \bibinfo{title}{{{MRI} {P}attern {R}ecognition in {M}ultiple
  {S}clerosis {N}ormal-{A}ppearing {B}rain {A}reas}},
\newblock \bibinfo{journal}{PLOS ONE} \bibinfo{volume}{6}
  (\bibinfo{year}{2011}) \bibinfo{pages}{e21138}.
\bibitem[{Kohavi(1995)}]{Kohavi1995}
\bibinfo{author}{R.~Kohavi},
\newblock \bibinfo{title}{{A study of cross-validation and bootstrap for
  accuracy estimation and model selection}},
\newblock in: \bibinfo{booktitle}{IJCAI}, \bibinfo{year}{1995}, pp.
  \bibinfo{pages}{1137--1145}.
\bibitem[{Varoquaux et~al.(2017)Varoquaux, Raamana, Engemann, Hoyos-Idrobo,
  Schwartz, and Thirion}]{varoquaux2017assessing}
\bibinfo{author}{G.~Varoquaux}, \bibinfo{author}{P.~R. Raamana},
  \bibinfo{author}{D.~A. Engemann}, \bibinfo{author}{A.~Hoyos-Idrobo},
  \bibinfo{author}{Y.~Schwartz}, \bibinfo{author}{B.~Thirion},
\newblock \bibinfo{title}{Assessing and tuning brain decoders:
  cross-validation, caveats, and guidelines},
\newblock \bibinfo{journal}{NeuroImage} \bibinfo{volume}{145}
  (\bibinfo{year}{2017}) \bibinfo{pages}{166--179}.
\bibitem[{Jollans et~al.(2019)Jollans, Boyle, Artiges, Banaschewski,
  Desrivi{\`e}res, Grigis, Martinot, Paus, Smolka, Walter
  et~al.}]{jollans2019quantifying}
\bibinfo{author}{L.~Jollans}, \bibinfo{author}{R.~Boyle},
  \bibinfo{author}{E.~Artiges}, \bibinfo{author}{T.~Banaschewski},
  \bibinfo{author}{S.~Desrivi{\`e}res}, \bibinfo{author}{A.~Grigis},
  \bibinfo{author}{J.-L. Martinot}, \bibinfo{author}{T.~Paus},
  \bibinfo{author}{M.~N. Smolka}, \bibinfo{author}{H.~Walter}, et~al.,
\newblock \bibinfo{title}{Quantifying performance of machine learning methods
  for neuroimaging data},
\newblock \bibinfo{journal}{NeuroImage} \bibinfo{volume}{199}
  (\bibinfo{year}{2019}) \bibinfo{pages}{351--365}.
\bibitem[{Koutsouleris et~al.(2018)Koutsouleris, Kambeitz-Ilankovic, Ruhrmann,
  Rosen, Ruef, Dwyer, Paolini, Chisholm, Kambeitz, Haidl
  et~al.}]{koutsouleris2018prediction}
\bibinfo{author}{N.~Koutsouleris}, \bibinfo{author}{L.~Kambeitz-Ilankovic},
  \bibinfo{author}{S.~Ruhrmann}, \bibinfo{author}{M.~Rosen},
  \bibinfo{author}{A.~Ruef}, \bibinfo{author}{D.~B. Dwyer},
  \bibinfo{author}{M.~Paolini}, \bibinfo{author}{K.~Chisholm},
  \bibinfo{author}{J.~Kambeitz}, \bibinfo{author}{T.~Haidl}, et~al.,
\newblock \bibinfo{title}{Prediction models of functional outcomes for
  individuals in the clinical high-risk state for psychosis or with
  recent-onset depression: a multimodal, multisite machine learning analysis},
\newblock \bibinfo{journal}{JAMA Psychiatry} \bibinfo{volume}{75}
  (\bibinfo{year}{2018}) \bibinfo{pages}{1156--1172}.
\bibitem[{Varma and Simon(2006)}]{varma2006bias}
\bibinfo{author}{S.~Varma}, \bibinfo{author}{R.~Simon},
\newblock \bibinfo{title}{Bias in error estimation when using cross-validation
  for model selection},
\newblock \bibinfo{journal}{BMC Bioinformatics} \bibinfo{volume}{7}
  (\bibinfo{year}{2006}) \bibinfo{pages}{91}.
\bibitem[{Szegedy et~al.(2015)Szegedy, Liu, Jia, Sermanet, Reed, Anguelov,
  Erhan, Vanhoucke, and Rabinovich}]{Szegedy2014}
\bibinfo{author}{C.~Szegedy}, \bibinfo{author}{W.~Liu},
  \bibinfo{author}{Y.~Jia}, \bibinfo{author}{P.~Sermanet},
  \bibinfo{author}{S.~Reed}, \bibinfo{author}{D.~Anguelov},
  \bibinfo{author}{D.~Erhan}, \bibinfo{author}{V.~Vanhoucke},
  \bibinfo{author}{A.~Rabinovich},
\newblock \bibinfo{title}{Going deeper with convolutions},
\newblock in: \bibinfo{booktitle}{Proceedings of the IEEE Conference on
  Computer Vision and Pattern Recognition}, \bibinfo{year}{2015}, pp.
  \bibinfo{pages}{1--9}.
\bibitem[{Sutskever et~al.(2014)Sutskever, Vinyals, and Le}]{Sutskever2014}
\bibinfo{author}{I.~Sutskever}, \bibinfo{author}{O.~Vinyals},
  \bibinfo{author}{Q.~V. Le},
\newblock \bibinfo{title}{{Sequence to sequence learning with neural
  networks}},
\newblock in: \bibinfo{booktitle}{Advances in Neural Information Processing
  Systems}, \bibinfo{year}{2014}, pp. \bibinfo{pages}{3104--3112}.
\bibitem[{McCulloch and Pitts(1943)}]{Mcculloch1943}
\bibinfo{author}{W.~S. McCulloch}, \bibinfo{author}{W.~Pitts},
\newblock \bibinfo{title}{A logical calculus of the ideas immanent in nervous
  activity},
\newblock \bibinfo{journal}{The Bulletin of Mathematical Biophysics}
  \bibinfo{volume}{5} (\bibinfo{year}{1943}) \bibinfo{pages}{115--133}.
\bibitem[{Hopfield(1982)}]{hopfield1982neural}
\bibinfo{author}{J.~J. Hopfield},
\newblock \bibinfo{title}{Neural networks and physical systems with emergent
  collective computational abilities},
\newblock \bibinfo{journal}{Proceedings of the National Academy of Sciences}
  \bibinfo{volume}{79} (\bibinfo{year}{1982}) \bibinfo{pages}{2554--2558}.
\bibitem[{Hochreiter and Schmidhuber(1997)}]{hochreiter1997long}
\bibinfo{author}{S.~Hochreiter}, \bibinfo{author}{J.~Schmidhuber},
\newblock \bibinfo{title}{Long short-term memory},
\newblock \bibinfo{journal}{Neural Computation} \bibinfo{volume}{9}
  (\bibinfo{year}{1997}) \bibinfo{pages}{1735--1780}.
\bibitem[{Glorot et~al.(2011)Glorot, Bordes, and Bengio}]{Glorot2011}
\bibinfo{author}{X.~Glorot}, \bibinfo{author}{A.~Bordes},
  \bibinfo{author}{Y.~Bengio},
\newblock \bibinfo{title}{Deep sparse rectifier neural networks},
\newblock in: \bibinfo{booktitle}{Proceedings of the fourteenth International
  Conference on Artificial Intelligence and Statistics}, \bibinfo{year}{2011},
  pp. \bibinfo{pages}{315--323}.
\bibitem[{Cybenko(1989)}]{Cybenko1989}
\bibinfo{author}{G.~Cybenko},
\newblock \bibinfo{title}{Approximation by superpositions of a sigmoidal
  function},
\newblock \bibinfo{journal}{Mathematics of Control, Signals and Systems}
  \bibinfo{volume}{2} (\bibinfo{year}{1989}) \bibinfo{pages}{303--314}.
\bibitem[{Zhou(2020)}]{zhou2020universality}
\bibinfo{author}{D.-X. Zhou},
\newblock \bibinfo{title}{Universality of deep convolutional neural networks},
\newblock \bibinfo{journal}{Applied and computational harmonic analysis}
  \bibinfo{volume}{48} (\bibinfo{year}{2020}) \bibinfo{pages}{787--794}.
\bibitem[{Bengio(2009)}]{Bengio2009}
\bibinfo{author}{Y.~Bengio}, \bibinfo{title}{Learning deep architectures for
  AI}, \bibinfo{publisher}{Now Publishers Inc}, \bibinfo{year}{2009}.
\bibitem[{Bengio et~al.(2013)Bengio, Courville, and
  Vincent}]{bengio2013representation}
\bibinfo{author}{Y.~Bengio}, \bibinfo{author}{A.~Courville},
  \bibinfo{author}{P.~Vincent},
\newblock \bibinfo{title}{Representation learning: A review and new
  perspectives},
\newblock \bibinfo{journal}{IEEE Transactions on Pattern Analysis and Machine
  Intelligence} \bibinfo{volume}{35} (\bibinfo{year}{2013})
  \bibinfo{pages}{1798--1828}.
\bibitem[{Rumelhart et~al.(1986)Rumelhart, Hinton, and
  Williams}]{rumelhart1986}
\bibinfo{author}{D.~E. Rumelhart}, \bibinfo{author}{G.~E. Hinton},
  \bibinfo{author}{R.~J. Williams},
\newblock \bibinfo{title}{Learning representations by back-propagating errors},
\newblock \bibinfo{journal}{Nature} \bibinfo{volume}{323}
  (\bibinfo{year}{1986}) \bibinfo{pages}{533--536}.
\bibitem[{Paszke et~al.(2019)Paszke, Gross, Massa, Lerer, Bradbury, Chanan,
  Killeen, Lin, Gimelshein, Antiga et~al.}]{Paszke2019}
\bibinfo{author}{A.~Paszke}, \bibinfo{author}{S.~Gross},
  \bibinfo{author}{F.~Massa}, \bibinfo{author}{A.~Lerer},
  \bibinfo{author}{J.~Bradbury}, \bibinfo{author}{G.~Chanan},
  \bibinfo{author}{T.~Killeen}, \bibinfo{author}{Z.~Lin},
  \bibinfo{author}{N.~Gimelshein}, \bibinfo{author}{L.~Antiga}, et~al.,
\newblock \bibinfo{title}{Pytorch: An imperative style, high-performance deep
  learning library},
\newblock in: \bibinfo{booktitle}{Advances in Neural Information Processing
  Systems}, \bibinfo{year}{2019}, pp. \bibinfo{pages}{8026--8037}.
\bibitem[{Abadi et~al.(2016)Abadi, Agarwal, Barham, Brevdo, Chen, Citro,
  Corrado, Davis, Dean, Devin et~al.}]{tensorflow2015-whitepaper}
\bibinfo{author}{M.~Abadi}, \bibinfo{author}{A.~Agarwal},
  \bibinfo{author}{P.~Barham}, \bibinfo{author}{E.~Brevdo},
  \bibinfo{author}{Z.~Chen}, \bibinfo{author}{C.~Citro}, \bibinfo{author}{G.~S.
  Corrado}, \bibinfo{author}{A.~Davis}, \bibinfo{author}{J.~Dean},
  \bibinfo{author}{M.~Devin}, et~al.,
\newblock \bibinfo{title}{Tensorflow: Large-scale machine learning on
  heterogeneous distributed systems},
\newblock \bibinfo{journal}{arXiv preprint arXiv:1603.04467}
  (\bibinfo{year}{2016}).
\bibitem[{Bergstra et~al.(2010)Bergstra, Breuleux, Bastien, Lamblin, Pascanu,
  Desjardins, Turian, Warde-Farley, and Bengio}]{Bergstra2010}
\bibinfo{author}{J.~Bergstra}, \bibinfo{author}{O.~Breuleux},
  \bibinfo{author}{F.~Bastien}, \bibinfo{author}{P.~Lamblin},
  \bibinfo{author}{R.~Pascanu}, \bibinfo{author}{G.~Desjardins},
  \bibinfo{author}{J.~Turian}, \bibinfo{author}{D.~Warde-Farley},
  \bibinfo{author}{Y.~Bengio},
\newblock \bibinfo{title}{Theano: A {CPU} and {GPU} math compiler in python},
\newblock in: \bibinfo{booktitle}{Proceedings of the 9th Python in Science
  Conference}, volume~\bibinfo{volume}{1}, \bibinfo{year}{2010}, pp.
  \bibinfo{pages}{3--10}.
\bibitem[{Elsken et~al.(2019)Elsken, Metzen, and Hutter}]{Elsken2018}
\bibinfo{author}{T.~Elsken}, \bibinfo{author}{J.~H. Metzen},
  \bibinfo{author}{F.~Hutter},
\newblock \bibinfo{title}{Neural architecture search: A survey},
\newblock \bibinfo{journal}{Journal of Machine Learning Research}
  \bibinfo{volume}{20} (\bibinfo{year}{2019}) \bibinfo{pages}{1--21}.
\bibitem[{Srivastava et~al.(2014)Srivastava, Hinton, Krizhevsky, Sutskever, and
  Salakhutdinov}]{Srivastava2014}
\bibinfo{author}{N.~Srivastava}, \bibinfo{author}{G.~Hinton},
  \bibinfo{author}{A.~Krizhevsky}, \bibinfo{author}{I.~Sutskever},
  \bibinfo{author}{R.~Salakhutdinov},
\newblock \bibinfo{title}{Dropout: a simple way to prevent neural networks from
  overfitting},
\newblock \bibinfo{journal}{The Journal of Machine Learning Research}
  \bibinfo{volume}{15} (\bibinfo{year}{2014}) \bibinfo{pages}{1929--1958}.
\bibitem[{Krogh and Hertz(1992)}]{Krogh1992}
\bibinfo{author}{A.~Krogh}, \bibinfo{author}{J.~A. Hertz},
\newblock \bibinfo{title}{A simple weight decay can improve generalization},
\newblock in: \bibinfo{booktitle}{Advances in Neural Information Processing
  Systems}, \bibinfo{year}{1992}, pp. \bibinfo{pages}{950--957}.
\bibitem[{Li et~al.(2018)Li, Xu, Taylor, Studer, and Goldstein}]{Li2018}
\bibinfo{author}{H.~Li}, \bibinfo{author}{Z.~Xu}, \bibinfo{author}{G.~Taylor},
  \bibinfo{author}{C.~Studer}, \bibinfo{author}{T.~Goldstein},
\newblock \bibinfo{title}{Visualizing the loss landscape of neural nets},
\newblock in: \bibinfo{booktitle}{Advances in Neural Information Processing
  Systems}, \bibinfo{year}{2018}, pp. \bibinfo{pages}{6389--6399}.
\bibitem[{Choromanska et~al.(2015)Choromanska, Henaff, Mathieu, Arous, and
  LeCun}]{Choromanska2015}
\bibinfo{author}{A.~Choromanska}, \bibinfo{author}{M.~Henaff},
  \bibinfo{author}{M.~Mathieu}, \bibinfo{author}{G.~B. Arous},
  \bibinfo{author}{Y.~LeCun},
\newblock \bibinfo{title}{The loss surfaces of multilayer networks},
\newblock in: \bibinfo{booktitle}{Artificial Intelligence and Statistics},
  \bibinfo{year}{2015}, pp. \bibinfo{pages}{192--204}.
\bibitem[{LeCun and Bengio(1995)}]{LeCun1995}
\bibinfo{author}{Y.~LeCun}, \bibinfo{author}{Y.~Bengio},
\newblock \bibinfo{title}{{Convolutional networks for images, speech, and time
  series}},
\newblock \bibinfo{journal}{The Handbook of Brain Theory and Neural Networks}
  \bibinfo{volume}{3361} (\bibinfo{year}{1995}).
\bibitem[{LeCun et~al.(1989)LeCun, Boser, Denker, Henderson, Howard, Hubbard,
  and Jackel}]{LeCun1989}
\bibinfo{author}{Y.~LeCun}, \bibinfo{author}{B.~Boser}, \bibinfo{author}{J.~S.
  Denker}, \bibinfo{author}{D.~Henderson}, \bibinfo{author}{R.~E. Howard},
  \bibinfo{author}{W.~Hubbard}, \bibinfo{author}{L.~D. Jackel},
\newblock \bibinfo{title}{{Backpropagation Applied to Handwritten Zip Code
  Recognition}},
\newblock \bibinfo{journal}{Neural Computation} \bibinfo{volume}{1}
  (\bibinfo{year}{1989}) \bibinfo{pages}{541--551}.
\bibitem[{Huang et~al.(2017)Huang, Liu, van~der Maaten, and
  Weinberger}]{Huang2017DenselyNetworks}
\bibinfo{author}{G.~Huang}, \bibinfo{author}{Z.~Liu},
  \bibinfo{author}{L.~van~der Maaten}, \bibinfo{author}{K.~Q. Weinberger},
\newblock \bibinfo{title}{{Densely Connected Convolutional Networks}},
\newblock in: \bibinfo{booktitle}{CVPR 2017}, \bibinfo{year}{2017}, pp.
  \bibinfo{pages}{4700--4708}.
\bibitem[{He et~al.(2016)He, Zhang, Ren, and Sun}]{He2015DeepRecognition}
\bibinfo{author}{K.~He}, \bibinfo{author}{X.~Zhang}, \bibinfo{author}{S.~Ren},
  \bibinfo{author}{J.~Sun},
\newblock \bibinfo{title}{Deep residual learning for image recognition},
\newblock in: \bibinfo{booktitle}{Proceedings of the IEEE Conference on
  Computer Vision and Pattern Recognition}, \bibinfo{year}{2016}, pp.
  \bibinfo{pages}{770--778}.
\bibitem[{Simonyan and Zisserman(2014)}]{Simonyan2014a}
\bibinfo{author}{K.~Simonyan}, \bibinfo{author}{A.~Zisserman},
\newblock \bibinfo{title}{{Very Deep Convolutional Networks for Large-Scale
  Image Recognition}},
\newblock \bibinfo{journal}{arXiv preprint arXiv:1409.1556}
  (\bibinfo{year}{2014}).
\bibitem[{Leenings et~al.(2020)Leenings, Winter, Plagwitz, Holstein, Ernsting,
  Steenweg, Gebker, Sarink, Emden, Grotegerd et~al.}]{leenings2020photon}
\bibinfo{author}{R.~Leenings}, \bibinfo{author}{N.~R. Winter},
  \bibinfo{author}{L.~Plagwitz}, \bibinfo{author}{V.~Holstein},
  \bibinfo{author}{J.~Ernsting}, \bibinfo{author}{J.~Steenweg},
  \bibinfo{author}{J.~Gebker}, \bibinfo{author}{K.~Sarink},
  \bibinfo{author}{D.~Emden}, \bibinfo{author}{D.~Grotegerd}, et~al.,
\newblock \bibinfo{title}{{PHOTONAI}--a python {API} for rapid machine learning
  model development},
\newblock \bibinfo{journal}{arXiv preprint arXiv:2002.05426}
  (\bibinfo{year}{2020}).
\bibitem[{Beers et~al.(2020)Beers, Brown, Chang, Hoebel, Patel, Ly, Tolaney,
  Brastianos, Rosen, Gerstner et~al.}]{beers2020deepneuro}
\bibinfo{author}{A.~Beers}, \bibinfo{author}{J.~Brown},
  \bibinfo{author}{K.~Chang}, \bibinfo{author}{K.~Hoebel},
  \bibinfo{author}{J.~Patel}, \bibinfo{author}{K.~I. Ly},
  \bibinfo{author}{S.~M. Tolaney}, \bibinfo{author}{P.~Brastianos},
  \bibinfo{author}{B.~Rosen}, \bibinfo{author}{E.~R. Gerstner}, et~al.,
\newblock \bibinfo{title}{{D}eep{N}euro: an open-source deep learning toolbox
  for neuroimaging},
\newblock \bibinfo{journal}{Neuroinformatics}  (\bibinfo{year}{2020})
  \bibinfo{pages}{1--14}.
\bibitem[{Castelvecchi(2016)}]{Castelvecchi2016}
\bibinfo{author}{D.~Castelvecchi},
\newblock \bibinfo{title}{{Can we open the black box of AI?}},
\newblock \bibinfo{journal}{Nature} \bibinfo{volume}{538}
  (\bibinfo{year}{2016}) \bibinfo{pages}{20--23}.
\bibitem[{Lapuschkin et~al.(2019)Lapuschkin, W{\"{a}}ldchen, Binder, Montavon,
  Samek, and M{\"{u}}ller}]{Lapuschkin2019}
\bibinfo{author}{S.~Lapuschkin}, \bibinfo{author}{S.~W{\"{a}}ldchen},
  \bibinfo{author}{A.~Binder}, \bibinfo{author}{G.~Montavon},
  \bibinfo{author}{W.~Samek}, \bibinfo{author}{K.-R. M{\"{u}}ller},
\newblock \bibinfo{title}{{Unmasking Clever Hans Predictors and Assessing What
  Machines Really Learn}},
\newblock \bibinfo{journal}{Nature Communications} \bibinfo{volume}{10}
  (\bibinfo{year}{2019}) \bibinfo{pages}{1096}.
\bibitem[{Xiao et~al.(2018)Xiao, Choi, and Sun}]{Xiao2018}
\bibinfo{author}{C.~Xiao}, \bibinfo{author}{E.~Choi}, \bibinfo{author}{J.~Sun},
\newblock \bibinfo{title}{{Opportunities and challenges in developing deep
  learning models using electronic health records data: a systematic review}},
\newblock \bibinfo{journal}{Journal of the American Medical Informatics
  Association} \bibinfo{volume}{25} (\bibinfo{year}{2018})
  \bibinfo{pages}{1419--1428}.
\bibitem[{Ribeiro et~al.(2016)Ribeiro, Singh, and Guestrin}]{ribeiro2016should}
\bibinfo{author}{M.~T. Ribeiro}, \bibinfo{author}{S.~Singh},
  \bibinfo{author}{C.~Guestrin},
\newblock \bibinfo{title}{``why should i trust you?" explaining the predictions
  of any classifier},
\newblock in: \bibinfo{booktitle}{Proceedings of the 22nd ACM SIGKDD
  International Conference on Knowledge Discovery and Data Mining},
  \bibinfo{year}{2016}, pp. \bibinfo{pages}{1135--1144}.
\bibitem[{Lundberg and Lee(2017)}]{lundberg2017shap}
\bibinfo{author}{S.~M. Lundberg}, \bibinfo{author}{S.-I. Lee},
\newblock \bibinfo{title}{A unified approach to interpreting model
  predictions},
\newblock in: \bibinfo{editor}{I.~Guyon}, \bibinfo{editor}{U.~V. Luxburg},
  \bibinfo{editor}{S.~Bengio}, \bibinfo{editor}{H.~Wallach},
  \bibinfo{editor}{R.~Fergus}, \bibinfo{editor}{S.~Vishwanathan},
  \bibinfo{editor}{R.~Garnett} (Eds.), \bibinfo{booktitle}{Advances in Neural
  Information Processing Systems 30}, \bibinfo{publisher}{Curran Associates,
  Inc.}, \bibinfo{year}{2017}, pp. \bibinfo{pages}{4765--4774}.
\bibitem[{Koh and Liang(2017)}]{Koh2017}
\bibinfo{author}{P.~W. Koh}, \bibinfo{author}{P.~Liang},
\newblock \bibinfo{title}{Understanding black-box predictions via influence
  functions},
\newblock in: \bibinfo{booktitle}{International Conference on Machine
  Learning}, \bibinfo{year}{2017}, pp. \bibinfo{pages}{1885--1894}.
\bibitem[{Dosovitskiy and Brox(2016)}]{dosovitskiy2016generating}
\bibinfo{author}{A.~Dosovitskiy}, \bibinfo{author}{T.~Brox},
\newblock \bibinfo{title}{Generating images with perceptual similarity metrics
  based on deep networks},
\newblock in: \bibinfo{booktitle}{Advances in Neural Information Processing
  Systems}, \bibinfo{year}{2016}, pp. \bibinfo{pages}{658--666}.
\bibitem[{Hinton et~al.(2015)Hinton, Vinyals, and Dean}]{hinton2015distilling}
\bibinfo{author}{G.~Hinton}, \bibinfo{author}{O.~Vinyals},
  \bibinfo{author}{J.~Dean},
\newblock \bibinfo{title}{Distilling the knowledge in a neural network},
\newblock \bibinfo{journal}{arXiv preprint arXiv:1503.02531}
  (\bibinfo{year}{2015}).
\bibitem[{Chen et~al.(2018)Chen, Song, Wainwright, and
  Jordan}]{chen2018learning}
\bibinfo{author}{J.~Chen}, \bibinfo{author}{L.~Song},
  \bibinfo{author}{M.~Wainwright}, \bibinfo{author}{M.~Jordan},
\newblock \bibinfo{title}{Learning to explain: An information-theoretic
  perspective on model interpretation},
\newblock in: \bibinfo{booktitle}{International Conference on Machine
  Learning}, \bibinfo{year}{2018}, pp. \bibinfo{pages}{883--892}.
\bibitem[{Ancona et~al.(2017)Ancona, Ceolini, {\"O}ztireli, and
  Gross}]{ancona2018unifiedattribution}
\bibinfo{author}{M.~Ancona}, \bibinfo{author}{E.~Ceolini},
  \bibinfo{author}{C.~{\"O}ztireli}, \bibinfo{author}{M.~Gross},
\newblock \bibinfo{title}{A unified view of gradient-based attribution methods
  for deep neural networks},
\newblock in: \bibinfo{booktitle}{NIPS Workshop on Interpreting, Explaining and
  Visualizing Deep Learning-Now What? (NIPS 2017)}, \bibinfo{organization}{ETH
  Zurich}, \bibinfo{year}{2017}.
\bibitem[{Zeiler and Fergus(2014)}]{Zeiler2014}
\bibinfo{author}{M.~Zeiler}, \bibinfo{author}{R.~Fergus},
\newblock \bibinfo{title}{{Visualizing and Understanding Convolutional
  Networks}},
\newblock in: \bibinfo{editor}{D.~Fleet}, \bibinfo{editor}{T.~Pajdla},
  \bibinfo{editor}{B.~Schiele}, \bibinfo{editor}{T.~Tuytelaars} (Eds.),
  \bibinfo{booktitle}{Computer Vision – ECCV 2014}, volume
  \bibinfo{volume}{8689} of \textit{\bibinfo{series}{Lecture Notes in Computer
  Science}}, \bibinfo{publisher}{Springer International Publishing},
  \bibinfo{year}{2014}, pp. \bibinfo{pages}{818--833}.
\bibitem[{Rieke et~al.(2018)Rieke, Eitel, Weygandt, Haynes, and
  Ritter}]{Rieke2018}
\bibinfo{author}{J.~Rieke}, \bibinfo{author}{F.~Eitel},
  \bibinfo{author}{M.~Weygandt}, \bibinfo{author}{J.~D. Haynes},
  \bibinfo{author}{K.~Ritter},
\newblock \bibinfo{title}{{Visualizing convolutional networks for MRI-based
  diagnosis of Alzheimer's disease}},
\newblock in: \bibinfo{booktitle}{Lecture Notes in Computer Science (including
  subseries Lecture Notes in Artificial Intelligence and Lecture Notes in
  Bioinformatics)}, volume \bibinfo{volume}{11038 LNCS},
  \bibinfo{publisher}{Springer, Cham}, \bibinfo{year}{2018}, pp.
  \bibinfo{pages}{24--31}.
\bibitem[{Simonyan et~al.(2013)Simonyan, Vedaldi, and Zisserman}]{Simonyan2013}
\bibinfo{author}{K.~Simonyan}, \bibinfo{author}{A.~Vedaldi},
  \bibinfo{author}{A.~Zisserman},
\newblock \bibinfo{title}{{Deep inside convolutional networks: Visualising
  image classification models and saliency maps}},
\newblock \bibinfo{journal}{arXiv preprint arXiv:1312.6034}
  (\bibinfo{year}{2013}).
\bibitem[{Sundararajan et~al.(2017)Sundararajan, Taly, and
  Yan}]{sundararajan2017integratedgradients}
\bibinfo{author}{M.~Sundararajan}, \bibinfo{author}{A.~Taly},
  \bibinfo{author}{Q.~Yan},
\newblock \bibinfo{title}{Axiomatic attribution for deep networks},
\newblock in: \bibinfo{booktitle}{Proceedings of the 34th International
  Conference on Machine Learning-Volume 70}, \bibinfo{year}{2017}, pp.
  \bibinfo{pages}{3319--3328}.
\bibitem[{Shrikumar et~al.(2017)Shrikumar, Greenside, and
  Kundaje}]{shrikumar2017deeplift}
\bibinfo{author}{A.~Shrikumar}, \bibinfo{author}{P.~Greenside},
  \bibinfo{author}{A.~Kundaje},
\newblock \bibinfo{title}{Learning important features through propagating
  activation differences},
\newblock in: \bibinfo{booktitle}{Proceedings of the 34th International
  Conference on Machine Learning-Volume 70}, \bibinfo{year}{2017}, pp.
  \bibinfo{pages}{3145--3153}.
\bibitem[{Bach et~al.(2015)Bach, Binder, Montavon, Klauschen, M{\"{u}}ller, and
  Samek}]{Bach2015}
\bibinfo{author}{S.~Bach}, \bibinfo{author}{A.~Binder},
  \bibinfo{author}{G.~Montavon}, \bibinfo{author}{F.~Klauschen},
  \bibinfo{author}{K.-R. M{\"{u}}ller}, \bibinfo{author}{W.~Samek},
\newblock \bibinfo{title}{{On pixel-wise explanations for non-linear classifier
  decisions by layer-wise relevance propagation}},
\newblock \bibinfo{journal}{PLOS ONE} \bibinfo{volume}{10}
  (\bibinfo{year}{2015}) \bibinfo{pages}{e0130140}.
\bibitem[{Lee et~al.(2009)Lee, Grosse, Ranganath, and
  Ng}]{lee2009convolutional}
\bibinfo{author}{H.~Lee}, \bibinfo{author}{R.~Grosse},
  \bibinfo{author}{R.~Ranganath}, \bibinfo{author}{A.~Y. Ng},
\newblock \bibinfo{title}{Convolutional deep belief networks for scalable
  unsupervised learning of hierarchical representations},
\newblock in: \bibinfo{booktitle}{Proceedings of the 26th annual International
  Conference on Machine Learning}, \bibinfo{year}{2009}, pp.
  \bibinfo{pages}{609--616}.
\bibitem[{Plis et~al.(2014)Plis, Hjelm, Salakhutdinov, Allen, Bockholt, Long,
  Johnson, Paulsen, Turner, and Calhoun}]{plis2014deep}
\bibinfo{author}{S.~M. Plis}, \bibinfo{author}{D.~R. Hjelm},
  \bibinfo{author}{R.~Salakhutdinov}, \bibinfo{author}{E.~A. Allen},
  \bibinfo{author}{H.~J. Bockholt}, \bibinfo{author}{J.~D. Long},
  \bibinfo{author}{H.~J. Johnson}, \bibinfo{author}{J.~S. Paulsen},
  \bibinfo{author}{J.~A. Turner}, \bibinfo{author}{V.~D. Calhoun},
\newblock \bibinfo{title}{Deep learning for neuroimaging: a validation study},
\newblock \bibinfo{journal}{Frontiers in Neuroscience} \bibinfo{volume}{8}
  (\bibinfo{year}{2014}) \bibinfo{pages}{229}.
\bibitem[{Tishby and Zaslavsky(2015)}]{tishby2015deep}
\bibinfo{author}{N.~Tishby}, \bibinfo{author}{N.~Zaslavsky},
\newblock \bibinfo{title}{Deep learning and the information bottleneck
  principle},
\newblock in: \bibinfo{booktitle}{2015 IEEE Information Theory Workshop (ITW)},
  \bibinfo{organization}{IEEE}, \bibinfo{year}{2015}, pp.
  \bibinfo{pages}{1--5}.
\bibitem[{Kramer(1991)}]{kramer1991nonlinear}
\bibinfo{author}{M.~A. Kramer},
\newblock \bibinfo{title}{Nonlinear principal component analysis using
  autoassociative neural networks},
\newblock \bibinfo{journal}{AIChE Journal} \bibinfo{volume}{37}
  (\bibinfo{year}{1991}) \bibinfo{pages}{233--243}.
\bibitem[{Mikolov et~al.(2013)Mikolov, Chen, Corrado, and
  Dean}]{mikolov2013efficient}
\bibinfo{author}{T.~Mikolov}, \bibinfo{author}{K.~Chen},
  \bibinfo{author}{G.~Corrado}, \bibinfo{author}{J.~Dean},
\newblock \bibinfo{title}{Efficient estimation of word representations in
  vector space},
\newblock \bibinfo{journal}{arXiv preprint arXiv:1301.3781}
  (\bibinfo{year}{2013}).
\bibitem[{Zhang et~al.(2017)Zhang, Isola, and Efros}]{zhang2017split}
\bibinfo{author}{R.~Zhang}, \bibinfo{author}{P.~Isola}, \bibinfo{author}{A.~A.
  Efros},
\newblock \bibinfo{title}{Split-brain autoencoders: Unsupervised learning by
  cross-channel prediction},
\newblock in: \bibinfo{booktitle}{Proceedings of the IEEE Conference on
  Computer Vision and Pattern Recognition}, \bibinfo{year}{2017}, pp.
  \bibinfo{pages}{1058--1067}.
\bibitem[{Balakrishnan et~al.(2019)Balakrishnan, Zhao, Sabuncu, Guttag, and
  Dalca}]{balakrishnan2019voxelmorph}
\bibinfo{author}{G.~Balakrishnan}, \bibinfo{author}{A.~Zhao},
  \bibinfo{author}{M.~R. Sabuncu}, \bibinfo{author}{J.~Guttag},
  \bibinfo{author}{A.~V. Dalca},
\newblock \bibinfo{title}{Voxelmorph: a learning framework for deformable
  medical image registration},
\newblock \bibinfo{journal}{IEEE transactions on medical imaging}
  \bibinfo{volume}{38} (\bibinfo{year}{2019}) \bibinfo{pages}{1788--1800}.
\bibitem[{Kingma and Welling(2013)}]{kingma2013auto}
\bibinfo{author}{D.~P. Kingma}, \bibinfo{author}{M.~Welling},
\newblock \bibinfo{title}{Auto-encoding variational bayes},
\newblock \bibinfo{journal}{arXiv preprint arXiv:1312.6114}
  (\bibinfo{year}{2013}).
\bibitem[{Cheung et~al.(2014)Cheung, Livezey, Bansal, and
  Olshausen}]{cheung2014discovering}
\bibinfo{author}{B.~Cheung}, \bibinfo{author}{J.~A. Livezey},
  \bibinfo{author}{A.~K. Bansal}, \bibinfo{author}{B.~A. Olshausen},
\newblock \bibinfo{title}{Discovering hidden factors of variation in deep
  networks},
\newblock \bibinfo{journal}{arXiv preprint arXiv:1412.6583}
  (\bibinfo{year}{2014}).
\bibitem[{He et~al.(2020)He, An, Feng, Bzdok, Holmes, Eickhoff, and
  Yeo}]{he2020meta}
\bibinfo{author}{T.~He}, \bibinfo{author}{L.~An}, \bibinfo{author}{J.~Feng},
  \bibinfo{author}{D.~Bzdok}, \bibinfo{author}{A.~J. Holmes},
  \bibinfo{author}{S.~B. Eickhoff}, \bibinfo{author}{B.~T.~T. Yeo},
\newblock \bibinfo{title}{Meta-matching: a simple framework to translate
  phenotypic predictive models from big to small data},
\newblock \bibinfo{journal}{bioRxiv}  (\bibinfo{year}{2020}).
\bibitem[{Schulz et~al.(2017)Schulz, Varoquaux, Gramfort, Thirion, and
  Bzdok}]{semi2017}
\bibinfo{author}{M.-A. Schulz}, \bibinfo{author}{G.~Varoquaux},
  \bibinfo{author}{A.~Gramfort}, \bibinfo{author}{B.~Thirion},
  \bibinfo{author}{D.~Bzdok},
\newblock \bibinfo{title}{Label scarcity in biomedicine: Data-rich latent
  factor discovery enhances phenotype prediction},
\newblock \bibinfo{journal}{NIPS - Machine Learning for Health Workshop}
  (\bibinfo{year}{2017}).
\bibitem[{Tan et~al.(2018)Tan, Sun, Kong, Zhang, Yang, and Liu}]{tan2018survey}
\bibinfo{author}{C.~Tan}, \bibinfo{author}{F.~Sun}, \bibinfo{author}{T.~Kong},
  \bibinfo{author}{W.~Zhang}, \bibinfo{author}{C.~Yang},
  \bibinfo{author}{C.~Liu},
\newblock \bibinfo{title}{A survey on deep transfer learning},
\newblock in: \bibinfo{booktitle}{International Conference on Artificial Neural
  Networks}, \bibinfo{organization}{Springer}, \bibinfo{year}{2018}, pp.
  \bibinfo{pages}{270--279}.
\bibitem[{Maqsood et~al.(2019)Maqsood, Nazir, Khan, Aadil, Jamal, Mehmood, and
  Song}]{maqsood2019transfer}
\bibinfo{author}{M.~Maqsood}, \bibinfo{author}{F.~Nazir},
  \bibinfo{author}{U.~Khan}, \bibinfo{author}{F.~Aadil},
  \bibinfo{author}{H.~Jamal}, \bibinfo{author}{I.~Mehmood},
  \bibinfo{author}{O.-y. Song},
\newblock \bibinfo{title}{Transfer learning assisted classification and
  detection of alzheimer’s disease stages using 3{D} {MRI} scans},
\newblock \bibinfo{journal}{Sensors} \bibinfo{volume}{19}
  (\bibinfo{year}{2019}) \bibinfo{pages}{2645}.
\bibitem[{{Hon} and {Khan}(2017)}]{hon2017transfer}
\bibinfo{author}{M.~{Hon}}, \bibinfo{author}{N.~M. {Khan}},
\newblock \bibinfo{title}{Towards alzheimer's disease classification through
  transfer learning},
\newblock in: \bibinfo{booktitle}{2017 IEEE International Conference on
  Bioinformatics and Biomedicine (BIBM)}, \bibinfo{year}{2017}, pp.
  \bibinfo{pages}{1166--1169}.
\bibitem[{Ghafoorian et~al.(2017)Ghafoorian, Mehrtash, Kapur, Karssemeijer,
  Marchiori, Pesteie, Guttmann, de~Leeuw, Tempany, van Ginneken, Fedorov,
  Abolmaesumi, Platel, and Wells}]{Ghafoorian2017TransferLearning}
\bibinfo{author}{M.~Ghafoorian}, \bibinfo{author}{A.~Mehrtash},
  \bibinfo{author}{T.~Kapur}, \bibinfo{author}{N.~Karssemeijer},
  \bibinfo{author}{E.~Marchiori}, \bibinfo{author}{M.~Pesteie},
  \bibinfo{author}{C.~R.~G. Guttmann}, \bibinfo{author}{F.-E. de~Leeuw},
  \bibinfo{author}{C.~M. Tempany}, \bibinfo{author}{B.~van Ginneken},
  \bibinfo{author}{A.~Fedorov}, \bibinfo{author}{P.~Abolmaesumi},
  \bibinfo{author}{B.~Platel}, \bibinfo{author}{W.~M. Wells},
\newblock \bibinfo{title}{{Transfer Learning for Domain Adaptation in MRI:
  Application in Brain Lesion Segmentation}},
\newblock in: \bibinfo{editor}{M.~Descoteaux}, \bibinfo{editor}{L.~Maier-Hein},
  \bibinfo{editor}{A.~Franz}, \bibinfo{editor}{P.~Jannin},
  \bibinfo{editor}{D.~L. Collins}, \bibinfo{editor}{S.~Duchesne} (Eds.),
  \bibinfo{booktitle}{Medical Image Computing and Computer Assisted
  Intervention - MICCAI 2017}, \bibinfo{publisher}{Springer International
  Publishing}, \bibinfo{address}{Cham}, \bibinfo{year}{2017}, pp.
  \bibinfo{pages}{516--524}.
\bibitem[{Kamnitsas et~al.(2017)Kamnitsas, Baumgartner, Ledig, Newcombe,
  Simpson, Kane, Menon, Nori, Criminisi, Rueckert
  et~al.}]{kamnitsas2017unsupervised}
\bibinfo{author}{K.~Kamnitsas}, \bibinfo{author}{C.~Baumgartner},
  \bibinfo{author}{C.~Ledig}, \bibinfo{author}{V.~Newcombe},
  \bibinfo{author}{J.~Simpson}, \bibinfo{author}{A.~Kane},
  \bibinfo{author}{D.~Menon}, \bibinfo{author}{A.~Nori},
  \bibinfo{author}{A.~Criminisi}, \bibinfo{author}{D.~Rueckert}, et~al.,
\newblock \bibinfo{title}{Unsupervised domain adaptation in brain lesion
  segmentation with adversarial networks},
\newblock in: \bibinfo{booktitle}{International Conference on Information
  Processing in Medical Imaging}, \bibinfo{organization}{Springer},
  \bibinfo{year}{2017}, pp. \bibinfo{pages}{597--609}.
\bibitem[{Akkus et~al.(2017)Akkus, Galimzianova, Hoogi, Rubin, and
  Erickson}]{akkus2017deep}
\bibinfo{author}{Z.~Akkus}, \bibinfo{author}{A.~Galimzianova},
  \bibinfo{author}{A.~Hoogi}, \bibinfo{author}{D.~L. Rubin},
  \bibinfo{author}{B.~J. Erickson},
\newblock \bibinfo{title}{Deep learning for brain {MRI} segmentation: state of
  the art and future directions},
\newblock \bibinfo{journal}{Journal of Digital Imaging} \bibinfo{volume}{30}
  (\bibinfo{year}{2017}) \bibinfo{pages}{449--459}.
\bibitem[{Valindria et~al.(2017)Valindria, Lavdas, Bai, Kamnitsas, Aboagye,
  Rockall, Rueckert, and Glocker}]{valindria2017reverse}
\bibinfo{author}{V.~V. Valindria}, \bibinfo{author}{I.~Lavdas},
  \bibinfo{author}{W.~Bai}, \bibinfo{author}{K.~Kamnitsas},
  \bibinfo{author}{E.~O. Aboagye}, \bibinfo{author}{A.~G. Rockall},
  \bibinfo{author}{D.~Rueckert}, \bibinfo{author}{B.~Glocker},
\newblock \bibinfo{title}{Reverse classification accuracy: predicting
  segmentation performance in the absence of ground truth},
\newblock \bibinfo{journal}{IEEE Transactions on Medical Imaging}
  \bibinfo{volume}{36} (\bibinfo{year}{2017}) \bibinfo{pages}{1597--1606}.
\bibitem[{Gupta et~al.(2013)Gupta, Ayhan, and Maida}]{Gupta2013}
\bibinfo{author}{A.~Gupta}, \bibinfo{author}{M.~Ayhan},
  \bibinfo{author}{A.~Maida},
\newblock \bibinfo{title}{{Natural {I}mage {B}ases to {R}epresent
  {N}euroimaging {D}ata}},
\newblock in: \bibinfo{editor}{S.~Dasgupta}, \bibinfo{editor}{D.~Mcallester}
  (Eds.), \bibinfo{booktitle}{Proceedings of the 30th International Conference
  on Machine Learning (ICML-13)}, volume~\bibinfo{volume}{28},
  \bibinfo{publisher}{JMLR Workshop and Conference Proceedings},
  \bibinfo{year}{2013}, pp. \bibinfo{pages}{987--994}.
\bibitem[{Hosseini-Asl et~al.(2018)Hosseini-Asl, Ghazal, Mahmoud, Aslantas,
  Shalaby, Casanova, Barnes, Gimel'farb, Keynton, and
  El-Baz}]{hosseini2018alzheimer}
\bibinfo{author}{E.~Hosseini-Asl}, \bibinfo{author}{M.~Ghazal},
  \bibinfo{author}{A.~Mahmoud}, \bibinfo{author}{A.~Aslantas},
  \bibinfo{author}{A.~Shalaby}, \bibinfo{author}{M.~Casanova},
  \bibinfo{author}{G.~Barnes}, \bibinfo{author}{G.~Gimel'farb},
  \bibinfo{author}{R.~Keynton}, \bibinfo{author}{A.~El-Baz},
\newblock \bibinfo{title}{Alzheimer's disease diagnostics by a 3{D} deeply
  supervised adaptable convolutional network.},
\newblock \bibinfo{journal}{Frontiers in Bioscience (Landmark edition)}
  \bibinfo{volume}{23} (\bibinfo{year}{2018}) \bibinfo{pages}{584}.
\bibitem[{Geman et~al.(1992)Geman, Bienenstock, and Doursat}]{geman1992neural}
\bibinfo{author}{S.~Geman}, \bibinfo{author}{E.~Bienenstock},
  \bibinfo{author}{R.~Doursat},
\newblock \bibinfo{title}{Neural networks and the bias/variance dilemma},
\newblock \bibinfo{journal}{Neural Computation} \bibinfo{volume}{4}
  (\bibinfo{year}{1992}) \bibinfo{pages}{1--58}.
\bibitem[{Battaglia et~al.(2018)Battaglia, Hamrick, Bapst, Sanchez-Gonzalez,
  Zambaldi, Malinowski, Tacchetti, Raposo, Santoro, Faulkner
  et~al.}]{battaglia2018relational}
\bibinfo{author}{P.~W. Battaglia}, \bibinfo{author}{J.~B. Hamrick},
  \bibinfo{author}{V.~Bapst}, \bibinfo{author}{A.~Sanchez-Gonzalez},
  \bibinfo{author}{V.~Zambaldi}, \bibinfo{author}{M.~Malinowski},
  \bibinfo{author}{A.~Tacchetti}, \bibinfo{author}{D.~Raposo},
  \bibinfo{author}{A.~Santoro}, \bibinfo{author}{R.~Faulkner}, et~al.,
\newblock \bibinfo{title}{Relational inductive biases, deep learning, and graph
  networks},
\newblock \bibinfo{journal}{arXiv preprint arXiv:1806.01261}
  (\bibinfo{year}{2018}).
\bibitem[{Yu et~al.(2019)Yu, Si, Hu, and Zhang}]{yu2019review}
\bibinfo{author}{Y.~Yu}, \bibinfo{author}{X.~Si}, \bibinfo{author}{C.~Hu},
  \bibinfo{author}{J.~Zhang},
\newblock \bibinfo{title}{A review of recurrent neural networks: {LSTM} cells
  and network architectures},
\newblock \bibinfo{journal}{Neural Computation} \bibinfo{volume}{31}
  (\bibinfo{year}{2019}) \bibinfo{pages}{1235--1270}.
\bibitem[{Wu et~al.(2020)Wu, Pan, Chen, Long, Zhang, and
  Philip}]{wu2020comprehensive}
\bibinfo{author}{Z.~Wu}, \bibinfo{author}{S.~Pan}, \bibinfo{author}{F.~Chen},
  \bibinfo{author}{G.~Long}, \bibinfo{author}{C.~Zhang}, \bibinfo{author}{S.~Y.
  Philip},
\newblock \bibinfo{title}{A comprehensive survey on graph neural networks},
\newblock \bibinfo{journal}{IEEE Transactions on Neural Networks and Learning
  Systems}  (\bibinfo{year}{2020}).
\bibitem[{Kawahara et~al.(2017)Kawahara, Brown, Miller, Booth, Chau, Grunau,
  Zwicker, and Hamarneh}]{kawahara2017brainnetcnn}
\bibinfo{author}{J.~Kawahara}, \bibinfo{author}{C.~J. Brown},
  \bibinfo{author}{S.~P. Miller}, \bibinfo{author}{B.~G. Booth},
  \bibinfo{author}{V.~Chau}, \bibinfo{author}{R.~E. Grunau},
  \bibinfo{author}{J.~G. Zwicker}, \bibinfo{author}{G.~Hamarneh},
\newblock \bibinfo{title}{Brain{N}et{CNN}: Convolutional neural networks for
  brain networks; towards predicting neurodevelopment},
\newblock \bibinfo{journal}{NeuroImage} \bibinfo{volume}{146}
  (\bibinfo{year}{2017}) \bibinfo{pages}{1038--1049}.
\bibitem[{Eitel et~al.(2020)Eitel, Albrecht, Weygandt, Paul, and
  Ritter}]{eitel2020harnessing}
\bibinfo{author}{F.~Eitel}, \bibinfo{author}{J.~P. Albrecht},
  \bibinfo{author}{M.~Weygandt}, \bibinfo{author}{F.~Paul},
  \bibinfo{author}{K.~Ritter},
\newblock \bibinfo{title}{Harnessing spatial homogeneity of neuroimaging data:
  patch individual filter layers for {CNN}s},
\newblock \bibinfo{journal}{arXiv preprint arXiv:2007.11899}
  (\bibinfo{year}{2020}).
\bibitem[{Defferrard et~al.(2016)Defferrard, Bresson, and
  Vandergheynst}]{defferrard2016convolutional}
\bibinfo{author}{M.~Defferrard}, \bibinfo{author}{X.~Bresson},
  \bibinfo{author}{P.~Vandergheynst},
\newblock \bibinfo{title}{Convolutional neural networks on graphs with fast
  localized spectral filtering},
\newblock in: \bibinfo{booktitle}{Advances in Neural Information Processing
  Systems}, \bibinfo{year}{2016}, pp. \bibinfo{pages}{3844--3852}.
\bibitem[{Bronstein et~al.(2017)Bronstein, Bruna, LeCun, Szlam, and
  Vandergheynst}]{bronstein2017geometric}
\bibinfo{author}{M.~M. Bronstein}, \bibinfo{author}{J.~Bruna},
  \bibinfo{author}{Y.~LeCun}, \bibinfo{author}{A.~Szlam},
  \bibinfo{author}{P.~Vandergheynst},
\newblock \bibinfo{title}{Geometric deep learning: going beyond euclidean
  data},
\newblock \bibinfo{journal}{IEEE Signal Processing Magazine}
  \bibinfo{volume}{34} (\bibinfo{year}{2017}) \bibinfo{pages}{18--42}.
\bibitem[{Kipf and Welling(2016)}]{kipf2016semi}
\bibinfo{author}{T.~N. Kipf}, \bibinfo{author}{M.~Welling},
\newblock \bibinfo{title}{Semi-supervised classification with graph
  convolutional networks},
\newblock \bibinfo{journal}{arXiv preprint arXiv:1609.02907}
  (\bibinfo{year}{2016}).
\bibitem[{Parisot et~al.(2017)Parisot, Ktena, Ferrante, Lee, Moreno, Glocker,
  and Rueckert}]{parisot2017spectral}
\bibinfo{author}{S.~Parisot}, \bibinfo{author}{S.~I. Ktena},
  \bibinfo{author}{E.~Ferrante}, \bibinfo{author}{M.~Lee},
  \bibinfo{author}{R.~G. Moreno}, \bibinfo{author}{B.~Glocker},
  \bibinfo{author}{D.~Rueckert},
\newblock \bibinfo{title}{Spectral graph convolutions for population-based
  disease prediction},
\newblock in: \bibinfo{booktitle}{International Conference on Medical Image
  Computing and Computer Assisted Intervention},
  \bibinfo{organization}{Springer}, \bibinfo{year}{2017}, pp.
  \bibinfo{pages}{177--185}.
\bibitem[{Parisot et~al.(2018)Parisot, Ktena, Ferrante, Lee, Guerrero, Glocker,
  and Rueckert}]{parisot2018disease}
\bibinfo{author}{S.~Parisot}, \bibinfo{author}{S.~I. Ktena},
  \bibinfo{author}{E.~Ferrante}, \bibinfo{author}{M.~Lee},
  \bibinfo{author}{R.~Guerrero}, \bibinfo{author}{B.~Glocker},
  \bibinfo{author}{D.~Rueckert},
\newblock \bibinfo{title}{Disease prediction using graph convolutional
  networks: Application to autism spectrum disorder and {A}lzheimer’s
  disease},
\newblock \bibinfo{journal}{Medical Image Analysis} \bibinfo{volume}{48}
  (\bibinfo{year}{2018}) \bibinfo{pages}{117--130}.
\bibitem[{Jo et~al.(2019)Jo, Nho, and Saykin}]{jo2019deep}
\bibinfo{author}{T.~Jo}, \bibinfo{author}{K.~Nho}, \bibinfo{author}{A.~J.
  Saykin},
\newblock \bibinfo{title}{Deep learning in {A}lzheimer’s disease: diagnostic
  classification and prognostic prediction using neuroimaging data},
\newblock \bibinfo{journal}{Frontiers in Aging Neuroscience}
  \bibinfo{volume}{11} (\bibinfo{year}{2019}) \bibinfo{pages}{220}.
\bibitem[{Patel et~al.(2015)Patel, Andreescu, Price, Edelman, Reynolds~III, and
  Aizenstein}]{patel2015responsepred}
\bibinfo{author}{M.~J. Patel}, \bibinfo{author}{C.~Andreescu},
  \bibinfo{author}{J.~C. Price}, \bibinfo{author}{K.~L. Edelman},
  \bibinfo{author}{C.~F. Reynolds~III}, \bibinfo{author}{H.~J. Aizenstein},
\newblock \bibinfo{title}{Machine learning approaches for integrating clinical
  and imaging features in late-life depression classification and response
  prediction},
\newblock \bibinfo{journal}{International Journal of Geriatric Psychiatry}
  \bibinfo{volume}{30} (\bibinfo{year}{2015}) \bibinfo{pages}{1056--1067}.
\bibitem[{Schnack et~al.(2014)Schnack, Nieuwenhuis, van Haren, Abramovic,
  Scheewe, Brouwer, Pol, and Kahn}]{schnack2014can}
\bibinfo{author}{H.~G. Schnack}, \bibinfo{author}{M.~Nieuwenhuis},
  \bibinfo{author}{N.~E. van Haren}, \bibinfo{author}{L.~Abramovic},
  \bibinfo{author}{T.~W. Scheewe}, \bibinfo{author}{R.~M. Brouwer},
  \bibinfo{author}{H.~E.~H. Pol}, \bibinfo{author}{R.~S. Kahn},
\newblock \bibinfo{title}{Can structural {MRI} aid in clinical classification?
  {A} machine learning study in two independent samples of patients with
  schizophrenia, bipolar disorder and healthy subjects},
\newblock \bibinfo{journal}{NeuroImage} \bibinfo{volume}{84}
  (\bibinfo{year}{2014}) \bibinfo{pages}{299--306}.
\bibitem[{Talpalaru et~al.(2019)Talpalaru, Bhagwat, Devenyi, Lepage, and
  Chakravarty}]{talpalaru2019identifying}
\bibinfo{author}{A.~Talpalaru}, \bibinfo{author}{N.~Bhagwat},
  \bibinfo{author}{G.~A. Devenyi}, \bibinfo{author}{M.~Lepage},
  \bibinfo{author}{M.~M. Chakravarty},
\newblock \bibinfo{title}{Identifying schizophrenia subgroups using clustering
  and supervised learning},
\newblock \bibinfo{journal}{Schizophrenia Research} \bibinfo{volume}{214}
  (\bibinfo{year}{2019}) \bibinfo{pages}{51--59}.
\bibitem[{{Liu} et~al.(2015){Liu}, {Liu}, {Cai}, {Che}, {Pujol}, {Kikinis},
  {Feng}, {Fulham}, and {ADNI}}]{liu2015multilabelmulticlass}
\bibinfo{author}{S.~{Liu}}, \bibinfo{author}{S.~{Liu}},
  \bibinfo{author}{W.~{Cai}}, \bibinfo{author}{H.~{Che}},
  \bibinfo{author}{S.~{Pujol}}, \bibinfo{author}{R.~{Kikinis}},
  \bibinfo{author}{D.~{Feng}}, \bibinfo{author}{M.~J. {Fulham}},
  \bibinfo{author}{{ADNI}},
\newblock \bibinfo{title}{Multimodal neuroimaging feature learning for
  multiclass diagnosis of {A}lzheimer's disease},
\newblock \bibinfo{journal}{IEEE Transactions on Biomedical Engineering}
  \bibinfo{volume}{62} (\bibinfo{year}{2015}) \bibinfo{pages}{1132--1140}.
\bibitem[{Bzdok et~al.(2020)Bzdok, Varoquaux, and
  Steyerberg}]{bzdok2020prediction}
\bibinfo{author}{D.~Bzdok}, \bibinfo{author}{G.~Varoquaux},
  \bibinfo{author}{E.~W. Steyerberg},
\newblock \bibinfo{title}{{Prediction, Not Association, Paves the Road to
  Precision Medicine}},
\newblock \bibinfo{journal}{JAMA Psychiatry}  (\bibinfo{year}{2020}).
\bibitem[{Saposnik et~al.(2016)Saposnik, Redelmeier, Ruff, and
  Tobler}]{saposnik2016cognitive}
\bibinfo{author}{G.~Saposnik}, \bibinfo{author}{D.~Redelmeier},
  \bibinfo{author}{C.~C. Ruff}, \bibinfo{author}{P.~N. Tobler},
\newblock \bibinfo{title}{Cognitive biases associated with medical decisions: a
  systematic review},
\newblock \bibinfo{journal}{BMC medical informatics and decision making}
  \bibinfo{volume}{16} (\bibinfo{year}{2016}) \bibinfo{pages}{138}.
\bibitem[{Hamberg(2008)}]{hamberg2008genderbias}
\bibinfo{author}{K.~Hamberg},
\newblock \bibinfo{title}{Gender bias in medicine},
\newblock \bibinfo{journal}{Women's Health} \bibinfo{volume}{4}
  (\bibinfo{year}{2008}) \bibinfo{pages}{237--243}.
\bibitem[{Habes et~al.(2016)Habes, Erus, Toledo, Zhang, Bryan, Launer, Rosseel,
  Janowitz, Doshi, Van~der Auwera et~al.}]{habes2016white}
\bibinfo{author}{M.~Habes}, \bibinfo{author}{G.~Erus}, \bibinfo{author}{J.~B.
  Toledo}, \bibinfo{author}{T.~Zhang}, \bibinfo{author}{N.~Bryan},
  \bibinfo{author}{L.~J. Launer}, \bibinfo{author}{Y.~Rosseel},
  \bibinfo{author}{D.~Janowitz}, \bibinfo{author}{J.~Doshi},
  \bibinfo{author}{S.~Van~der Auwera}, et~al.,
\newblock \bibinfo{title}{White matter hyperintensities and imaging patterns of
  brain ageing in the general population},
\newblock \bibinfo{journal}{Brain} \bibinfo{volume}{139} (\bibinfo{year}{2016})
  \bibinfo{pages}{1164--1179}.
\bibitem[{Pagnozzi et~al.(2019)Pagnozzi, Fripp, and Rose}]{PAGNOZZI2019116018}
\bibinfo{author}{A.~M. Pagnozzi}, \bibinfo{author}{J.~Fripp},
  \bibinfo{author}{S.~E. Rose},
\newblock \bibinfo{title}{Quantifying deep grey matter atrophy using automated
  segmentation approaches: A systematic review of structural {MRI} studies},
\newblock \bibinfo{journal}{NeuroImage} \bibinfo{volume}{201}
  (\bibinfo{year}{2019}) \bibinfo{pages}{116018}.
\bibitem[{Suk et~al.(2014)Suk, Lee, and Shen}]{Suk2014}
\bibinfo{author}{H.-I. Suk}, \bibinfo{author}{S.-W. Lee},
  \bibinfo{author}{D.~Shen},
\newblock \bibinfo{title}{{Hierarchical feature representation and multimodal
  fusion with deep learning for {AD}/{MCI} diagnosis}},
\newblock \bibinfo{journal}{NeuroImage} \bibinfo{volume}{101}
  (\bibinfo{year}{2014}) \bibinfo{pages}{569--582}.
\bibitem[{Alaa and van~der Schaar(2019)}]{alaa2019attentive}
\bibinfo{author}{A.~M. Alaa}, \bibinfo{author}{M.~van~der Schaar},
\newblock \bibinfo{title}{Attentive state-space modeling of disease
  progression},
\newblock in: \bibinfo{booktitle}{Advances in Neural Information Processing
  Systems}, \bibinfo{year}{2019}, pp. \bibinfo{pages}{11338--11348}.
\bibitem[{Thompson et~al.(2020)Thompson, Jahanshad, Ching, Salminen,
  Thomopoulos, Bright, Baune, Bertol{\'\i}n, Bralten, Bruin
  et~al.}]{thompson2020enigma}
\bibinfo{author}{P.~M. Thompson}, \bibinfo{author}{N.~Jahanshad},
  \bibinfo{author}{C.~R. Ching}, \bibinfo{author}{L.~E. Salminen},
  \bibinfo{author}{S.~I. Thomopoulos}, \bibinfo{author}{J.~Bright},
  \bibinfo{author}{B.~T. Baune}, \bibinfo{author}{S.~Bertol{\'\i}n},
  \bibinfo{author}{J.~Bralten}, \bibinfo{author}{W.~B. Bruin}, et~al.,
\newblock \bibinfo{title}{{ENIGMA} and global neuroscience: A decade of
  large-scale studies of the brain in health and disease across more than 40
  countries},
\newblock \bibinfo{journal}{Translational Psychiatry} \bibinfo{volume}{10}
  (\bibinfo{year}{2020}) \bibinfo{pages}{1--28}.
\bibitem[{Lehman(2000)}]{lehman2000diagnostic}
\bibinfo{author}{J.~F. Lehman},
\newblock \bibinfo{title}{The diagnostic and statistical manual of mental
  disorders}  (\bibinfo{year}{2000}).
\bibitem[{Karalunas et~al.(2014)Karalunas, Fair, Musser, Aykes, Iyer, and
  Nigg}]{Karalunas2014}
\bibinfo{author}{S.~L. Karalunas}, \bibinfo{author}{D.~Fair},
  \bibinfo{author}{E.~D. Musser}, \bibinfo{author}{K.~Aykes},
  \bibinfo{author}{S.~P. Iyer}, \bibinfo{author}{J.~T. Nigg},
\newblock \bibinfo{title}{{Subtyping attention-deficit/hyperactivity disorder
  using temperament dimensions: toward biologically based nosologic criteria}},
\newblock \bibinfo{journal}{JAMA Psychiatry} \bibinfo{volume}{71}
  (\bibinfo{year}{2014}) \bibinfo{pages}{1015--1024}.
\bibitem[{Tamminga et~al.(2014)Tamminga, Pearlson, Keshavan, Sweeney, Clementz,
  and Thaker}]{tamminga2014bipolar}
\bibinfo{author}{C.~A. Tamminga}, \bibinfo{author}{G.~Pearlson},
  \bibinfo{author}{M.~Keshavan}, \bibinfo{author}{J.~Sweeney},
  \bibinfo{author}{B.~Clementz}, \bibinfo{author}{G.~Thaker},
\newblock \bibinfo{title}{Bipolar and schizophrenia network for intermediate
  phenotypes: outcomes across the psychosis continuum},
\newblock \bibinfo{journal}{Schizophrenia Bulletin} \bibinfo{volume}{40}
  (\bibinfo{year}{2014}) \bibinfo{pages}{S131--S137}.
\bibitem[{L{\"o}we et~al.(2008)L{\"o}we, Spitzer, Williams, Mussell,
  Schellberg, and Kroenke}]{lowe2008depression}
\bibinfo{author}{B.~L{\"o}we}, \bibinfo{author}{R.~L. Spitzer},
  \bibinfo{author}{J.~B. Williams}, \bibinfo{author}{M.~Mussell},
  \bibinfo{author}{D.~Schellberg}, \bibinfo{author}{K.~Kroenke},
\newblock \bibinfo{title}{Depression, anxiety and somatization in primary care:
  syndrome overlap and functional impairment},
\newblock \bibinfo{journal}{General Hospital Psychiatry} \bibinfo{volume}{30}
  (\bibinfo{year}{2008}) \bibinfo{pages}{191--199}.
\bibitem[{Insel et~al.(2010)Insel, Cuthbert, Garvey, Heinssen, Pine, Quinn,
  Sanislow, and Wang}]{Insel2010}
\bibinfo{author}{T.~Insel}, \bibinfo{author}{B.~Cuthbert},
  \bibinfo{author}{M.~Garvey}, \bibinfo{author}{R.~Heinssen},
  \bibinfo{author}{D.~S. Pine}, \bibinfo{author}{K.~Quinn},
  \bibinfo{author}{C.~Sanislow}, \bibinfo{author}{P.~Wang},
\newblock \bibinfo{title}{{Research Domain Criteria (RDoC): Toward a New
  Classification Framework for Research on Mental Disorders}},
\newblock \bibinfo{journal}{American Journal of Psychiatry}
  \bibinfo{volume}{167} (\bibinfo{year}{2010}) \bibinfo{pages}{748--751}.
\bibitem[{Karim et~al.(2020)Karim, Beyan, Zappa, Costa, Rebholz-Schuhmann,
  Cochez, and Decker}]{karim2020deep}
\bibinfo{author}{M.~R. Karim}, \bibinfo{author}{O.~Beyan},
  \bibinfo{author}{A.~Zappa}, \bibinfo{author}{I.~G. Costa},
  \bibinfo{author}{D.~Rebholz-Schuhmann}, \bibinfo{author}{M.~Cochez},
  \bibinfo{author}{S.~Decker},
\newblock \bibinfo{title}{Deep learning-based clustering approaches for
  bioinformatics},
\newblock \bibinfo{journal}{Briefings in Bioinformatics}
  (\bibinfo{year}{2020}).
\bibitem[{Mirzaei and Adeli(2018)}]{mirzaei2018segmentation}
\bibinfo{author}{G.~Mirzaei}, \bibinfo{author}{H.~Adeli},
\newblock \bibinfo{title}{Segmentation and clustering in brain {MRI} imaging},
\newblock \bibinfo{journal}{Reviews in the Neurosciences} \bibinfo{volume}{30}
  (\bibinfo{year}{2018}) \bibinfo{pages}{31--44}.
\bibitem[{Schulz et~al.(2020)Schulz, Chapman-Rounds, Verma, Bzdok, and
  Georgatzis}]{schulz2020inferring}
\bibinfo{author}{M.-A. Schulz}, \bibinfo{author}{M.~Chapman-Rounds},
  \bibinfo{author}{M.~Verma}, \bibinfo{author}{D.~Bzdok},
  \bibinfo{author}{K.~Georgatzis},
\newblock \bibinfo{title}{Inferring disease subtypes from clusters in
  explanation space},
\newblock \bibinfo{journal}{Scientific Reports} \bibinfo{volume}{10}
  (\bibinfo{year}{2020}) \bibinfo{pages}{1--6}.
\bibitem[{Panta et~al.(2016)Panta, Wang, Fries, Kalyanam, Speer, Banich, Kiehl,
  King, Milham, Wager et~al.}]{panta2016tool}
\bibinfo{author}{S.~R. Panta}, \bibinfo{author}{R.~Wang},
  \bibinfo{author}{J.~Fries}, \bibinfo{author}{R.~Kalyanam},
  \bibinfo{author}{N.~Speer}, \bibinfo{author}{M.~Banich},
  \bibinfo{author}{K.~Kiehl}, \bibinfo{author}{M.~King},
  \bibinfo{author}{M.~Milham}, \bibinfo{author}{T.~D. Wager}, et~al.,
\newblock \bibinfo{title}{A tool for interactive data visualization:
  application to over 10,000 brain imaging and phantom {MRI} data sets},
\newblock \bibinfo{journal}{Frontiers in Neuroinformatics} \bibinfo{volume}{10}
  (\bibinfo{year}{2016}) \bibinfo{pages}{9}.
\bibitem[{Drysdale et~al.(2017)Drysdale, Grosenick, Downar, Dunlop, Mansouri,
  Meng, Fetcho, Zebley, Oathes, Etkin, Schatzberg, Sudheimer, Keller, Mayberg,
  Gunning, Alexopoulos, Fox, Pascual-Leone, Voss, Casey, Dubin, and
  Liston}]{Drysdale2017Resting-stateDepression}
\bibinfo{author}{A.~T. Drysdale}, \bibinfo{author}{L.~Grosenick},
  \bibinfo{author}{J.~Downar}, \bibinfo{author}{K.~Dunlop},
  \bibinfo{author}{F.~Mansouri}, \bibinfo{author}{Y.~Meng},
  \bibinfo{author}{R.~N. Fetcho}, \bibinfo{author}{B.~Zebley},
  \bibinfo{author}{D.~J. Oathes}, \bibinfo{author}{A.~Etkin},
  \bibinfo{author}{A.~F. Schatzberg}, \bibinfo{author}{K.~Sudheimer},
  \bibinfo{author}{J.~Keller}, \bibinfo{author}{H.~S. Mayberg},
  \bibinfo{author}{F.~M. Gunning}, \bibinfo{author}{G.~S. Alexopoulos},
  \bibinfo{author}{M.~D. Fox}, \bibinfo{author}{A.~Pascual-Leone},
  \bibinfo{author}{H.~U. Voss}, \bibinfo{author}{B.~Casey},
  \bibinfo{author}{M.~J. Dubin}, \bibinfo{author}{C.~Liston},
\newblock \bibinfo{title}{{Resting-state connectivity biomarkers define
  neurophysiological subtypes of depression}},
\newblock \bibinfo{journal}{Nature Medicine} \bibinfo{volume}{23}
  (\bibinfo{year}{2017}) \bibinfo{pages}{28--38}.
\bibitem[{Hardoon et~al.(2004)Hardoon, Szedmak, and
  Shawe-Taylor}]{hardoon2004canonical}
\bibinfo{author}{D.~R. Hardoon}, \bibinfo{author}{S.~Szedmak},
  \bibinfo{author}{J.~Shawe-Taylor},
\newblock \bibinfo{title}{Canonical correlation analysis: An overview with
  application to learning methods},
\newblock \bibinfo{journal}{Neural Computation} \bibinfo{volume}{16}
  (\bibinfo{year}{2004}) \bibinfo{pages}{2639--2664}.
\bibitem[{Dinga et~al.(2019)Dinga, Schmaal, Penninx, van Tol, Veltman, van
  Velzen, Mennes, van~der Wee, and Marquand}]{dinga2019evaluating}
\bibinfo{author}{R.~Dinga}, \bibinfo{author}{L.~Schmaal},
  \bibinfo{author}{B.~W. Penninx}, \bibinfo{author}{M.~J. van Tol},
  \bibinfo{author}{D.~J. Veltman}, \bibinfo{author}{L.~van Velzen},
  \bibinfo{author}{M.~Mennes}, \bibinfo{author}{N.~J. van~der Wee},
  \bibinfo{author}{A.~F. Marquand},
\newblock \bibinfo{title}{Evaluating the evidence for biotypes of depression:
  Methodological replication and extension of},
\newblock \bibinfo{journal}{NeuroImage: Clinical} \bibinfo{volume}{22}
  (\bibinfo{year}{2019}) \bibinfo{pages}{101796}.
\bibitem[{Andrew et~al.(2013)Andrew, Arora, Bilmes, and
  Livescu}]{andrew2013deep}
\bibinfo{author}{G.~Andrew}, \bibinfo{author}{R.~Arora},
  \bibinfo{author}{J.~Bilmes}, \bibinfo{author}{K.~Livescu},
\newblock \bibinfo{title}{Deep canonical correlation analysis},
\newblock in: \bibinfo{booktitle}{International Conference on Machine
  Learning}, \bibinfo{organization}{PMLR}, \bibinfo{year}{2013}, pp.
  \bibinfo{pages}{1247--1255}.
\bibitem[{Marquand et~al.(2019)Marquand, Kia, Zabihi, Wolfers, Buitelaar, and
  Beckmann}]{marquand2019conceptualizing}
\bibinfo{author}{A.~F. Marquand}, \bibinfo{author}{S.~M. Kia},
  \bibinfo{author}{M.~Zabihi}, \bibinfo{author}{T.~Wolfers},
  \bibinfo{author}{J.~K. Buitelaar}, \bibinfo{author}{C.~F. Beckmann},
\newblock \bibinfo{title}{Conceptualizing mental disorders as deviations from
  normative functioning},
\newblock \bibinfo{journal}{Molecular Psychiatry} \bibinfo{volume}{24}
  (\bibinfo{year}{2019}) \bibinfo{pages}{1415--1424}.
\bibitem[{Pinaya et~al.(2019)Pinaya, Mechelli, and Sato}]{pinaya2019using}
\bibinfo{author}{W.~H. Pinaya}, \bibinfo{author}{A.~Mechelli},
  \bibinfo{author}{J.~R. Sato},
\newblock \bibinfo{title}{Using deep autoencoders to identify abnormal brain
  structural patterns in neuropsychiatric disorders: A large-scale multi-sample
  study},
\newblock \bibinfo{journal}{Human Brain Mapping} \bibinfo{volume}{40}
  (\bibinfo{year}{2019}) \bibinfo{pages}{944--954}.
\bibitem[{Kia and Marquand(2019)}]{kia2019neural}
\bibinfo{author}{S.~M. Kia}, \bibinfo{author}{A.~F. Marquand},
\newblock \bibinfo{title}{Neural processes mixed-effect models for deep
  normative modeling of clinical neuroimaging data},
\newblock in: \bibinfo{booktitle}{International Conference on Medical Imaging
  with Deep Learning}, \bibinfo{organization}{PMLR}, \bibinfo{year}{2019}, pp.
  \bibinfo{pages}{297--314}.
\bibitem[{Pinaya et~al.(2020)Pinaya, Scarpazza, Garcia-Dias, Vieira, Baecker,
  da~Costa, Redolfi, Frisoni, Pievani, Calhoun et~al.}]{pinaya2020normative}
\bibinfo{author}{W.~H. Pinaya}, \bibinfo{author}{C.~Scarpazza},
  \bibinfo{author}{R.~Garcia-Dias}, \bibinfo{author}{S.~Vieira},
  \bibinfo{author}{L.~Baecker}, \bibinfo{author}{P.~F. da~Costa},
  \bibinfo{author}{A.~Redolfi}, \bibinfo{author}{G.~B. Frisoni},
  \bibinfo{author}{M.~Pievani}, \bibinfo{author}{V.~D. Calhoun}, et~al.,
\newblock \bibinfo{title}{Normative modelling using deep autoencoders: a
  multi-cohort study on mild cognitive impairment and {A}lzheimer's disease},
\newblock \bibinfo{journal}{bioRxiv}  (\bibinfo{year}{2020}).
\bibitem[{Wolfers et~al.(2018)Wolfers, Doan, Kaufmann, Aln{\ae}s, Moberget,
  Agartz, Buitelaar, Ueland, Melle, Franke et~al.}]{wolfers2018mapping}
\bibinfo{author}{T.~Wolfers}, \bibinfo{author}{N.~T. Doan},
  \bibinfo{author}{T.~Kaufmann}, \bibinfo{author}{D.~Aln{\ae}s},
  \bibinfo{author}{T.~Moberget}, \bibinfo{author}{I.~Agartz},
  \bibinfo{author}{J.~K. Buitelaar}, \bibinfo{author}{T.~Ueland},
  \bibinfo{author}{I.~Melle}, \bibinfo{author}{B.~Franke}, et~al.,
\newblock \bibinfo{title}{Mapping the heterogeneous phenotype of schizophrenia
  and bipolar disorder using normative models},
\newblock \bibinfo{journal}{JAMA Psychiatry} \bibinfo{volume}{75}
  (\bibinfo{year}{2018}) \bibinfo{pages}{1146--1155}.
\bibitem[{Achterberg et~al.(2019)Achterberg, Sørensen, Wolters, Niessen,
  Vernooij, Ikram, Nielsen, and {de Bruijne}}]{ACHTERBERG2019hippocampal}
\bibinfo{author}{H.~C. Achterberg}, \bibinfo{author}{L.~Sørensen},
  \bibinfo{author}{F.~J. Wolters}, \bibinfo{author}{W.~J. Niessen},
  \bibinfo{author}{M.~W. Vernooij}, \bibinfo{author}{M.~A. Ikram},
  \bibinfo{author}{M.~Nielsen}, \bibinfo{author}{M.~{de Bruijne}},
\newblock \bibinfo{title}{The value of hippocampal volume, shape, and texture
  for 11-year prediction of dementia: a population-based study},
\newblock \bibinfo{journal}{Neurobiology of Aging} \bibinfo{volume}{81}
  (\bibinfo{year}{2019}) \bibinfo{pages}{58 -- 66}.
\bibitem[{Dyrba et~al.(2020)Dyrba, Pallath, and Marzban}]{Dyrba2020vis}
\bibinfo{author}{M.~Dyrba}, \bibinfo{author}{A.~H. Pallath},
  \bibinfo{author}{E.~N. Marzban},
\newblock \bibinfo{title}{Comparison of {CNN} visualization methods to aid
  model interpretability for detecting {A}lzheimer's disease},
\newblock in: \bibinfo{editor}{T.~Tolxdorff}, \bibinfo{editor}{T.~M. Deserno},
  \bibinfo{editor}{H.~Handels}, \bibinfo{editor}{A.~Maier},
  \bibinfo{editor}{K.~H. Maier-Hein}, \bibinfo{editor}{C.~Palm} (Eds.),
  \bibinfo{booktitle}{Bildverarbeitung f{\"u}r die Medizin 2020},
  \bibinfo{publisher}{Springer Fachmedien Wiesbaden},
  \bibinfo{address}{Wiesbaden}, \bibinfo{year}{2020}, pp.
  \bibinfo{pages}{307--312}.
\bibitem[{Jo et~al.(2020)Jo, Nho, Risacher, and Saykin}]{Jo2020taubetvis}
\bibinfo{author}{T.~Jo}, \bibinfo{author}{K.~Nho}, \bibinfo{author}{S.~L.
  Risacher}, \bibinfo{author}{A.~J. Saykin},
\newblock \bibinfo{title}{Deep learning detection of informative features in
  tau {PET} for {A}lzheimer{\textquoteright}s disease classification},
\newblock \bibinfo{journal}{bioRxiv}  (\bibinfo{year}{2020}).
\bibitem[{Whelan et~al.(2014)Whelan, Watts, Orr, Althoff, Artiges,
  Banaschewski, Barker, Bokde, B{\"u}chel, Carvalho
  et~al.}]{whelan2014neuropsychosocial}
\bibinfo{author}{R.~Whelan}, \bibinfo{author}{R.~Watts}, \bibinfo{author}{C.~A.
  Orr}, \bibinfo{author}{R.~R. Althoff}, \bibinfo{author}{E.~Artiges},
  \bibinfo{author}{T.~Banaschewski}, \bibinfo{author}{G.~J. Barker},
  \bibinfo{author}{A.~L. Bokde}, \bibinfo{author}{C.~B{\"u}chel},
  \bibinfo{author}{F.~M. Carvalho}, et~al.,
\newblock \bibinfo{title}{Neuropsychosocial profiles of current and future
  adolescent alcohol misusers},
\newblock \bibinfo{journal}{Nature} \bibinfo{volume}{512}
  (\bibinfo{year}{2014}) \bibinfo{pages}{185--189}.
\bibitem[{Pettersson-Yeo et~al.(2013)Pettersson-Yeo, Benetti, Marquand,
  Dell‘Acqua, Williams, Allen, Prata, McGuire, and
  Mechelli}]{pettersson2013multimodal}
\bibinfo{author}{W.~Pettersson-Yeo}, \bibinfo{author}{S.~Benetti},
  \bibinfo{author}{A.~F. Marquand}, \bibinfo{author}{F.~Dell‘Acqua},
  \bibinfo{author}{S.~C.~R. Williams}, \bibinfo{author}{P.~Allen},
  \bibinfo{author}{D.~Prata}, \bibinfo{author}{P.~McGuire},
  \bibinfo{author}{A.~Mechelli},
\newblock \bibinfo{title}{Using genetic, cognitive and multi-modal neuroimaging
  data to identify ultra-high-risk and first-episode psychosis at the
  individual level},
\newblock \bibinfo{journal}{Psychological Medicine} \bibinfo{volume}{43}
  (\bibinfo{year}{2013}) \bibinfo{pages}{2547–2562}.
\bibitem[{Nie et~al.(2016)Nie, Zhang, Adeli, Liu, and Shen}]{nie2016multimodal}
\bibinfo{author}{D.~Nie}, \bibinfo{author}{H.~Zhang},
  \bibinfo{author}{E.~Adeli}, \bibinfo{author}{L.~Liu},
  \bibinfo{author}{D.~Shen},
\newblock \bibinfo{title}{3{D} deep learning for multi-modal imaging-guided
  survival time prediction of brain tumor patients},
\newblock in: \bibinfo{editor}{S.~Ourselin}, \bibinfo{editor}{L.~Joskowicz},
  \bibinfo{editor}{M.~R. Sabuncu}, \bibinfo{editor}{G.~Unal},
  \bibinfo{editor}{W.~Wells} (Eds.), \bibinfo{booktitle}{Medical Image
  Computing and Computer Assisted Intervention -- MICCAI 2016},
  \bibinfo{publisher}{Springer International Publishing},
  \bibinfo{address}{Cham}, \bibinfo{year}{2016}, pp. \bibinfo{pages}{212--220}.
\bibitem[{Maglanoc et~al.(2020)Maglanoc, Kaufmann, Jonassen, Hilland, Beck,
  Landr{\o}, and Westlye}]{maglanoc2020multimodal}
\bibinfo{author}{L.~A. Maglanoc}, \bibinfo{author}{T.~Kaufmann},
  \bibinfo{author}{R.~Jonassen}, \bibinfo{author}{E.~Hilland},
  \bibinfo{author}{D.~Beck}, \bibinfo{author}{N.~I. Landr{\o}},
  \bibinfo{author}{L.~T. Westlye},
\newblock \bibinfo{title}{Multimodal fusion of structural and functional brain
  imaging in depression using linked independent component analysis},
\newblock \bibinfo{journal}{Human Brain Mapping} \bibinfo{volume}{41}
  (\bibinfo{year}{2020}) \bibinfo{pages}{241--255}.
\bibitem[{Kriegeskorte et~al.(2009)Kriegeskorte, Simmons, Bellgowan, and
  Baker}]{Kriegeskorte2009}
\bibinfo{author}{N.~Kriegeskorte}, \bibinfo{author}{W.~K. Simmons},
  \bibinfo{author}{P.~S.~F. Bellgowan}, \bibinfo{author}{C.~I. Baker},
\newblock \bibinfo{title}{{Circular analysis in systems neuroscience: the
  dangers of double dipping}},
\newblock \bibinfo{journal}{Nature Neuroscience} \bibinfo{volume}{12}
  (\bibinfo{year}{2009}) \bibinfo{pages}{535--540}.
\bibitem[{Dwyer et~al.(2018)Dwyer, Falkai, and Koutsouleris}]{dwyer2018machine}
\bibinfo{author}{D.~B. Dwyer}, \bibinfo{author}{P.~Falkai},
  \bibinfo{author}{N.~Koutsouleris},
\newblock \bibinfo{title}{Machine learning approaches for clinical psychology
  and psychiatry},
\newblock \bibinfo{journal}{Annual Review of Clinical Psychology}
  \bibinfo{volume}{14} (\bibinfo{year}{2018}) \bibinfo{pages}{91--118}.
\bibitem[{Kl{\"o}ppel et~al.(2012)Kl{\"o}ppel, Abdulkadir, Jack~Jr,
  Koutsouleris, Mour{\~a}o-Miranda, and Vemuri}]{kloppel2012diagnostic}
\bibinfo{author}{S.~Kl{\"o}ppel}, \bibinfo{author}{A.~Abdulkadir},
  \bibinfo{author}{C.~R. Jack~Jr}, \bibinfo{author}{N.~Koutsouleris},
  \bibinfo{author}{J.~Mour{\~a}o-Miranda}, \bibinfo{author}{P.~Vemuri},
\newblock \bibinfo{title}{Diagnostic neuroimaging across diseases},
\newblock \bibinfo{journal}{NeuroImage} \bibinfo{volume}{61}
  (\bibinfo{year}{2012}) \bibinfo{pages}{457--463}.
\bibitem[{Koutsouleris et~al.(2009)Koutsouleris, Meisenzahl, Davatzikos,
  Bottlender, Frodl, Scheuerecker, Schmitt, Zetzsche, Decker, Reiser
  et~al.}]{koutsouleris2009use}
\bibinfo{author}{N.~Koutsouleris}, \bibinfo{author}{E.~M. Meisenzahl},
  \bibinfo{author}{C.~Davatzikos}, \bibinfo{author}{R.~Bottlender},
  \bibinfo{author}{T.~Frodl}, \bibinfo{author}{J.~Scheuerecker},
  \bibinfo{author}{G.~Schmitt}, \bibinfo{author}{T.~Zetzsche},
  \bibinfo{author}{P.~Decker}, \bibinfo{author}{M.~Reiser}, et~al.,
\newblock \bibinfo{title}{Use of neuroanatomical pattern classification to
  identify subjects in at-risk mental states of psychosis and predict disease
  transition},
\newblock \bibinfo{journal}{Archives of General Psychiatry}
  \bibinfo{volume}{66} (\bibinfo{year}{2009}) \bibinfo{pages}{700--712}.
\bibitem[{Klöppel et~al.(2008)Klöppel, Stonnington, Barnes, Chen, Chu, Good,
  Mader, Mitchell, Patel, Roberts, Fox, Jack, Ashburner, and {and Richard S. J.
  Frackowiak}}]{Kloppel2008a}
\bibinfo{author}{S.~Klöppel}, \bibinfo{author}{C.~M. Stonnington},
  \bibinfo{author}{J.~Barnes}, \bibinfo{author}{F.~Chen},
  \bibinfo{author}{C.~Chu}, \bibinfo{author}{C.~D. Good},
  \bibinfo{author}{I.~Mader}, \bibinfo{author}{L.~A. Mitchell},
  \bibinfo{author}{A.~C. Patel}, \bibinfo{author}{C.~C. Roberts},
  \bibinfo{author}{N.~C. Fox}, \bibinfo{author}{C.~R. Jack},
  \bibinfo{author}{J.~Ashburner}, \bibinfo{author}{{and Richard S. J.
  Frackowiak}},
\newblock \bibinfo{title}{{Accuracy of dementia diagnosis - a direct comparison
  between radiologists and a computerized method}},
\newblock \bibinfo{journal}{Brain} \bibinfo{volume}{131} (\bibinfo{year}{2008})
  \bibinfo{pages}{2969--2974}.
\bibitem[{Weygandt et~al.(2012)Weygandt, Blecker, Sch{\"{a}}fer, Hackmack,
  Haynes, Vaitl, Stark, and Schienle}]{Weygandt2012}
\bibinfo{author}{M.~Weygandt}, \bibinfo{author}{C.~R. Blecker},
  \bibinfo{author}{A.~Sch{\"{a}}fer}, \bibinfo{author}{K.~Hackmack},
  \bibinfo{author}{J.-D. Haynes}, \bibinfo{author}{D.~Vaitl},
  \bibinfo{author}{R.~Stark}, \bibinfo{author}{A.~Schienle},
\newblock \bibinfo{title}{{fMRI pattern recognition in obsessive-compulsive
  disorder}},
\newblock \bibinfo{journal}{NeuroImage} \bibinfo{volume}{60}
  (\bibinfo{year}{2012}) \bibinfo{pages}{1186--1193}.
\bibitem[{Varoquaux(2018)}]{VAROQUAUX2018samplesizes}
\bibinfo{author}{G.~Varoquaux},
\newblock \bibinfo{title}{Cross-validation failure: Small sample sizes lead to
  large error bars},
\newblock \bibinfo{journal}{NeuroImage} \bibinfo{volume}{180}
  (\bibinfo{year}{2018}) \bibinfo{pages}{68 -- 77}. \bibinfo{note}{New Advances
  in Encoding and Decoding of Brain Signals}.
\bibitem[{Arbabshirani et~al.(2017)Arbabshirani, Plis, Sui, and
  Calhoun}]{arbabshirani2017single}
\bibinfo{author}{M.~R. Arbabshirani}, \bibinfo{author}{S.~Plis},
  \bibinfo{author}{J.~Sui}, \bibinfo{author}{V.~D. Calhoun},
\newblock \bibinfo{title}{Single subject prediction of brain disorders in
  neuroimaging: Promises and pitfalls},
\newblock \bibinfo{journal}{NeuroImage} \bibinfo{volume}{145}
  (\bibinfo{year}{2017}) \bibinfo{pages}{137--165}.
\bibitem[{Flint et~al.(2019)Flint, Cearns, Opel, Redlich, Mehler, Emden,
  Winter, Leenings, Eickhoff, Kircher et~al.}]{flint2019systematic}
\bibinfo{author}{C.~Flint}, \bibinfo{author}{M.~Cearns},
  \bibinfo{author}{N.~Opel}, \bibinfo{author}{R.~Redlich},
  \bibinfo{author}{D.~Mehler}, \bibinfo{author}{D.~Emden},
  \bibinfo{author}{N.~R. Winter}, \bibinfo{author}{R.~Leenings},
  \bibinfo{author}{S.~B. Eickhoff}, \bibinfo{author}{T.~Kircher}, et~al.,
\newblock \bibinfo{title}{Systematic overestimation of machine learning
  performance in neuroimaging studies of depression},
\newblock \bibinfo{journal}{arXiv preprint arXiv:1912.06686}
  (\bibinfo{year}{2019}).
\bibitem[{Koppe et~al.(2020)Koppe, Meyer-Lindenberg, and
  Durstewitz}]{koppe2020deep}
\bibinfo{author}{G.~Koppe}, \bibinfo{author}{A.~Meyer-Lindenberg},
  \bibinfo{author}{D.~Durstewitz},
\newblock \bibinfo{title}{Deep learning for small and big data in psychiatry},
\newblock \bibinfo{journal}{Neuropsychopharmacology}  (\bibinfo{year}{2020})
  \bibinfo{pages}{1--17}.
\bibitem[{Bhogal et~al.(2013)Bhogal, Mahoney, Graeme-Baker, Roy, Shah, Fraioli,
  Cowley, and Jager}]{Bhogal2013}
\bibinfo{author}{P.~Bhogal}, \bibinfo{author}{C.~Mahoney},
  \bibinfo{author}{S.~Graeme-Baker}, \bibinfo{author}{A.~Roy},
  \bibinfo{author}{S.~Shah}, \bibinfo{author}{F.~Fraioli},
  \bibinfo{author}{P.~Cowley}, \bibinfo{author}{H.~R. Jager},
\newblock \bibinfo{title}{The common dementias: {a} pictorial review},
\newblock \bibinfo{journal}{European Radiology} \bibinfo{volume}{23}
  (\bibinfo{year}{2013}) \bibinfo{pages}{3405--3417}.
\bibitem[{Weiner et~al.(2013)Weiner, Veitch, Aisen, Beckett, Cairns, Green,
  Harvey, Jack, Jagust, Liu, Morris, Petersen, Saykin, Schmidt, Shaw, Shen,
  Siuciak, Soares, Toga, and Trojanowski}]{Weiner2013}
\bibinfo{author}{M.~W. Weiner}, \bibinfo{author}{D.~P. Veitch},
  \bibinfo{author}{P.~S. Aisen}, \bibinfo{author}{L.~A. Beckett},
  \bibinfo{author}{N.~J. Cairns}, \bibinfo{author}{R.~C. Green},
  \bibinfo{author}{D.~Harvey}, \bibinfo{author}{C.~R. Jack},
  \bibinfo{author}{W.~Jagust}, \bibinfo{author}{E.~Liu}, \bibinfo{author}{J.~C.
  Morris}, \bibinfo{author}{R.~C. Petersen}, \bibinfo{author}{A.~J. Saykin},
  \bibinfo{author}{M.~E. Schmidt}, \bibinfo{author}{L.~Shaw},
  \bibinfo{author}{L.~Shen}, \bibinfo{author}{J.~A. Siuciak},
  \bibinfo{author}{H.~Soares}, \bibinfo{author}{A.~W. Toga},
  \bibinfo{author}{J.~Q. Trojanowski},
\newblock \bibinfo{title}{{The {A}lzheimer's {D}isease {N}euroimaging
  {I}nitiative: A review of papers published since its inception}},
\newblock \bibinfo{journal}{Alzheimer's {\&} Dementia} \bibinfo{volume}{9}
  (\bibinfo{year}{2013}) \bibinfo{pages}{e111 -- e194}.
\bibitem[{Payan and Montana(2015)}]{Payan2015}
\bibinfo{author}{A.~Payan}, \bibinfo{author}{G.~Montana},
\newblock \bibinfo{title}{Predicting {A}lzheimer's disease: a neuroimaging
  study with 3{D} convolutional neural networks},
\newblock \bibinfo{journal}{arXiv preprint arXiv:1502.02506}
  (\bibinfo{year}{2015}).
\bibitem[{Korolev et~al.(2017)Korolev, Safiullin, Belyaev, and
  Dodonova}]{korolev2017residual}
\bibinfo{author}{S.~Korolev}, \bibinfo{author}{A.~Safiullin},
  \bibinfo{author}{M.~Belyaev}, \bibinfo{author}{Y.~Dodonova},
\newblock \bibinfo{title}{Residual and plain convolutional neural networks for
  3{D} brain {MRI} classification},
\newblock in: \bibinfo{booktitle}{2017 IEEE 14th International Symposium on
  Biomedical Imaging (ISBI 2017)}, \bibinfo{organization}{IEEE},
  \bibinfo{year}{2017}, pp. \bibinfo{pages}{835--838}.
\bibitem[{Wood et~al.(2019)Wood, Cole, and Booth}]{wood2019neuro}
\bibinfo{author}{D.~Wood}, \bibinfo{author}{J.~Cole},
  \bibinfo{author}{T.~Booth},
\newblock \bibinfo{title}{{NEURO}-{DRAM}: a 3{D} recurrent visual attention
  model for interpretable neuroimaging classification},
\newblock \bibinfo{journal}{arXiv preprint arXiv:1910.04721}
  (\bibinfo{year}{2019}).
\bibitem[{Shenton et~al.(2001)Shenton, Dickey, Frumin, and
  McCarley}]{shenton2001review}
\bibinfo{author}{M.~E. Shenton}, \bibinfo{author}{C.~C. Dickey},
  \bibinfo{author}{M.~Frumin}, \bibinfo{author}{R.~W. McCarley},
\newblock \bibinfo{title}{A review of {MRI} findings in schizophrenia},
\newblock \bibinfo{journal}{Schizophrenia Research} \bibinfo{volume}{49}
  (\bibinfo{year}{2001}) \bibinfo{pages}{1--52}.
\bibitem[{Narr et~al.(2004)Narr, Bilder, Kim, Thompson, Szeszko, Robinson,
  Luders, and Toga}]{Narr2004}
\bibinfo{author}{K.~L. Narr}, \bibinfo{author}{R.~M. Bilder},
  \bibinfo{author}{S.~Kim}, \bibinfo{author}{P.~M. Thompson},
  \bibinfo{author}{P.~Szeszko}, \bibinfo{author}{D.~Robinson},
  \bibinfo{author}{E.~Luders}, \bibinfo{author}{A.~W. Toga},
\newblock \bibinfo{title}{{Abnormal gyral complexity in first-episode
  schizophrenia}},
\newblock \bibinfo{journal}{Biological Psychiatry} \bibinfo{volume}{55}
  (\bibinfo{year}{2004}) \bibinfo{pages}{859--867}.
\bibitem[{van~der Meer et~al.(2010)van~der Meer, Costafreda, Aleman, and
  David}]{van2010self}
\bibinfo{author}{L.~van~der Meer}, \bibinfo{author}{S.~Costafreda},
  \bibinfo{author}{A.~Aleman}, \bibinfo{author}{A.~S. David},
\newblock \bibinfo{title}{Self-reflection and the brain: a theoretical review
  and meta-analysis of neuroimaging studies with implications for
  schizophrenia},
\newblock \bibinfo{journal}{Neuroscience \& Biobehavioral Reviews}
  \bibinfo{volume}{34} (\bibinfo{year}{2010}) \bibinfo{pages}{935--946}.
\bibitem[{Kircher et~al.(2019)Kircher, W{\"o}hr, Nenadic, Schwarting, Schratt,
  Alferink, Culmsee, Garn, Hahn, M{\"u}ller-Myhsok
  et~al.}]{kircher2019neurobiology}
\bibinfo{author}{T.~Kircher}, \bibinfo{author}{M.~W{\"o}hr},
  \bibinfo{author}{I.~Nenadic}, \bibinfo{author}{R.~Schwarting},
  \bibinfo{author}{G.~Schratt}, \bibinfo{author}{J.~Alferink},
  \bibinfo{author}{C.~Culmsee}, \bibinfo{author}{H.~Garn},
  \bibinfo{author}{T.~Hahn}, \bibinfo{author}{B.~M{\"u}ller-Myhsok}, et~al.,
\newblock \bibinfo{title}{Neurobiology of the major psychoses: a translational
  perspective on brain structure and function—the {FOR}2107 consortium},
\newblock \bibinfo{journal}{European Archives of Psychiatry and Clinical
  Neuroscience} \bibinfo{volume}{269} (\bibinfo{year}{2019})
  \bibinfo{pages}{949--962}.
\bibitem[{Davatzikos et~al.(2009)Davatzikos, Xu, An, Fan, and
  Resnick}]{Davatzikos2009}
\bibinfo{author}{C.~Davatzikos}, \bibinfo{author}{F.~Xu},
  \bibinfo{author}{Y.~An}, \bibinfo{author}{Y.~Fan}, \bibinfo{author}{S.~M.
  Resnick},
\newblock \bibinfo{title}{Longitudinal progression of {A}lzheimer's-like
  patterns of atrophy in normal older adults: the {SPARE}-{AD} index},
\newblock \bibinfo{journal}{Brain} \bibinfo{volume}{132} (\bibinfo{year}{2009})
  \bibinfo{pages}{2026--2035}.
\bibitem[{Greenstein et~al.(2012)Greenstein, Weisinger, Malley, Clasen, and
  Gogtay}]{greenstein2012ROIschizophrenia}
\bibinfo{author}{D.~Greenstein}, \bibinfo{author}{B.~Weisinger},
  \bibinfo{author}{J.~Malley}, \bibinfo{author}{L.~Clasen},
  \bibinfo{author}{N.~Gogtay},
\newblock \bibinfo{title}{Using multivariate machine learning methods and
  structural {MRI} to classify childhood onset schizophrenia and healthy
  controls},
\newblock \bibinfo{journal}{Frontiers in Psychiatry} \bibinfo{volume}{3}
  (\bibinfo{year}{2012}) \bibinfo{pages}{53}.
\bibitem[{Gould et~al.(2014)Gould, Shepherd, Laurens, Cairns, Carr, and
  Green}]{gould2014multivariate}
\bibinfo{author}{I.~C. Gould}, \bibinfo{author}{A.~M. Shepherd},
  \bibinfo{author}{K.~R. Laurens}, \bibinfo{author}{M.~J. Cairns},
  \bibinfo{author}{V.~J. Carr}, \bibinfo{author}{M.~J. Green},
\newblock \bibinfo{title}{Multivariate neuroanatomical classification of
  cognitive subtypes in schizophrenia: a support vector machine learning
  approach},
\newblock \bibinfo{journal}{NeuroImage: Clinical} \bibinfo{volume}{6}
  (\bibinfo{year}{2014}) \bibinfo{pages}{229--236}.
\bibitem[{Du et~al.(2018)Du, Fu, and Calhoun}]{du2018classification}
\bibinfo{author}{Y.~Du}, \bibinfo{author}{Z.~Fu}, \bibinfo{author}{V.~D.
  Calhoun},
\newblock \bibinfo{title}{Classification and prediction of brain disorders
  using functional connectivity: promising but challenging},
\newblock \bibinfo{journal}{Frontiers in Neuroscience} \bibinfo{volume}{12}
  (\bibinfo{year}{2018}) \bibinfo{pages}{525}.
\bibitem[{Vieira et~al.(2020)Vieira, Gong, Pinaya, Scarpazza, Tognin,
  Crespo-Facorro, Tordesillas-Gutierrez, Ortiz-Garc{\'\i}a, Setien-Suero,
  Scheepers et~al.}]{vieira2020using}
\bibinfo{author}{S.~Vieira}, \bibinfo{author}{Q.-y. Gong},
  \bibinfo{author}{W.~H. Pinaya}, \bibinfo{author}{C.~Scarpazza},
  \bibinfo{author}{S.~Tognin}, \bibinfo{author}{B.~Crespo-Facorro},
  \bibinfo{author}{D.~Tordesillas-Gutierrez},
  \bibinfo{author}{V.~Ortiz-Garc{\'\i}a}, \bibinfo{author}{E.~Setien-Suero},
  \bibinfo{author}{F.~E. Scheepers}, et~al.,
\newblock \bibinfo{title}{Using machine learning and structural neuroimaging to
  detect first episode psychosis: reconsidering the evidence},
\newblock \bibinfo{journal}{Schizophrenia Bulletin} \bibinfo{volume}{46}
  (\bibinfo{year}{2020}) \bibinfo{pages}{17--26}.
\bibitem[{de~Filippis et~al.(2019)de~Filippis, Carbone, Gaetano, Bruni,
  Pugliese, Segura-Garcia, and De~Fazio}]{de2019machine}
\bibinfo{author}{R.~de~Filippis}, \bibinfo{author}{E.~A. Carbone},
  \bibinfo{author}{R.~Gaetano}, \bibinfo{author}{A.~Bruni},
  \bibinfo{author}{V.~Pugliese}, \bibinfo{author}{C.~Segura-Garcia},
  \bibinfo{author}{P.~De~Fazio},
\newblock \bibinfo{title}{Machine learning techniques in a structural and
  functional {MRI} diagnostic approach in schizophrenia: a systematic review},
\newblock \bibinfo{journal}{Neuropsychiatric Disease and Treatment}
  \bibinfo{volume}{15} (\bibinfo{year}{2019}) \bibinfo{pages}{1605}.
\bibitem[{Pinaya et~al.(2016)Pinaya, Gadelha, Doyle, Noto, Zugman, Cordeiro,
  Jackowski, Bressan, and Sato}]{pinaya2016using}
\bibinfo{author}{W.~H. Pinaya}, \bibinfo{author}{A.~Gadelha},
  \bibinfo{author}{O.~M. Doyle}, \bibinfo{author}{C.~Noto},
  \bibinfo{author}{A.~Zugman}, \bibinfo{author}{Q.~Cordeiro},
  \bibinfo{author}{A.~P. Jackowski}, \bibinfo{author}{R.~A. Bressan},
  \bibinfo{author}{J.~R. Sato},
\newblock \bibinfo{title}{Using deep belief network modelling to characterize
  differences in brain morphometry in schizophrenia},
\newblock \bibinfo{journal}{Scientific Reports} \bibinfo{volume}{6}
  (\bibinfo{year}{2016}) \bibinfo{pages}{38897}.
\bibitem[{Oh et~al.(2020)Oh, Oh, Lee, Chae, and Yun}]{oh2020identifying}
\bibinfo{author}{J.~Oh}, \bibinfo{author}{B.-L. Oh}, \bibinfo{author}{K.-U.
  Lee}, \bibinfo{author}{J.-H. Chae}, \bibinfo{author}{K.~Yun},
\newblock \bibinfo{title}{Identifying schizophrenia using structural {MRI} with
  a deep learning algorithm},
\newblock \bibinfo{journal}{Frontiers in Psychiatry} \bibinfo{volume}{11}
  (\bibinfo{year}{2020}) \bibinfo{pages}{16}.
\bibitem[{Kim et~al.(2016)Kim, Calhoun, Shim, and Lee}]{kim2016deep}
\bibinfo{author}{J.~Kim}, \bibinfo{author}{V.~D. Calhoun},
  \bibinfo{author}{E.~Shim}, \bibinfo{author}{J.-H. Lee},
\newblock \bibinfo{title}{Deep neural network with weight sparsity control and
  pre-training extracts hierarchical features and enhances classification
  performance: Evidence from whole-brain resting-state functional connectivity
  patterns of schizophrenia},
\newblock \bibinfo{journal}{NeuroImage} \bibinfo{volume}{124}
  (\bibinfo{year}{2016}) \bibinfo{pages}{127--146}.
\bibitem[{Zeng et~al.(2018)Zeng, Wang, Hu, Yang, Pu, Shen, Chen, Liu, Yin, Tan
  et~al.}]{zeng2018multi}
\bibinfo{author}{L.-L. Zeng}, \bibinfo{author}{H.~Wang},
  \bibinfo{author}{P.~Hu}, \bibinfo{author}{B.~Yang}, \bibinfo{author}{W.~Pu},
  \bibinfo{author}{H.~Shen}, \bibinfo{author}{X.~Chen},
  \bibinfo{author}{Z.~Liu}, \bibinfo{author}{H.~Yin}, \bibinfo{author}{Q.~Tan},
  et~al.,
\newblock \bibinfo{title}{Multi-site diagnostic classification of schizophrenia
  using discriminant deep learning with functional connectivity {MRI}},
\newblock \bibinfo{journal}{EBioMedicine} \bibinfo{volume}{30}
  (\bibinfo{year}{2018}) \bibinfo{pages}{74--85}.
\bibitem[{Li et~al.(2020)Li, Han, Wang, Hu, Calhoun, and
  Wang}]{li2020application}
\bibinfo{author}{G.~Li}, \bibinfo{author}{D.~Han}, \bibinfo{author}{C.~Wang},
  \bibinfo{author}{W.~Hu}, \bibinfo{author}{V.~D. Calhoun},
  \bibinfo{author}{Y.-P. Wang},
\newblock \bibinfo{title}{Application of deep canonically correlated sparse
  autoencoder for the classification of schizophrenia},
\newblock \bibinfo{journal}{Computer Methods and Programs in Biomedicine}
  \bibinfo{volume}{183} (\bibinfo{year}{2020}) \bibinfo{pages}{105073}.
\bibitem[{Matsubara et~al.(2019)Matsubara, Tashiro, and
  Uehara}]{matsubara2019deep}
\bibinfo{author}{T.~Matsubara}, \bibinfo{author}{T.~Tashiro},
  \bibinfo{author}{K.~Uehara},
\newblock \bibinfo{title}{Deep neural generative model of functional mri images
  for psychiatric disorder diagnosis},
\newblock \bibinfo{journal}{IEEE Transactions on Biomedical Engineering}
  \bibinfo{volume}{66} (\bibinfo{year}{2019}) \bibinfo{pages}{2768--2779}.
\bibitem[{Qureshi et~al.(2019)Qureshi, Oh, and Lee}]{qureshi20193d}
\bibinfo{author}{M.~N.~I. Qureshi}, \bibinfo{author}{J.~Oh},
  \bibinfo{author}{B.~Lee},
\newblock \bibinfo{title}{3{D}-{CNN} based discrimination of schizophrenia
  using resting-state f{MRI}},
\newblock \bibinfo{journal}{Artificial Intelligence in Medicine}
  \bibinfo{volume}{98} (\bibinfo{year}{2019}) \bibinfo{pages}{10--17}.
\bibitem[{Oh et~al.(2019)Oh, Kim, Shen, Piao, Kang, Oh, and
  Chung}]{oh2019classification}
\bibinfo{author}{K.~Oh}, \bibinfo{author}{W.~Kim}, \bibinfo{author}{G.~Shen},
  \bibinfo{author}{Y.~Piao}, \bibinfo{author}{N.-I. Kang},
  \bibinfo{author}{I.-S. Oh}, \bibinfo{author}{Y.~C. Chung},
\newblock \bibinfo{title}{Classification of schizophrenia and normal controls
  using 3{D} convolutional neural network and outcome visualization},
\newblock \bibinfo{journal}{Schizophrenia Research} \bibinfo{volume}{212}
  (\bibinfo{year}{2019}) \bibinfo{pages}{186--195}.
\bibitem[{Lei et~al.(2020)Lei, Pinaya, van Amelsvoort, Marcelis, Donohoe,
  Mothersill, Corvin, Gill, Vieira, Huang, and et~al.}]{lei2019detecting}
\bibinfo{author}{D.~Lei}, \bibinfo{author}{W.~H.~L. Pinaya},
  \bibinfo{author}{T.~van Amelsvoort}, \bibinfo{author}{M.~Marcelis},
  \bibinfo{author}{G.~Donohoe}, \bibinfo{author}{D.~O. Mothersill},
  \bibinfo{author}{A.~Corvin}, \bibinfo{author}{M.~Gill},
  \bibinfo{author}{S.~Vieira}, \bibinfo{author}{X.~Huang},
  \bibinfo{author}{et~al.},
\newblock \bibinfo{title}{Detecting schizophrenia at the level of the
  individual: relative diagnostic value of whole-brain images, connectome-wide
  functional connectivity and graph-based metrics},
\newblock \bibinfo{journal}{Psychological Medicine} \bibinfo{volume}{50}
  (\bibinfo{year}{2020}) \bibinfo{pages}{1852–1861}.
\bibitem[{Plis et~al.(2018)Plis, Amin, Chekroud, Hjelm, Damaraju, Lee,
  Bustillo, Cho, Pearlson, and Calhoun}]{plis2018reading}
\bibinfo{author}{S.~M. Plis}, \bibinfo{author}{M.~F. Amin},
  \bibinfo{author}{A.~Chekroud}, \bibinfo{author}{D.~Hjelm},
  \bibinfo{author}{E.~Damaraju}, \bibinfo{author}{H.~J. Lee},
  \bibinfo{author}{J.~R. Bustillo}, \bibinfo{author}{K.~Cho},
  \bibinfo{author}{G.~D. Pearlson}, \bibinfo{author}{V.~D. Calhoun},
\newblock \bibinfo{title}{Reading the (functional) writing on the (structural)
  wall: multimodal fusion of brain structure and function via a deep neural
  network based translation approach reveals novel impairments in
  schizophrenia},
\newblock \bibinfo{journal}{NeuroImage} \bibinfo{volume}{181}
  (\bibinfo{year}{2018}) \bibinfo{pages}{734--747}.
\bibitem[{Yan et~al.(2019)Yan, Calhoun, Song, Cui, Yan, Liu, Fan, Zuo, Yang, Xu
  et~al.}]{yan2019discriminating}
\bibinfo{author}{W.~Yan}, \bibinfo{author}{V.~Calhoun},
  \bibinfo{author}{M.~Song}, \bibinfo{author}{Y.~Cui},
  \bibinfo{author}{H.~Yan}, \bibinfo{author}{S.~Liu}, \bibinfo{author}{L.~Fan},
  \bibinfo{author}{N.~Zuo}, \bibinfo{author}{Z.~Yang}, \bibinfo{author}{K.~Xu},
  et~al.,
\newblock \bibinfo{title}{Discriminating schizophrenia using recurrent neural
  network applied on time courses of multi-site f{MRI} data},
\newblock \bibinfo{journal}{EBioMedicine} \bibinfo{volume}{47}
  (\bibinfo{year}{2019}) \bibinfo{pages}{543--552}.
\bibitem[{Ionescu et~al.(2013)Ionescu, Niciu, Mathews, Richards, and
  Zarate~Jr}]{ionescu2013neurobiology}
\bibinfo{author}{D.~F. Ionescu}, \bibinfo{author}{M.~J. Niciu},
  \bibinfo{author}{D.~C. Mathews}, \bibinfo{author}{E.~M. Richards},
  \bibinfo{author}{C.~A. Zarate~Jr},
\newblock \bibinfo{title}{Neurobiology of anxious depression: a review},
\newblock \bibinfo{journal}{Depression and Anxiety} \bibinfo{volume}{30}
  (\bibinfo{year}{2013}) \bibinfo{pages}{374--385}.
\bibitem[{Gao et~al.(2018)Gao, Calhoun, and Sui}]{gao2018machine}
\bibinfo{author}{S.~Gao}, \bibinfo{author}{V.~D. Calhoun},
  \bibinfo{author}{J.~Sui},
\newblock \bibinfo{title}{Machine learning in major depression: From
  classification to treatment outcome prediction},
\newblock \bibinfo{journal}{CNS Neuroscience \& Therapeutics}
  \bibinfo{volume}{24} (\bibinfo{year}{2018}) \bibinfo{pages}{1037--1052}.
\bibitem[{Jiang et~al.(2018)Jiang, Abbott, Jiang, Du, Espinoza, Narr, Wade, Yu,
  Song, Lin et~al.}]{jiang2018smri}
\bibinfo{author}{R.~Jiang}, \bibinfo{author}{C.~C. Abbott},
  \bibinfo{author}{T.~Jiang}, \bibinfo{author}{Y.~Du},
  \bibinfo{author}{R.~Espinoza}, \bibinfo{author}{K.~L. Narr},
  \bibinfo{author}{B.~Wade}, \bibinfo{author}{Q.~Yu},
  \bibinfo{author}{M.~Song}, \bibinfo{author}{D.~Lin}, et~al.,
\newblock \bibinfo{title}{S{MRI} biomarkers predict electroconvulsive treatment
  outcomes: accuracy with independent data sets},
\newblock \bibinfo{journal}{Neuropsychopharmacology} \bibinfo{volume}{43}
  (\bibinfo{year}{2018}) \bibinfo{pages}{1078--1087}.
\bibitem[{Nunes et~al.(2018)Nunes, Schnack, Ching, Agartz, Akudjedu, Alda,
  Aln{\ae}s, Alonso-Lana, Bauer, Baune et~al.}]{nunes2018using}
\bibinfo{author}{A.~Nunes}, \bibinfo{author}{H.~G. Schnack},
  \bibinfo{author}{C.~R. Ching}, \bibinfo{author}{I.~Agartz},
  \bibinfo{author}{T.~N. Akudjedu}, \bibinfo{author}{M.~Alda},
  \bibinfo{author}{D.~Aln{\ae}s}, \bibinfo{author}{S.~Alonso-Lana},
  \bibinfo{author}{J.~Bauer}, \bibinfo{author}{B.~T. Baune}, et~al.,
\newblock \bibinfo{title}{Using structural {MRI} to identify bipolar
  disorders--13 site machine learning study in 3020 individuals from the
  {ENIGMA} {B}ipolar {D}isorders working group},
\newblock \bibinfo{journal}{Molecular Psychiatry}  (\bibinfo{year}{2018})
  \bibinfo{pages}{1--14}.
\bibitem[{Pominova et~al.(2018)Pominova, Artemov, Sharaev, Kondrateva,
  Bernstein, and Burnaev}]{pominova2018voxelwise}
\bibinfo{author}{M.~Pominova}, \bibinfo{author}{A.~Artemov},
  \bibinfo{author}{M.~Sharaev}, \bibinfo{author}{E.~Kondrateva},
  \bibinfo{author}{A.~Bernstein}, \bibinfo{author}{E.~Burnaev},
\newblock \bibinfo{title}{Voxelwise 3{D} convolutional and recurrent neural
  networks for epilepsy and depression diagnostics from structural and
  functional {MRI} data},
\newblock in: \bibinfo{booktitle}{2018 IEEE International Conference on Data
  Mining Workshops (ICDMW)}, \bibinfo{organization}{IEEE},
  \bibinfo{year}{2018}, pp. \bibinfo{pages}{299--307}.
\bibitem[{Bruin et~al.(2019)Bruin, Denys, and van Wingen}]{bruin2019diagnostic}
\bibinfo{author}{W.~Bruin}, \bibinfo{author}{D.~Denys}, \bibinfo{author}{G.~van
  Wingen},
\newblock \bibinfo{title}{Diagnostic neuroimaging markers of
  obsessive-compulsive disorder: initial evidence from structural and
  functional {MRI} studies},
\newblock \bibinfo{journal}{Progress in Neuro-Psychopharmacology and Biological
  Psychiatry} \bibinfo{volume}{91} (\bibinfo{year}{2019})
  \bibinfo{pages}{49--59}.
\bibitem[{Khodatars et~al.(2020)Khodatars, Shoeibi, Ghassemi, Jafari, Khadem,
  Sadeghi, Moridian, Hussain, Alizadehsani, Zare et~al.}]{khodatars2020deep}
\bibinfo{author}{M.~Khodatars}, \bibinfo{author}{A.~Shoeibi},
  \bibinfo{author}{N.~Ghassemi}, \bibinfo{author}{M.~Jafari},
  \bibinfo{author}{A.~Khadem}, \bibinfo{author}{D.~Sadeghi},
  \bibinfo{author}{P.~Moridian}, \bibinfo{author}{S.~Hussain},
  \bibinfo{author}{R.~Alizadehsani}, \bibinfo{author}{A.~Zare}, et~al.,
\newblock \bibinfo{title}{Deep learning for neuroimaging-based diagnosis and
  rehabilitation of autism spectrum disorder: A review},
\newblock \bibinfo{journal}{arXiv preprint arXiv:2007.01285}
  (\bibinfo{year}{2020}).
\bibitem[{Seidman et~al.(2005)Seidman, Valera, and
  Makris}]{seidman2005structural}
\bibinfo{author}{L.~J. Seidman}, \bibinfo{author}{E.~M. Valera},
  \bibinfo{author}{N.~Makris},
\newblock \bibinfo{title}{Structural brain imaging of
  attention-deficit/hyperactivity disorder},
\newblock \bibinfo{journal}{Biological Psychiatry} \bibinfo{volume}{57}
  (\bibinfo{year}{2005}) \bibinfo{pages}{1263--1272}.
\bibitem[{Yu-Feng et~al.(2007)Yu-Feng, Yong, Chao-Zhe, Qing-Jiu, Man-Qiu, Meng,
  Li-Xia, Tian-Zi, and Yu-Feng}]{yu2007altered}
\bibinfo{author}{Z.~Yu-Feng}, \bibinfo{author}{H.~Yong},
  \bibinfo{author}{Z.~Chao-Zhe}, \bibinfo{author}{C.~Qing-Jiu},
  \bibinfo{author}{S.~Man-Qiu}, \bibinfo{author}{L.~Meng},
  \bibinfo{author}{T.~Li-Xia}, \bibinfo{author}{J.~Tian-Zi},
  \bibinfo{author}{W.~Yu-Feng},
\newblock \bibinfo{title}{Altered baseline brain activity in children with
  {ADHD} revealed by resting-state functional {MRI}},
\newblock \bibinfo{journal}{Brain and Development} \bibinfo{volume}{29}
  (\bibinfo{year}{2007}) \bibinfo{pages}{83--91}.
\bibitem[{Brown et~al.(2012)Brown, Sidhu, Greiner, Asgarian, Bastani,
  Silverstone, Greenshaw, and Dursun}]{brown2012adhd}
\bibinfo{author}{M.~R. Brown}, \bibinfo{author}{G.~S. Sidhu},
  \bibinfo{author}{R.~Greiner}, \bibinfo{author}{N.~Asgarian},
  \bibinfo{author}{M.~Bastani}, \bibinfo{author}{P.~H. Silverstone},
  \bibinfo{author}{A.~J. Greenshaw}, \bibinfo{author}{S.~M. Dursun},
\newblock \bibinfo{title}{{ADHD}-200 global competition: diagnosing {ADHD}
  using personal characteristic data can outperform resting state f{MRI}
  measurements},
\newblock \bibinfo{journal}{Frontiers in Systems Neuroscience}
  \bibinfo{volume}{6} (\bibinfo{year}{2012}) \bibinfo{pages}{69}.
\bibitem[{Kuang et~al.(2014)Kuang, Guo, An, Zhao, and
  He}]{kuang2014discrimination}
\bibinfo{author}{D.~Kuang}, \bibinfo{author}{X.~Guo}, \bibinfo{author}{X.~An},
  \bibinfo{author}{Y.~Zhao}, \bibinfo{author}{L.~He},
\newblock \bibinfo{title}{Discrimination of {ADHD} based on f{MRI} data with
  deep belief network},
\newblock in: \bibinfo{booktitle}{International Conference on Intelligent
  Computing}, \bibinfo{organization}{Springer}, \bibinfo{year}{2014}, pp.
  \bibinfo{pages}{225--232}.
\bibitem[{Deshpande et~al.(2015)Deshpande, Wang, Rangaprakash, and
  Wilamowski}]{deshpande2015fully}
\bibinfo{author}{G.~Deshpande}, \bibinfo{author}{P.~Wang},
  \bibinfo{author}{D.~Rangaprakash}, \bibinfo{author}{B.~Wilamowski},
\newblock \bibinfo{title}{Fully connected cascade artificial neural network
  architecture for attention deficit hyperactivity disorder classification from
  functional magnetic resonance imaging data},
\newblock \bibinfo{journal}{IEEE Transactions on Cybernetics}
  \bibinfo{volume}{45} (\bibinfo{year}{2015}) \bibinfo{pages}{2668--2679}.
\bibitem[{Riaz et~al.(2020)Riaz, Asad, Alonso, and Slabaugh}]{riaz2020deepfmri}
\bibinfo{author}{A.~Riaz}, \bibinfo{author}{M.~Asad},
  \bibinfo{author}{E.~Alonso}, \bibinfo{author}{G.~Slabaugh},
\newblock \bibinfo{title}{Deep{FMRI}: End-to-end deep learning for functional
  connectivity and classification of {ADHD} using f{MRI}},
\newblock \bibinfo{journal}{Journal of Neuroscience Methods}
  \bibinfo{volume}{335} (\bibinfo{year}{2020}) \bibinfo{pages}{108506}.
\bibitem[{Mao et~al.(2019)Mao, Su, Xu, Wang, Huang, Yue, Sun, and
  Xiong}]{mao2019spatio}
\bibinfo{author}{Z.~Mao}, \bibinfo{author}{Y.~Su}, \bibinfo{author}{G.~Xu},
  \bibinfo{author}{X.~Wang}, \bibinfo{author}{Y.~Huang},
  \bibinfo{author}{W.~Yue}, \bibinfo{author}{L.~Sun},
  \bibinfo{author}{N.~Xiong},
\newblock \bibinfo{title}{Spatio-temporal deep learning method for {ADHD}
  f{MRI} classification},
\newblock \bibinfo{journal}{Information Sciences} \bibinfo{volume}{499}
  (\bibinfo{year}{2019}) \bibinfo{pages}{1--11}.
\bibitem[{Chen et~al.(2019)Chen, Li, Wang, Dillman, Parikh, and
  He}]{chen2019multichannel}
\bibinfo{author}{M.~Chen}, \bibinfo{author}{H.~Li}, \bibinfo{author}{J.~Wang},
  \bibinfo{author}{J.~R. Dillman}, \bibinfo{author}{N.~A. Parikh},
  \bibinfo{author}{L.~He},
\newblock \bibinfo{title}{A multichannel deep neural network model analyzing
  multiscale functional brain connectome data for attention deficit
  hyperactivity disorder detection},
\newblock \bibinfo{journal}{Radiology: Artificial Intelligence}
  \bibinfo{volume}{2} (\bibinfo{year}{2019}) \bibinfo{pages}{e190012}.
\bibitem[{Van~Rooij et~al.(2018)Van~Rooij, Anagnostou, Arango, Auzias,
  Behrmann, Busatto, Calderoni, Daly, Deruelle, Di~Martino
  et~al.}]{van2018cortical}
\bibinfo{author}{D.~Van~Rooij}, \bibinfo{author}{E.~Anagnostou},
  \bibinfo{author}{C.~Arango}, \bibinfo{author}{G.~Auzias},
  \bibinfo{author}{M.~Behrmann}, \bibinfo{author}{G.~F. Busatto},
  \bibinfo{author}{S.~Calderoni}, \bibinfo{author}{E.~Daly},
  \bibinfo{author}{C.~Deruelle}, \bibinfo{author}{A.~Di~Martino}, et~al.,
\newblock \bibinfo{title}{Cortical and subcortical brain morphometry
  differences between patients with autism spectrum disorder and healthy
  individuals across the lifespan: results from the {ENIGMA} {ASD} working
  group},
\newblock \bibinfo{journal}{American Journal of Psychiatry}
  \bibinfo{volume}{175} (\bibinfo{year}{2018}) \bibinfo{pages}{359--369}.
\bibitem[{Kennedy and Courchesne(2008)}]{kennedy2008intrinsic}
\bibinfo{author}{D.~P. Kennedy}, \bibinfo{author}{E.~Courchesne},
\newblock \bibinfo{title}{The intrinsic functional organization of the brain is
  altered in autism},
\newblock \bibinfo{journal}{NeuroImage} \bibinfo{volume}{39}
  (\bibinfo{year}{2008}) \bibinfo{pages}{1877--1885}.
\bibitem[{Heinsfeld et~al.(2018)Heinsfeld, Franco, Craddock, Buchweitz, and
  Meneguzzi}]{heinsfeld2018identification}
\bibinfo{author}{A.~S. Heinsfeld}, \bibinfo{author}{A.~R. Franco},
  \bibinfo{author}{R.~C. Craddock}, \bibinfo{author}{A.~Buchweitz},
  \bibinfo{author}{F.~Meneguzzi},
\newblock \bibinfo{title}{Identification of autism spectrum disorder using deep
  learning and the {ABIDE} dataset},
\newblock \bibinfo{journal}{NeuroImage: Clinical} \bibinfo{volume}{17}
  (\bibinfo{year}{2018}) \bibinfo{pages}{16--23}.
\bibitem[{B{\"u}hler and Mann(2011)}]{buhler2011alcohol}
\bibinfo{author}{M.~B{\"u}hler}, \bibinfo{author}{K.~Mann},
\newblock \bibinfo{title}{Alcohol and the human brain: a systematic review of
  different neuroimaging methods},
\newblock \bibinfo{journal}{Alcoholism: Clinical and Experimental Research}
  \bibinfo{volume}{35} (\bibinfo{year}{2011}) \bibinfo{pages}{1771--1793}.
\bibitem[{Zahr and Pfefferbaum(2017)}]{zahr2017alcohol}
\bibinfo{author}{N.~M. Zahr}, \bibinfo{author}{A.~Pfefferbaum},
\newblock \bibinfo{title}{Alcohol’s effects on the brain: neuroimaging
  results in humans and animal models.},
\newblock \bibinfo{journal}{Alcohol Research: Current Reviews}
  (\bibinfo{year}{2017}).
\bibitem[{Chanraud et~al.(2007)Chanraud, Martelli, Delain, Kostogianni, Douaud,
  Aubin, Reynaud, and Martinot}]{chanraud2007brain}
\bibinfo{author}{S.~Chanraud}, \bibinfo{author}{C.~Martelli},
  \bibinfo{author}{F.~Delain}, \bibinfo{author}{N.~Kostogianni},
  \bibinfo{author}{G.~Douaud}, \bibinfo{author}{H.-J. Aubin},
  \bibinfo{author}{M.~Reynaud}, \bibinfo{author}{J.-L. Martinot},
\newblock \bibinfo{title}{Brain morphometry and cognitive performance in
  detoxified alcohol-dependents with preserved psychosocial functioning},
\newblock \bibinfo{journal}{Neuropsychopharmacology} \bibinfo{volume}{32}
  (\bibinfo{year}{2007}) \bibinfo{pages}{429--438}.
\bibitem[{Hu et~al.(2018)Hu, Ide, Chao, Zhornitsky, Fischer, Wang, Zhang, and
  Chiang-shan}]{hu2018resting}
\bibinfo{author}{S.~Hu}, \bibinfo{author}{J.~S. Ide}, \bibinfo{author}{H.~H.
  Chao}, \bibinfo{author}{S.~Zhornitsky}, \bibinfo{author}{K.~A. Fischer},
  \bibinfo{author}{W.~Wang}, \bibinfo{author}{S.~Zhang}, \bibinfo{author}{R.~L.
  Chiang-shan},
\newblock \bibinfo{title}{Resting state functional connectivity of the amygdala
  and problem drinking in non-dependent alcohol drinkers},
\newblock \bibinfo{journal}{Drug and Alcohol Dependence} \bibinfo{volume}{185}
  (\bibinfo{year}{2018}) \bibinfo{pages}{173--180}.
\bibitem[{Wang et~al.(2018)Wang, Zhao, Nie, Liu, and Chen}]{wang2018disrupted}
\bibinfo{author}{Y.~Wang}, \bibinfo{author}{Y.~Zhao}, \bibinfo{author}{H.~Nie},
  \bibinfo{author}{C.~Liu}, \bibinfo{author}{J.~Chen},
\newblock \bibinfo{title}{Disrupted brain network efficiency and decreased
  functional connectivity in multi-sensory modality regions in male patients
  with alcohol use disorder},
\newblock \bibinfo{journal}{Frontiers in Human Neuroscience}
  \bibinfo{volume}{12} (\bibinfo{year}{2018}) \bibinfo{pages}{513}.
\bibitem[{Guggenmos et~al.(2018)Guggenmos, Scheel, Sekutowicz, Garbusow,
  Sebold, Sommer, Charlet, Beck, Wittchen, Zimmermann
  et~al.}]{guggenmos2018decoding}
\bibinfo{author}{M.~Guggenmos}, \bibinfo{author}{M.~Scheel},
  \bibinfo{author}{M.~Sekutowicz}, \bibinfo{author}{M.~Garbusow},
  \bibinfo{author}{M.~Sebold}, \bibinfo{author}{C.~Sommer},
  \bibinfo{author}{K.~Charlet}, \bibinfo{author}{A.~Beck},
  \bibinfo{author}{H.-U. Wittchen}, \bibinfo{author}{U.~Zimmermann}, et~al.,
\newblock \bibinfo{title}{Decoding diagnosis and lifetime consumption in
  alcohol dependence from grey-matter pattern information},
\newblock \bibinfo{journal}{Acta Psychiatrica Scandinavica}
  \bibinfo{volume}{137} (\bibinfo{year}{2018}) \bibinfo{pages}{252--262}.
\bibitem[{Guggenmos et~al.(2020)Guggenmos, Schmack, Veer, Lett, Sekutowicz,
  Sebold, Garbusow, Sommer, Wittchen, Zimmermann
  et~al.}]{guggenmos2020multimodal}
\bibinfo{author}{M.~Guggenmos}, \bibinfo{author}{K.~Schmack},
  \bibinfo{author}{I.~M. Veer}, \bibinfo{author}{T.~Lett},
  \bibinfo{author}{M.~Sekutowicz}, \bibinfo{author}{M.~Sebold},
  \bibinfo{author}{M.~Garbusow}, \bibinfo{author}{C.~Sommer},
  \bibinfo{author}{H.-U. Wittchen}, \bibinfo{author}{U.~S. Zimmermann}, et~al.,
\newblock \bibinfo{title}{A multimodal neuroimaging classifier for alcohol
  dependence},
\newblock \bibinfo{journal}{Scientific Reports} \bibinfo{volume}{10}
  (\bibinfo{year}{2020}) \bibinfo{pages}{1--12}.
\bibitem[{Seo et~al.(2015)Seo, Mohr, Beck, W{\"u}stenberg, Heinz, and
  Obermayer}]{seo2015predicting}
\bibinfo{author}{S.~Seo}, \bibinfo{author}{J.~Mohr}, \bibinfo{author}{A.~Beck},
  \bibinfo{author}{T.~W{\"u}stenberg}, \bibinfo{author}{A.~Heinz},
  \bibinfo{author}{K.~Obermayer},
\newblock \bibinfo{title}{Predicting the future relapse of alcohol-dependent
  patients from structural and functional brain images},
\newblock \bibinfo{journal}{Addiction Biology} \bibinfo{volume}{20}
  (\bibinfo{year}{2015}) \bibinfo{pages}{1042--1055}.
\bibitem[{Zhu et~al.(2018)Zhu, Du, Kerich, Lohoff, and Momenan}]{zhu2018random}
\bibinfo{author}{X.~Zhu}, \bibinfo{author}{X.~Du}, \bibinfo{author}{M.~Kerich},
  \bibinfo{author}{F.~W. Lohoff}, \bibinfo{author}{R.~Momenan},
\newblock \bibinfo{title}{Random forest based classification of alcohol
  dependence patients and healthy controls using resting state {MRI}},
\newblock \bibinfo{journal}{Neuroscience Letters} \bibinfo{volume}{676}
  (\bibinfo{year}{2018}) \bibinfo{pages}{27--33}.
\bibitem[{Fede et~al.(2019)Fede, Grodin, Dean, Diazgranados, and
  Momenan}]{fede2019resting}
\bibinfo{author}{S.~J. Fede}, \bibinfo{author}{E.~N. Grodin},
  \bibinfo{author}{S.~F. Dean}, \bibinfo{author}{N.~Diazgranados},
  \bibinfo{author}{R.~Momenan},
\newblock \bibinfo{title}{Resting state connectivity best predicts alcohol use
  severity in moderate to heavy alcohol users},
\newblock \bibinfo{journal}{NeuroImage: Clinical} \bibinfo{volume}{22}
  (\bibinfo{year}{2019}) \bibinfo{pages}{101782}.
\bibitem[{Squeglia et~al.(2017)Squeglia, Ball, Jacobus, Brumback, McKenna,
  Nguyen-Louie, Sorg, Paulus, and Tapert}]{squeglia2017neural}
\bibinfo{author}{L.~M. Squeglia}, \bibinfo{author}{T.~M. Ball},
  \bibinfo{author}{J.~Jacobus}, \bibinfo{author}{T.~Brumback},
  \bibinfo{author}{B.~S. McKenna}, \bibinfo{author}{T.~T. Nguyen-Louie},
  \bibinfo{author}{S.~F. Sorg}, \bibinfo{author}{M.~P. Paulus},
  \bibinfo{author}{S.~F. Tapert},
\newblock \bibinfo{title}{Neural predictors of initiating alcohol use during
  adolescence},
\newblock \bibinfo{journal}{American Journal of Psychiatry}
  \bibinfo{volume}{174} (\bibinfo{year}{2017}) \bibinfo{pages}{172--185}.
\bibitem[{Spechler et~al.(2019)Spechler, Allgaier, Chaarani, Whelan, Watts,
  Orr, Albaugh, D'Alberto, Higgins, Hudson et~al.}]{spechler2019initiation}
\bibinfo{author}{P.~A. Spechler}, \bibinfo{author}{N.~Allgaier},
  \bibinfo{author}{B.~Chaarani}, \bibinfo{author}{R.~Whelan},
  \bibinfo{author}{R.~Watts}, \bibinfo{author}{C.~Orr}, \bibinfo{author}{M.~D.
  Albaugh}, \bibinfo{author}{N.~D'Alberto}, \bibinfo{author}{S.~T. Higgins},
  \bibinfo{author}{K.~E. Hudson}, et~al.,
\newblock \bibinfo{title}{The initiation of cannabis use in adolescence is
  predicted by sex-specific psychosocial and neurobiological features},
\newblock \bibinfo{journal}{European Journal of Neuroscience}
  \bibinfo{volume}{50} (\bibinfo{year}{2019}) \bibinfo{pages}{2346--2356}.
\bibitem[{Wang et~al.(2018)Wang, Lv, Sui, Liu, Wang, and
  Zhang}]{wang2018alcoholism}
\bibinfo{author}{S.-H. Wang}, \bibinfo{author}{Y.-D. Lv},
  \bibinfo{author}{Y.~Sui}, \bibinfo{author}{S.~Liu}, \bibinfo{author}{S.-J.
  Wang}, \bibinfo{author}{Y.-D. Zhang},
\newblock \bibinfo{title}{Alcoholism detection by data augmentation and
  convolutional neural network with stochastic pooling},
\newblock \bibinfo{journal}{Journal of Medical Systems} \bibinfo{volume}{42}
  (\bibinfo{year}{2018}) \bibinfo{pages}{2}.
\bibitem[{Wang et~al.(2019)Wang, Xie, Chen, Guttery, Tang, Sun, and
  Zhang}]{wang2019alcoholism}
\bibinfo{author}{S.-H. Wang}, \bibinfo{author}{S.~Xie},
  \bibinfo{author}{X.~Chen}, \bibinfo{author}{D.~S. Guttery},
  \bibinfo{author}{C.~Tang}, \bibinfo{author}{J.~Sun}, \bibinfo{author}{Y.-D.
  Zhang},
\newblock \bibinfo{title}{Alcoholism identification based on an alexnet
  transfer learning model},
\newblock \bibinfo{journal}{Frontiers in Psychiatry} \bibinfo{volume}{10}
  (\bibinfo{year}{2019}) \bibinfo{pages}{205}.
\bibitem[{Wang et~al.(2020)Wang, Muhammad, Hong, Sangaiah, and
  Zhang}]{wang2020alcoholism}
\bibinfo{author}{S.-H. Wang}, \bibinfo{author}{K.~Muhammad},
  \bibinfo{author}{J.~Hong}, \bibinfo{author}{A.~K. Sangaiah},
  \bibinfo{author}{Y.-D. Zhang},
\newblock \bibinfo{title}{Alcoholism identification via convolutional neural
  network based on parametric {R}e{LU}, dropout, and batch normalization},
\newblock \bibinfo{journal}{Neural Computing and Applications}
  \bibinfo{volume}{32} (\bibinfo{year}{2020}) \bibinfo{pages}{665--680}.
\bibitem[{Elliott et~al.(2019)Elliott, Belsky, Knodt, Ireland, Melzer, Poulton,
  Ramrakha, Caspi, Moffitt, and Hariri}]{elliott2019brain}
\bibinfo{author}{M.~L. Elliott}, \bibinfo{author}{D.~W. Belsky},
  \bibinfo{author}{A.~R. Knodt}, \bibinfo{author}{D.~Ireland},
  \bibinfo{author}{T.~R. Melzer}, \bibinfo{author}{R.~Poulton},
  \bibinfo{author}{S.~Ramrakha}, \bibinfo{author}{A.~Caspi},
  \bibinfo{author}{T.~E. Moffitt}, \bibinfo{author}{A.~R. Hariri},
\newblock \bibinfo{title}{Brain-age in midlife is associated with accelerated
  biological aging and cognitive decline in a longitudinal birth cohort},
\newblock \bibinfo{journal}{Molecular Psychiatry}  (\bibinfo{year}{2019})
  \bibinfo{pages}{1--10}.
\bibitem[{Bashyam et~al.(2020)Bashyam, Erus, Doshi, Habes, Nasralah,
  Truelove-Hill, Srinivasan, Mamourian, Pomponio, Fan et~al.}]{bashyam2020mri}
\bibinfo{author}{V.~M. Bashyam}, \bibinfo{author}{G.~Erus},
  \bibinfo{author}{J.~Doshi}, \bibinfo{author}{M.~Habes},
  \bibinfo{author}{I.~Nasralah}, \bibinfo{author}{M.~Truelove-Hill},
  \bibinfo{author}{D.~Srinivasan}, \bibinfo{author}{L.~Mamourian},
  \bibinfo{author}{R.~Pomponio}, \bibinfo{author}{Y.~Fan}, et~al.,
\newblock \bibinfo{title}{{MRI} signatures of brain age and disease over the
  lifespan based on a deep brain network and 14468 individuals worldwide},
\newblock \bibinfo{journal}{Brain} \bibinfo{volume}{143} (\bibinfo{year}{2020})
  \bibinfo{pages}{2312--2324}.
\bibitem[{Cole et~al.(2018)Cole, Ritchie, Bastin, Hern{\'a}ndez, Maniega,
  Royle, Corley, Pattie, Harris, Zhang et~al.}]{cole2018brain}
\bibinfo{author}{J.~H. Cole}, \bibinfo{author}{S.~J. Ritchie},
  \bibinfo{author}{M.~E. Bastin}, \bibinfo{author}{M.~V. Hern{\'a}ndez},
  \bibinfo{author}{S.~M. Maniega}, \bibinfo{author}{N.~Royle},
  \bibinfo{author}{J.~Corley}, \bibinfo{author}{A.~Pattie},
  \bibinfo{author}{S.~E. Harris}, \bibinfo{author}{Q.~Zhang}, et~al.,
\newblock \bibinfo{title}{Brain age predicts mortality},
\newblock \bibinfo{journal}{Molecular Psychiatry} \bibinfo{volume}{23}
  (\bibinfo{year}{2018}) \bibinfo{pages}{1385--1392}.
\bibitem[{Jiang et~al.(2019)Jiang, Lu, Chen, Yao, Li, Zhang, and
  Guo}]{jiang2019predicting}
\bibinfo{author}{H.~Jiang}, \bibinfo{author}{N.~Lu}, \bibinfo{author}{K.~Chen},
  \bibinfo{author}{L.~Yao}, \bibinfo{author}{K.~Li},
  \bibinfo{author}{J.~Zhang}, \bibinfo{author}{X.~Guo},
\newblock \bibinfo{title}{Predicting brain age of healthy adults based on
  structural {MRI} parcellation using convolutional neural networks},
\newblock \bibinfo{journal}{Frontiers in Neurology} \bibinfo{volume}{10}
  (\bibinfo{year}{2019}).
\bibitem[{Uttal(2011)}]{uttal2011mind}
\bibinfo{author}{W.~R. Uttal}, \bibinfo{title}{Mind and Brain: A Critical
  Appraisal of Cognitive Neuroscience}, \bibinfo{publisher}{MIT Press},
  \bibinfo{year}{2011}.
\bibitem[{Mihalik et~al.(2019)Mihalik, Brudfors, Robu, Ferreira, Lin, Rau, Wu,
  Blumberg, Kanber, Tariq et~al.}]{mihalik2019abcd}
\bibinfo{author}{A.~Mihalik}, \bibinfo{author}{M.~Brudfors},
  \bibinfo{author}{M.~Robu}, \bibinfo{author}{F.~S. Ferreira},
  \bibinfo{author}{H.~Lin}, \bibinfo{author}{A.~Rau}, \bibinfo{author}{T.~Wu},
  \bibinfo{author}{S.~B. Blumberg}, \bibinfo{author}{B.~Kanber},
  \bibinfo{author}{M.~Tariq}, et~al.,
\newblock \bibinfo{title}{{ABCD} neurocognitive prediction challenge 2019:
  Predicting individual fluid intelligence scores from structural {MRI} using
  probabilistic segmentation and kernel ridge regression},
\newblock in: \bibinfo{booktitle}{Challenge in Adolescent Brain Cognitive
  Development Neurocognitive Prediction}, \bibinfo{organization}{Springer},
  \bibinfo{year}{2019}, pp. \bibinfo{pages}{133--142}.
\bibitem[{Hilger et~al.(2020)Hilger, Winter, Leenings, Sassenhagen, Hahn,
  Basten, and Fiebach}]{hilger2020predicting}
\bibinfo{author}{K.~Hilger}, \bibinfo{author}{N.~R. Winter},
  \bibinfo{author}{R.~Leenings}, \bibinfo{author}{J.~Sassenhagen},
  \bibinfo{author}{T.~Hahn}, \bibinfo{author}{U.~Basten},
  \bibinfo{author}{C.~J. Fiebach},
\newblock \bibinfo{title}{Predicting intelligence from brain gray matter
  volume},
\newblock \bibinfo{journal}{Brain Structure and Function}
  (\bibinfo{year}{2020}) \bibinfo{pages}{1--19}.
\bibitem[{Aine(1995)}]{Aine1995}
\bibinfo{author}{C.~Aine},
\newblock \bibinfo{title}{A conceptual overview and critique of functional
  neuroimaging techniques in humans: {I}. {MRI}/f{MRI} and {PET}},
\newblock \bibinfo{journal}{Critical Reviews in Neurobiology}
  \bibinfo{volume}{9} (\bibinfo{year}{1995}) \bibinfo{pages}{229--309}.
\bibitem[{Bellon et~al.(1986)Bellon, Haacke, Coleman, Sacco, Steiger, and
  Gangarosa}]{bellon1986mr}
\bibinfo{author}{E.~M. Bellon}, \bibinfo{author}{E.~M. Haacke},
  \bibinfo{author}{P.~E. Coleman}, \bibinfo{author}{D.~C. Sacco},
  \bibinfo{author}{D.~A. Steiger}, \bibinfo{author}{R.~E. Gangarosa},
\newblock \bibinfo{title}{{MR} artifacts: a review},
\newblock \bibinfo{journal}{American Journal of Roentgenology}
  \bibinfo{volume}{147} (\bibinfo{year}{1986}) \bibinfo{pages}{1271--1281}.
\bibitem[{Liu(2016)}]{liu2016noise}
\bibinfo{author}{T.~T. Liu},
\newblock \bibinfo{title}{Noise contributions to the f{MRI} signal: An
  overview},
\newblock \bibinfo{journal}{NeuroImage} \bibinfo{volume}{143}
  (\bibinfo{year}{2016}) \bibinfo{pages}{141--151}.
\bibitem[{Kather et~al.(2019)Kather, Pearson, Halama, J{\"a}ger, Krause,
  Loosen, Marx, Boor, Tacke, Neumann et~al.}]{kather2019deep}
\bibinfo{author}{J.~N. Kather}, \bibinfo{author}{A.~T. Pearson},
  \bibinfo{author}{N.~Halama}, \bibinfo{author}{D.~J{\"a}ger},
  \bibinfo{author}{J.~Krause}, \bibinfo{author}{S.~H. Loosen},
  \bibinfo{author}{A.~Marx}, \bibinfo{author}{P.~Boor},
  \bibinfo{author}{F.~Tacke}, \bibinfo{author}{U.~P. Neumann}, et~al.,
\newblock \bibinfo{title}{Deep learning can predict microsatellite instability
  directly from histology in gastrointestinal cancer},
\newblock \bibinfo{journal}{Nature Medicine} \bibinfo{volume}{25}
  (\bibinfo{year}{2019}) \bibinfo{pages}{1054--1056}.
\bibitem[{Marinescu et~al.(2018)Marinescu, Oxtoby, Young, Bron, Toga, Weiner,
  Barkhof, Fox, Klein, Alexander et~al.}]{marinescu2018tadpole}
\bibinfo{author}{R.~V. Marinescu}, \bibinfo{author}{N.~P. Oxtoby},
  \bibinfo{author}{A.~L. Young}, \bibinfo{author}{E.~E. Bron},
  \bibinfo{author}{A.~W. Toga}, \bibinfo{author}{M.~W. Weiner},
  \bibinfo{author}{F.~Barkhof}, \bibinfo{author}{N.~C. Fox},
  \bibinfo{author}{S.~Klein}, \bibinfo{author}{D.~C. Alexander}, et~al.,
\newblock \bibinfo{title}{{TADPOLE} challenge: Prediction of longitudinal
  evolution in {A}lzheimer's disease},
\newblock \bibinfo{journal}{arXiv preprint arXiv:1805.03909}
  (\bibinfo{year}{2018}).
\bibitem[{Russakovsky et~al.(2015)Russakovsky, Deng, Su, Krause, Satheesh, Ma,
  Huang, Karpathy, Khosla, Bernstein et~al.}]{russakovsky2015imagenet}
\bibinfo{author}{O.~Russakovsky}, \bibinfo{author}{J.~Deng},
  \bibinfo{author}{H.~Su}, \bibinfo{author}{J.~Krause},
  \bibinfo{author}{S.~Satheesh}, \bibinfo{author}{S.~Ma},
  \bibinfo{author}{Z.~Huang}, \bibinfo{author}{A.~Karpathy},
  \bibinfo{author}{A.~Khosla}, \bibinfo{author}{M.~Bernstein}, et~al.,
\newblock \bibinfo{title}{Imagenet large scale visual recognition challenge},
\newblock \bibinfo{journal}{International Journal of Computer Vision}
  \bibinfo{volume}{115} (\bibinfo{year}{2015}) \bibinfo{pages}{211--252}.
\bibitem[{Cordts et~al.(2016)Cordts, Omran, Ramos, Rehfeld, Enzweiler,
  Benenson, Franke, Roth, and Schiele}]{cordts2016cityscapes}
\bibinfo{author}{M.~Cordts}, \bibinfo{author}{M.~Omran},
  \bibinfo{author}{S.~Ramos}, \bibinfo{author}{T.~Rehfeld},
  \bibinfo{author}{M.~Enzweiler}, \bibinfo{author}{R.~Benenson},
  \bibinfo{author}{U.~Franke}, \bibinfo{author}{S.~Roth},
  \bibinfo{author}{B.~Schiele},
\newblock \bibinfo{title}{The cityscapes dataset for semantic urban scene
  understanding},
\newblock in: \bibinfo{booktitle}{Proceedings of the IEEE conference on
  Computer Vision and Pattern Recognition}, \bibinfo{year}{2016}, pp.
  \bibinfo{pages}{3213--3223}.
\bibitem[{Glocker et~al.(2019)Glocker, Robinson, de~Castro, Dou, and
  Konukoglu}]{Glocker2019MachineLW}
\bibinfo{author}{B.~Glocker}, \bibinfo{author}{R.~Robinson},
  \bibinfo{author}{D.~C. de~Castro}, \bibinfo{author}{Q.~Dou},
  \bibinfo{author}{E.~Konukoglu},
\newblock \bibinfo{title}{Machine learning with multi-site imaging data: An
  empirical study on the impact of scanner effects},
\newblock \bibinfo{journal}{arXiv} \bibinfo{volume}{abs/1910.04597}
  (\bibinfo{year}{2019}).
\bibitem[{Bhagwat et~al.(2020)Bhagwat, Barry, Dickie, Brown, Devenyi, Hatano,
  DuPre, Dagher, Chakravarty, Greenwood, Misic, Kennedy, and
  Poline}]{Bhagwat2020preprocessing}
\bibinfo{author}{N.~Bhagwat}, \bibinfo{author}{A.~Barry},
  \bibinfo{author}{E.~W. Dickie}, \bibinfo{author}{S.~T. Brown},
  \bibinfo{author}{G.~A. Devenyi}, \bibinfo{author}{K.~Hatano},
  \bibinfo{author}{E.~DuPre}, \bibinfo{author}{A.~Dagher},
  \bibinfo{author}{M.~M. Chakravarty}, \bibinfo{author}{C.~M.~T. Greenwood},
  \bibinfo{author}{B.~Misic}, \bibinfo{author}{D.~N. Kennedy},
  \bibinfo{author}{J.-B. Poline},
\newblock \bibinfo{title}{Understanding the impact of preprocessing pipelines
  on neuroimaging cortical surface analyses},
\newblock \bibinfo{journal}{bioRxiv}  (\bibinfo{year}{2020}).
\bibitem[{Schnack and Kahn(2016)}]{Schnack2016samplesize}
\bibinfo{author}{H.~G. Schnack}, \bibinfo{author}{R.~S. Kahn},
\newblock \bibinfo{title}{Detecting neuroimaging biomarkers for psychiatric
  disorders: Sample size matters},
\newblock \bibinfo{journal}{Frontiers in Psychiatry} \bibinfo{volume}{7}
  (\bibinfo{year}{2016}) \bibinfo{pages}{50}.
\bibitem[{Szegedy et~al.(2013)Szegedy, Zaremba, Sutskever, Bruna, Erhan,
  Goodfellow, and Fergus}]{szegedy2013intriguing}
\bibinfo{author}{C.~Szegedy}, \bibinfo{author}{W.~Zaremba},
  \bibinfo{author}{I.~Sutskever}, \bibinfo{author}{J.~Bruna},
  \bibinfo{author}{D.~Erhan}, \bibinfo{author}{I.~Goodfellow},
  \bibinfo{author}{R.~Fergus},
\newblock \bibinfo{title}{Intriguing properties of neural networks},
\newblock \bibinfo{journal}{arXiv preprint arXiv:1312.6199}
  (\bibinfo{year}{2013}).
\bibitem[{Goodfellow et~al.(2014)Goodfellow, Shlens, and
  Szegedy}]{goodfellow2014explaining}
\bibinfo{author}{I.~J. Goodfellow}, \bibinfo{author}{J.~Shlens},
  \bibinfo{author}{C.~Szegedy},
\newblock \bibinfo{title}{Explaining and harnessing adversarial examples},
\newblock \bibinfo{journal}{arXiv preprint arXiv:1412.6572}
  (\bibinfo{year}{2014}).
\bibitem[{Wachinger et~al.(2019)Wachinger, Becker, Rieckmann, and
  P{\"o}lsterl}]{wachinger2019quantifying}
\bibinfo{author}{C.~Wachinger}, \bibinfo{author}{B.~G. Becker},
  \bibinfo{author}{A.~Rieckmann}, \bibinfo{author}{S.~P{\"o}lsterl},
\newblock \bibinfo{title}{Quantifying confounding bias in neuroimaging datasets
  with causal inference},
\newblock in: \bibinfo{booktitle}{International Conference on Medical Image
  Computing and Computer Assisted Intervention},
  \bibinfo{organization}{Springer}, \bibinfo{year}{2019}, pp.
  \bibinfo{pages}{484--492}.
\bibitem[{Tommasi et~al.(2017)Tommasi, Patricia, Caputo, and
  Tuytelaars}]{tommasi2017deeper}
\bibinfo{author}{T.~Tommasi}, \bibinfo{author}{N.~Patricia},
  \bibinfo{author}{B.~Caputo}, \bibinfo{author}{T.~Tuytelaars},
\newblock \bibinfo{title}{A deeper look at dataset bias},
\newblock in: \bibinfo{booktitle}{Domain adaptation in computer vision
  applications}, \bibinfo{publisher}{Springer}, \bibinfo{year}{2017}, pp.
  \bibinfo{pages}{37--55}.
\bibitem[{LeWinn et~al.(2017)LeWinn, Sheridan, Keyes, Hamilton, and
  McLaughlin}]{lewinn2017sample}
\bibinfo{author}{K.~Z. LeWinn}, \bibinfo{author}{M.~A. Sheridan},
  \bibinfo{author}{K.~M. Keyes}, \bibinfo{author}{A.~Hamilton},
  \bibinfo{author}{K.~A. McLaughlin},
\newblock \bibinfo{title}{Sample composition alters associations between age
  and brain structure},
\newblock \bibinfo{journal}{Nature Communications} \bibinfo{volume}{8}
  (\bibinfo{year}{2017}) \bibinfo{pages}{1--14}.
\bibitem[{Snoek et~al.(2019)Snoek, Mileti{\'c}, and Scholte}]{snoek2019control}
\bibinfo{author}{L.~Snoek}, \bibinfo{author}{S.~Mileti{\'c}},
  \bibinfo{author}{H.~S. Scholte},
\newblock \bibinfo{title}{How to control for confounds in decoding analyses of
  neuroimaging data},
\newblock \bibinfo{journal}{NeuroImage} \bibinfo{volume}{184}
  (\bibinfo{year}{2019}) \bibinfo{pages}{741--760}.
\bibitem[{Narla et~al.(2018)Narla, Kuprel, Sarin, Novoa, and
  Ko}]{narla2018automated}
\bibinfo{author}{A.~Narla}, \bibinfo{author}{B.~Kuprel},
  \bibinfo{author}{K.~Sarin}, \bibinfo{author}{R.~Novoa},
  \bibinfo{author}{J.~Ko},
\newblock \bibinfo{title}{Automated classification of skin lesions: from pixels
  to practice},
\newblock \bibinfo{journal}{Journal of Investigative Dermatology}
  \bibinfo{volume}{138} (\bibinfo{year}{2018}) \bibinfo{pages}{2108--2110}.
\bibitem[{Editorial(2019)}]{genomics2019}
\bibinfo{author}{Editorial},
\newblock \bibinfo{title}{Whose genomics?},
\newblock \bibinfo{journal}{Nature Human Behaviour}  (\bibinfo{year}{2019}).
\bibitem[{Sirugo et~al.(2019)Sirugo, Williams, and
  Tishkoff}]{sirugo2019missing}
\bibinfo{author}{G.~Sirugo}, \bibinfo{author}{S.~M. Williams},
  \bibinfo{author}{S.~A. Tishkoff},
\newblock \bibinfo{title}{The missing diversity in human genetic studies},
\newblock \bibinfo{journal}{Cell} \bibinfo{volume}{177} (\bibinfo{year}{2019})
  \bibinfo{pages}{26--31}.
\bibitem[{Rao et~al.(2017)Rao, Monteiro, Mourao-Miranda, and {Alzheimer's
  Disease Initiative}}]{Rao2017-jj}
\bibinfo{author}{A.~Rao}, \bibinfo{author}{J.~M. Monteiro},
  \bibinfo{author}{J.~Mourao-Miranda}, \bibinfo{author}{{Alzheimer's Disease
  Initiative}},
\newblock \bibinfo{title}{Predictive modelling using neuroimaging data in the
  presence of confounds},
\newblock \bibinfo{journal}{Neuroimage} \bibinfo{volume}{150}
  (\bibinfo{year}{2017}) \bibinfo{pages}{23--49}.
\bibitem[{Chyzhyk et~al.(2018)Chyzhyk, Varoquaux, Thirion, and
  Milham}]{Chyzhyk2018-fi}
\bibinfo{author}{D.~Chyzhyk}, \bibinfo{author}{G.~Varoquaux},
  \bibinfo{author}{B.~Thirion}, \bibinfo{author}{M.~Milham},
\newblock \bibinfo{title}{Controlling a confound in predictive models with a
  test set minimizing its effect},
\newblock in: \bibinfo{booktitle}{2018 International Workshop on Pattern
  Recognition in Neuroimaging ({PRNI})},
  \bibinfo{publisher}{ieeexplore.ieee.org}, \bibinfo{year}{2018}, pp.
  \bibinfo{pages}{1--4}.
\bibitem[{Dukart et~al.(2011)Dukart, Schroeter, Mueller, and {Alzheimer's
  Disease Neuroimaging Initiative}}]{Dukart2011-ra}
\bibinfo{author}{J.~Dukart}, \bibinfo{author}{M.~L. Schroeter},
  \bibinfo{author}{K.~Mueller}, \bibinfo{author}{{Alzheimer's Disease
  Neuroimaging Initiative}},
\newblock \bibinfo{title}{Age correction in dementia--matching to a healthy
  brain},
\newblock \bibinfo{journal}{PLoS One} \bibinfo{volume}{6}
  (\bibinfo{year}{2011}) \bibinfo{pages}{e22193}.
\bibitem[{Todd et~al.(2013)Todd, Nystrom, and Cohen}]{Todd2013-yh}
\bibinfo{author}{M.~T. Todd}, \bibinfo{author}{L.~E. Nystrom},
  \bibinfo{author}{J.~D. Cohen},
\newblock \bibinfo{title}{Confounds in multivariate pattern analysis: Theory
  and rule representation case study},
\newblock \bibinfo{journal}{Neuroimage} \bibinfo{volume}{77}
  (\bibinfo{year}{2013}) \bibinfo{pages}{157--165}.
\bibitem[{Kostro et~al.(2014)Kostro, Abdulkadir, Durr, Roos, Leavitt, Johnson,
  Cash, Tabrizi, Scahill, Ronneberger, Kl{\"o}ppel, and {Track-HD
  Investigators}}]{Kostro2014-gs}
\bibinfo{author}{D.~Kostro}, \bibinfo{author}{A.~Abdulkadir},
  \bibinfo{author}{A.~Durr}, \bibinfo{author}{R.~Roos}, \bibinfo{author}{B.~R.
  Leavitt}, \bibinfo{author}{H.~Johnson}, \bibinfo{author}{D.~Cash},
  \bibinfo{author}{S.~J. Tabrizi}, \bibinfo{author}{R.~I. Scahill},
  \bibinfo{author}{O.~Ronneberger}, \bibinfo{author}{S.~Kl{\"o}ppel},
  \bibinfo{author}{{Track-HD Investigators}},
\newblock \bibinfo{title}{Correction of inter-scanner and within-subject
  variance in structural {MRI} based automated diagnosing},
\newblock \bibinfo{journal}{Neuroimage} \bibinfo{volume}{98}
  (\bibinfo{year}{2014}) \bibinfo{pages}{405--415}.
\bibitem[{Dinga et~al.(2020)Dinga, Schmaal, {Brenda W J}, Veltman, and
  Marquand}]{Dinga2020-pa}
\bibinfo{author}{R.~Dinga}, \bibinfo{author}{L.~Schmaal},
  \bibinfo{author}{{Brenda W J}}, \bibinfo{author}{D.~J. Veltman},
  \bibinfo{author}{A.~F. Marquand}, \bibinfo{title}{Controlling for effects of
  confounding variables on machine learning predictions}, \bibinfo{year}{2020}.
\bibitem[{Ma et~al.(2018)Ma, Zhang, Zanetti, Shen, Satterthwaite, Wolf, Gur,
  Fan, Hu, Busatto, and Davatzikos}]{Ma2018-lt}
\bibinfo{author}{Q.~Ma}, \bibinfo{author}{T.~Zhang}, \bibinfo{author}{M.~V.
  Zanetti}, \bibinfo{author}{H.~Shen}, \bibinfo{author}{T.~D. Satterthwaite},
  \bibinfo{author}{D.~H. Wolf}, \bibinfo{author}{R.~E. Gur},
  \bibinfo{author}{Y.~Fan}, \bibinfo{author}{D.~Hu}, \bibinfo{author}{G.~F.
  Busatto}, \bibinfo{author}{C.~Davatzikos},
\newblock \bibinfo{title}{Classification of multi-site {MR} images in the
  presence of heterogeneity using multi-task learning},
\newblock \bibinfo{journal}{Neuroimage Clin} \bibinfo{volume}{19}
  (\bibinfo{year}{2018}) \bibinfo{pages}{476--486}.
\bibitem[{Adeli et~al.(2019)Adeli, Zhao, Pfefferbaum, Sullivan, Fei-Fei,
  Niebles, and Pohl}]{Adeli2019-qp}
\bibinfo{author}{E.~Adeli}, \bibinfo{author}{Q.~Zhao},
  \bibinfo{author}{A.~Pfefferbaum}, \bibinfo{author}{E.~V. Sullivan},
  \bibinfo{author}{L.~Fei-Fei}, \bibinfo{author}{J.~C. Niebles},
  \bibinfo{author}{K.~M. Pohl},
\newblock \bibinfo{title}{Representation learning with statistical independence
  to mitigate bias}  (\bibinfo{year}{2019}).
\bibitem[{Haufe et~al.(2014)Haufe, Meinecke, Görgen, Dähne, Haynes,
  Blankertz, and Bießmann}]{HAUFE201496}
\bibinfo{author}{S.~Haufe}, \bibinfo{author}{F.~Meinecke},
  \bibinfo{author}{K.~Görgen}, \bibinfo{author}{S.~Dähne},
  \bibinfo{author}{J.-D. Haynes}, \bibinfo{author}{B.~Blankertz},
  \bibinfo{author}{F.~Bießmann},
\newblock \bibinfo{title}{On the interpretation of weight vectors of linear
  models in multivariate neuroimaging},
\newblock \bibinfo{journal}{NeuroImage} \bibinfo{volume}{87}
  (\bibinfo{year}{2014}) \bibinfo{pages}{96 -- 110}.
\bibitem[{Kindermans et~al.(2017)Kindermans, Sch{\"u}tt, Alber, M{\"u}ller,
  Erhan, Kim, and D{\"a}hne}]{kindermans2017learning}
\bibinfo{author}{P.-J. Kindermans}, \bibinfo{author}{K.~T. Sch{\"u}tt},
  \bibinfo{author}{M.~Alber}, \bibinfo{author}{K.-R. M{\"u}ller},
  \bibinfo{author}{D.~Erhan}, \bibinfo{author}{B.~Kim},
  \bibinfo{author}{S.~D{\"a}hne},
\newblock \bibinfo{title}{Learning how to explain neural networks:
  {P}attern{N}et and {P}attern{A}ttribution},
\newblock \bibinfo{journal}{arXiv preprint arXiv:1705.05598}
  (\bibinfo{year}{2017}).
\bibitem[{London(2019)}]{london2019accvsexpl}
\bibinfo{author}{A.~J. London},
\newblock \bibinfo{title}{Artificial intelligence and black-box medical
  decisions: Accuracy versus explainability},
\newblock \bibinfo{journal}{Hastings Center Report} \bibinfo{volume}{49}
  (\bibinfo{year}{2019}) \bibinfo{pages}{15--21}.
\bibitem[{Adebayo et~al.(2018)Adebayo, Gilmer, Muelly, Goodfellow, Hardt, and
  Kim}]{adebayo2018sanity}
\bibinfo{author}{J.~Adebayo}, \bibinfo{author}{J.~Gilmer},
  \bibinfo{author}{M.~Muelly}, \bibinfo{author}{I.~Goodfellow},
  \bibinfo{author}{M.~Hardt}, \bibinfo{author}{B.~Kim},
\newblock \bibinfo{title}{Sanity checks for saliency maps},
\newblock in: \bibinfo{booktitle}{Advances in Neural Information Processing
  Systems}, \bibinfo{year}{2018}, pp. \bibinfo{pages}{9505--9515}.
\bibitem[{Sixt et~al.(2019)Sixt, Granz, and Landgraf}]{sixt2019explanations}
\bibinfo{author}{L.~Sixt}, \bibinfo{author}{M.~Granz},
  \bibinfo{author}{T.~Landgraf},
\newblock \bibinfo{title}{When explanations lie: Why modified {BP} attribution
  fails},
\newblock \bibinfo{journal}{arXiv preprint arXiv:1912.09818}
  (\bibinfo{year}{2019}).
\bibitem[{Eitel and Ritter(2019)}]{eitel2019testing}
\bibinfo{author}{F.~Eitel}, \bibinfo{author}{K.~Ritter},
\newblock \bibinfo{title}{Testing the robustness of attribution methods for
  convolutional neural networks in {MRI}-based {A}lzheimer's disease
  classification},
\newblock in: \bibinfo{editor}{K.~Suzuki}, \bibinfo{editor}{M.~Reyes},
  \bibinfo{editor}{T.~Syeda-Mahmood}, \bibinfo{editor}{B.~Glocker},
  \bibinfo{editor}{R.~Wiest}, \bibinfo{editor}{Y.~Gur},
  \bibinfo{editor}{H.~Greenspan}, \bibinfo{editor}{A.~Madabhushi} (Eds.),
  \bibinfo{booktitle}{Interpretability of Machine Intelligence in Medical Image
  Computing and Multimodal Learning for Clinical Decision Support},
  \bibinfo{publisher}{Springer International Publishing},
  \bibinfo{address}{Cham}, \bibinfo{year}{2019}, pp. \bibinfo{pages}{3--11}.

\end{thebibliography}







\end{document}